\documentclass[10pt, a4paper, onecolumn, twoside]{amsart}
\usepackage{titlesec, color, amsthm}
\definecolor{blue1}{rgb}{0,0,0}
\renewcommand\thesection{\arabic{section}}

\titleformat{\section}[hang]{\color{blue1}\large\bfseries\sffamily}{\thesection}{0mm}{. }[]
\titleformat{\subsection}[hang] {\color{blue1}\bfseries\sffamily}{\thesubsection}{0em}{. }[]
\titleformat{\subsubsection}[hang] {\color{blue1}\bfseries\sffamily}{\thesubsection}{0em}{. }[]
\titlespacing*{\section}{1em}{3.5ex plus .2ex minus .2ex}{1ex plus .2ex}
\titlespacing*{\subsection}{0em}{3ex plus .2ex minus .2ex}{1ex plus .2ex}
\titlespacing*{\subsubsection}{0em}{3ex plus .2ex minus .2ex}{1ex plus .2ex}
\newenvironment{Keywords}{{\color{blue1}\small\bfseries Keywords.}\footnotesize}{\par \vskip .1in}
\renewenvironment{abstract}{{\color{blue1}\small\bfseries Abstract.}\footnotesize}{\par \vskip .1in}
\makeatletter
\def\@setauthors{
\begingroup 
\def \thanks{\protect\thanks@warning}
\trivlist \centering\footnotesize \@topsep30\p@\relax \advance\@topsep by -\baselineskip
\item\relax \author@andify \authors \def\\{\protect\linebreak} {\color{blue1}\large\authors} \endtrivlist \endgroup}
\def\@settitle{\centering{\color{blue1} \Large \bfseries \bfseries \@title \par}}
\makeatother
\newtheorem{theorem}{\color{blue1} Theorem}
\usepackage[normal, small, up, textfont=it]{caption}
\usepackage{cite} \usepackage{amsmath} \usepackage{algorithm}
\usepackage{amssymb, bm, graphicx, algorithmic, hyperref, enumerate, psfrag, rotating}

\newcommand{\refeq}[1]{(\ref{#1})} 
\newcommand{\Cbb}{\ensuremath{\mathbb{C}}} 
\newcommand{\Rbb}{\ensuremath{\mathbb{R}}} 
\newcommand{\Nbb}{\ensuremath{\mathbb{N}}}
\newcommand{\scp}[2]{\ensuremath{\left\langle #1, #2 \right\rangle}}
\newcommand{\adjoint}{\ensuremath{{\intercal}}}
\renewcommand{\th}{\ensuremath{\text{th}}}
\newcommand{\inv}[1]{\ensuremath{\frac{1}{#1}}}
\renewcommand{\leq}{\ensuremath{\leqslant}}
\renewcommand{\geq}{\ensuremath{\geqslant}}
\newcommand{\norm}[1]{\ensuremath{\| #1\|}}
\newcommand{\abs}[1]{\ensuremath{\left| #1 \right|}}
\newcommand{\ma}[1]{\ensuremath{\mathsf{#1}}}
\renewcommand{\vec}[1]{\ensuremath{\bm{#1}}}
\newcommand{\set}[1]{\ensuremath{\mathcal{#1}}}
\newcommand{\dom}{\ensuremath{{\rm dom}}}

\DeclareMathOperator*{\argmin}{argmin}
\newcommand{\csig}{\ensuremath{x}} 
\newcommand{\sig}{\ensuremath{\vec{x}}} 
\newcommand{\dimSig}{\ensuremath{n}} 
\newcommand{\Dict}{\ensuremath{\ma{\Psi}}} 
\newcommand{\dimSpSig}{\ensuremath{p}} 
\newcommand{\meas}{\ensuremath{\vec{y}}} 
\newcommand{\SenMa}{\ensuremath{\ma{A}}} 
\newcommand{\dimMeas}{\ensuremath{m}} 
\newcommand{\param}{\ensuremath{\vec{\theta}}} 
\newcommand{\paramUp}{\ensuremath{\vec{\bar{{\theta}}}}} 
\newcommand{\paramLow}{\ensuremath{\vec{\underline{\theta}}}} 
\newcommand{\nbParam}{\ensuremath{q}} 
\newcommand{\InterpMa}{\ensuremath{\ma{S}}} 
\newcommand{\kernel}{\ensuremath{\varphi}} 
\newcommand{\trans}{\ensuremath{\tau}} 
\newcommand{\pos}{\ensuremath{\vec{u}}} 
\newcommand{\nbObs}{\ensuremath{l}} 
\newcommand{\noise}{\ensuremath{\vec{n}}} 
\newcommand{\boundNoise}{\ensuremath{\epsilon}} 
\newcommand{\obj}{\ensuremath{{L}}} 
\newcommand{\prior}{\ensuremath{f}} 
\newcommand{\qua}{\ensuremath{Q}} 
\newcommand{\reg}{\ensuremath{\kappa}} 
\newcommand{\ind}{\ensuremath{i}} 
\newcommand{\Ball}{\ensuremath{\mathcal{B}}} 
\newcommand{\ConsParam}{\ensuremath{\Theta}} 
\newcommand{\huber}{\ensuremath{h}_{\mu}} 
\newcommand{\regMove}{\ensuremath{\lambda}} 
\newcommand{\regMoveMin}{\ensuremath{\underline{\lambda}}} 
\newcommand{\regMoveMax}{\ensuremath{\bar{\lambda}}} 
\newcommand{\Jacob}{\ensuremath{\ma{J}}} 
\newcommand{\Hess}{\ensuremath{\ma{H}}} 
\newcommand{\KL}{{Kurdyca-{\L}ojasiewicz }}
\newcommand{\longtitle}{{Robust image reconstruction from multi-view measurements}}
\newcommand{\shortitle}{{Image reconstruction from multi-view measurements}}
\newcommand{\GPlong}{{Gilles~Puy }}
\newcommand{\PVlong}{{Pierre~Vandergheynst }}
\newcommand{\GPshort}{{G.~Puy }}
\newcommand{\PVshort}{{P.~Vandergheynst }}
\newcommand{\EPFL}{{Institute of Electrical Engineering, Ecole Polytechnique F{\'e}d{\'e}rale de Lausanne (EPFL), CH-1015 Lausanne, Switzerland}}
\newcommand{\Hasler}{{This work was partly funded by the Hasler Foundation (project number $12080$)}}

\title[\shortitle]{\longtitle}
\author[\GPshort]{\GPlong}
\author[\PVshort]{\PVlong}
\thanks{\Hasler.}
\thanks{\GPshort and \PVshort are with the \EPFL. E-mail: \url{gilles.puy@epfl.ch} and \url{pierre.vandergheynst@epfl.ch}.}
\begin{document}

\maketitle

\begin{abstract}
We propose a novel method to accurately reconstruct a set of images representing a single scene from few linear multi-view measurements. Each observed image is modeled as the sum of a background image and a foreground one. The background image is common to all observed images but undergoes geometric transformations, as the scene is observed from different viewpoints. In this paper, we assume that these geometric transformations are represented by a few parameters, e.g., translations, rotations, affine transformations, etc.. The foreground images differ from one observed image to another, and are used to model possible occlusions of the scene. The proposed reconstruction algorithm estimates jointly the images and the transformation parameters from the available multi-view measurements. The ideal solution of this multi-view imaging problem minimizes a non-convex functional, and the reconstruction technique is an alternating descent method built to minimize this functional. The convergence of the proposed algorithm is studied, and conditions under which the sequence of estimated images and parameters converges to a critical point of the non-convex functional are provided. Finally, the efficiency of the algorithm is demonstrated using numerical simulations for applications such as compressed sensing or super-resolution. 
\end{abstract}


\section{Introduction}

Multi-view imaging has become more and more popular in the signal processing community during the past years with, for example, the design of camera arrays \cite{afshari12} and the rise of new applications such as $3$D reconstruction of a scene \cite{kubota07}.

In most multi-view imaging applications, the pose of the cameras, or of the sensing systems, is estimated by pre-calibration or by using side-information. In this paper, we study the problem of reconstructing a set of images, representing a single scene, from few measurements made at different viewpoints in absence of (or with inaccurate) prior pose estimation. This situation may occur in different scenarios.

For example, let us imagine that we have a camera that can be sent up to take pictures of the ground and that we are interested in obtaining higher resolution images from these acquisitions. Without more knowledge on the acquisition strategy, we can treat each image independently using a single-frame super-resolution technique. However, if we know that the duration between the capture of two images is not too long, it is highly probable that parts of the scene visible in one frame remain visible in the subsequent ones. Consequently, we can surely benefit from the high inter-correlation between observations to obtain better super-resolved images by jointly reconstructing several frames together, as in, e.g., \cite{farsiu04, vandewalle06}. The inter-correlation may be modeled using geometric transformations, which depend on the camera pose, that register the images with respect to each other. In the absence of any side information, these transformations have to be estimated along with the high resolution images during the reconstruction process.

As another example, let us consider the problem of designing a similar camera as above but with strong power consumption constraints. To design such an energy-efficient system, compressed sensing (CS) is a powerful tool \cite{candes06, mamaghanian12}. Indeed, this theory states that sparse signals can be sampled with just a few linear and non-adaptive measurements. In scenarios where signals related to common phenomena are acquired, it is actually possible to reduce even further the number of measurements by jointly reconstructing the signals with a method exploiting the inter-correlation between them \cite{duarte05, baron06}. A technique, as in \cite{wakin09, park12}, able to estimate both the geometric transformations between the observed images and the images themselves from the compressed measurements can thus help us to meet our power consumption constraints.

In the present work, we propose a novel method that estimates jointly a set of correlated images and the geometric transformations that align these images on each other. This technique also robustly handles the appearance of new objects in the scene and can be applied in different settings, such as the ones described above.

\subsection{Problem formulation}
\label{sec:problem_formulation}
Let $\pos = (u_1, u_2) \in \Rbb^2$ represent the Cartesian coordinates of a point on the Euclidean plane, and $G = \{\pos_k\}_{1 \leq k \leq \dimSig}$ be a square uniform grid of $\dimSig$ points. We model continuous images as functions $\csig \colon \Rbb^2 \rightarrow \Rbb$ living in a space $\{\csig(\pos) = \sum_{i=1}^\dimSig x_i\ \kernel(\pos - \pos_i),\ x_i \in \Rbb,\ \pos_i \in G,\ i = 1, \ldots, \dimSig\}$, where $\kernel \colon \Rbb^2 \rightarrow \Rbb$ is a generating function \cite{unser00}. For simplicity, we suppose that $\kernel(\pos_k - \pos_i) = \delta(\pos_k - \pos_i)$ for all $\pos_k, \pos_i \in G$, so that the discrete image $\sig \in \Rbb^\dimSig$ obtained by sampling $\csig$ on the grid $G$ has samples $x_1, \ldots, x_\dimSig$.

In our setting, several observers provide different observations $\meas_1, \ldots, \meas_\nbObs \in \Rbb^\dimMeas$, with $\dimMeas \leq \dimSig$, of a scene from different viewpoints. As a first approximation, one can consider that these observations describe a single image $\csig_0$, referred as the \emph{background} image hereafter. This image is ``viewed'' from different perspectives and thus undergoes geometric transformations. We consider here that these transformations are unknown and need to be estimated. However, we restrict ourselves to global transformations represented by few parameters, such as translations, rotations, or affine transformations. For each observer $j$, with $j = 1, \ldots, \nbObs$, these transformations are modeled using a function $\trans_{\param_j} \colon \Rbb^2 \rightarrow \Rbb^2$, depending on $\nbParam$ parameters $\param_j \in \Rbb^\nbParam$, that maps the coordinates $\pos$ into new coordinates $\trans_{\param_j}(\pos)$. The background image transformed by $\trans_{\param_j}$ is thus $\csig_0 \circ \trans_{\param_j}$.

To complete the model, we also take into account possible occlusions of the scene. These occlusions may obviously be different from one observer to another. We model them here using $\nbObs$ \emph{foreground} images $\csig_1, \ldots, \csig_\nbObs$ and write the $j^\th$ observed image as $\csig_0 \circ \trans_{\param_j} + \csig_j$.  

Finally, we assume that the observations $\meas_j$ are obtained by linear projection of $\csig_0 \circ \trans_{\param_j} + \csig_j$ onto $\dimMeas$ known functions $a_1^j, \ldots, a_\dimMeas^j \colon \Rbb^2 \rightarrow \Rbb$. These functions model the acquisition system. They can represent a blurring operator for a deconvolution problem, or random waveforms in a compressed sensing setting. The $i^\th$ entry of $\meas_j$ satisfies
\begin{align*}
y_{ij} & = \int_{\Rbb^2} \left(\csig_0 \circ \trans_{\param_j}(\pos)+ \csig_j(\pos)\right) a_i^j(\pos)\ {\rm d}\pos\\
& = \int_{\Rbb^2} \sum_{k=1}^{\dimSig} (x_k^0\ \kernel(\trans_{\param_j}(\pos) - \pos_k) + x_k^j\ \kernel(\pos - \pos_k))\ a_i^j(\pos)\ {\rm d}\pos,
\end{align*}
where $x_k^0$ and $x_k^j$, with $k = 1, \ldots, \dimSig$, are the samples describing $\csig_0$ and $\csig_j$ respectively. To facilitate the implementation of this observation model, we assume that the grid $G$ has a sufficiently high resolution so that the integral above is ``well'' approximated by its Riemann sum on this grid. In the following, we consider that
\begin{align*}
y_{ij} = \sum_{k'=1}^{\dimSig} \sum_{k=1}^{\dimSig} x_k^0\ \kernel(\trans_{\param_j}(\pos_{k'}) - \pos_k) a_i^j(\pos_{k'}) + \sum_{k=1}^{\dimSig} x_k^j\ a_i^j(\pos_{k}),
\end{align*}
where we supposed that the pixels of the grid have a unit square area. Therefore, the measurements vector $\meas_j$ is equal to $\SenMa_j \left( \InterpMa(\param_j) \sig_0 + \sig_j \right)$, where $\SenMa_j = (a_i^j(\pos_{k}))_{ik} \in \Rbb^{\dimMeas \times \dimSig}$, and $\InterpMa \colon \Rbb^\nbParam \rightarrow \Rbb^{\dimSig \times \dimSig}$ satisfies
\begin{align}
\label{eq:interpolation_matrix}
\InterpMa(\param_j) =
\left[
\begin{array}{ccc}
\kernel\left(\trans_{\param_j}(\pos_{1}) - \pos_1\right) & \ldots & \kernel\left(\trans_{\param_j}(\pos_{1}) - \pos_\dimSig\right)\\
\vdots & & \vdots\\
\kernel\left(\trans_{\param_j}(\pos_{\dimSig}) - \pos_1\right) & \ldots & \kernel\left(\trans_{\param_j}(\pos_{\dimSig}) - \pos_\dimSig\right)
\end{array}
\right].
\end{align}
Concatenating all the observations in a single vector, we have
\begin{align}
\label{eq:measurement_model}
\left[ 
\begin{array}{c}
\meas_1\\ \vdots \\ \meas_\nbObs
\end{array}
\right]
=
\left[
\begin{array}{cccc}
\SenMa_1 \InterpMa(\param_1) & \SenMa_1 & \ldots & \ma{0}\\
\vdots & \vdots & \ddots & \vdots \\
\SenMa_\nbObs \InterpMa(\param_\nbObs) & \ma{0} & \ldots & \SenMa_\nbObs \\
\end{array}
\right]
\left[
\begin{array}{c}
\sig_0\\ \vdots \\ \sig_\nbObs
\end{array}
\right]
+
\left[ 
\begin{array}{c}
\noise_1\\ \vdots \\ \noise_\nbObs
\end{array}
\right],
\end{align}
where we introduced $\nbObs$ vectors $\noise_1, \ldots, \noise_\nbObs \in \Rbb^\dimMeas$ to model additive measurement noise. Our goal is now to design a method that estimates the discrete images $\sig = (\sig_0^\adjoint, \ldots, \sig_\nbObs^\adjoint)^\adjoint \in \Rbb^{(\nbObs+1)\dimSig}$, and the transformation parameters $\param^\adjoint = (\param_1^\adjoint, \ldots, \param_\nbObs^\adjoint)^\adjoint \in \Rbb^{\nbObs\nbParam}$, using $\meas= (\meas_1^\adjoint, \ldots, \meas_\nbObs^\adjoint)^\adjoint \in \Rbb^{\nbObs\dimMeas}$ as sole information.

The inter-correlation between the observed images appears explicitly in the measurement model \refeq{eq:measurement_model}. We consider that the observed images share a common signal - the background image - which lives on a low-dimensional manifold. In the literature, other models that exploit the inter-correlation between the measurements have been proposed.

\subsection{Related works}
\label{sec:related_work}
The problem of reconstructing an image from multi-view measurements has been studied by several authors in contexts such as compressed sensing, super-resolution, or robust image alignment. We try here to give a brief overview of the different types of techniques used.

When the measurements are obtained by compressed sensing, some techniques, like ours, rely on the estimation of geometric transformations between images. For example, the authors in \cite{chen09} propose to reconstruct the observed images by assuming that they are sparse in an over-complete dictionary created by scaling, translating and rotating several times a mother waveform. Then, they assume that the decompositions of these images in this dictionary are related by local geometrical transformations. In this method, one image is chosen as a reference image, and the final reconstruction quality depends on this choice. In \cite{wakin09, park12}, the authors suppose that the images live on a low-dimensional manifold. Then they use a manifold lifting algorithm to obtain a first estimation of the transformation parameters and a standard compressed sensing algorithm for the estimation of the images. An alternate minimization method is then proposed to refine the estimation of the transformation parameters and of the images. However, no proof of convergence of the method is provided. Furthermore, to work efficiently, the algorithm requires the use of a specific measurement matrix, implemented with noiselets, to estimate the transformation parameters in a multiscale fashion. We also remark that the presence of occlusions in the images is not considered in this model.

Continuing with compressed sensing, other techniques rely on more general interpolation model between images. In \cite{trocan10, trocan10a}, the images are first estimated with a bidirectional interpolation using two known neighbor images and then corrected using the acquired measurements. An iterative process, alternating between interpolation and reconstruction, is then proposed to refine the estimation of the images. Note that a similar idea is used in \cite{asif12} for dynamic MRI. In \cite{li10}, a slightly simpler model is used. The authors assume that the difference between two images is sparse in a wavelet basis and that the background image is sparse in the gradient domain. Then, a convex minimization problem is solved to reconstruct jointly all the images. In \cite{fu11, dai12}, the interpolation relies on ``disparity maps'' that indicate the correspondences between the pixels of two images. These maps are used to estimate the images from the measurements, and the newly estimated pictures can be used to refine the disparity maps. The process is then iterated several times. We remark that the authors do not provide any proof of convergence. We note that the authors deal with the problem of occlusions by supposing that they are sparse in image space. Let us also mention that a similar idea is proposed in \cite{thirumalai12}. Finally, in \cite{sankaranarayanan10}, the authors use a linear dynamical system to model the intercorrelation between frames, and reconstruct a video sequence.

When the matrices $\SenMa_1, \ldots, \SenMa_m$ implement a blurring and downsampling operator, one can try to obtain a high-resolution background image from the images observed at low-resolution by solving \refeq{eq:measurement_model}. Super-resolution from multiple frames has also been widely studied and several image priors and correlation models between images have been proposed to solve this problem. As before, some techniques model the correlation between images using geometric transformations \cite{farsiu04, vandewalle06} while other ones use more general correlation models \cite{protter09, shen07}.

Let us now highlight a similarity between our model and the one used in \cite{peng10, peng12} for robust image alignment, which consists in aligning a set of images despite the presence of large occlusions. With the measurement model \refeq{eq:measurement_model}, one can attempt to solve this problem by initializing the vectors $\meas_1, \ldots, \meas_\nbObs$ with the observed images and the observations matrices $\SenMa_1, \ldots, \SenMa_\nbObs$ with the identity. Then, the algorithm should estimate the background image, the transformation parameters that align this background image on the observed ones, and the foreground images. In \cite{peng10, peng12}, a different approach, {\color{black} called RASL}, is proposed. The authors concatenate the initial images $\meas_1, \ldots, \meas_\nbObs$ in a matrix and explain that, after alignment of these images, this matrix can be decomposed as the sum of a low-rank matrix (modeling correlated components) and a sparse matrix (modeling occlusions). This model serves as the criterion to align the images. Note that in the limit where the rank of the final matrix modeling the correlated components is $1$, their method, like ours, also decomposes the initial images into a single background image and $\nbObs$ foreground images. {\color{black} Finally, let us mention that even though TILT \cite{zhang12} resembles RASL at first sight, their purpose is different. TILT is a method to correct and align low-rank textures and takes as input only one image. This is a different problem than the one studied here.}

\subsection{Main contribution and organization of the paper}
\label{sec:contributions}
{\color{black} In the present work, we propose a novel algorithm for \emph{joint registration and reconstruction} of a set of misaligned images from linear measurements and, in contrast to the methods described above, provide a \emph{convergence proof} of our algorithm. Furthermore, the model used permits us to easily deal with the problem of occlusions. It is also general enough to allow us to handle many image reconstruction problems. In order to study the efficiency of our algorithm, we use extensive numerical simulations in several applications. These numerical experiments confirm that our algorithm is competitive with well-known methods for robust image alignment, compressed sensing, and super-resolution. 
}

In Section \ref{sec:method}, we identify the solution of the multi-view imaging problem with the global minimizer of a non-convex functional, and propose an algorithm for the minimization of this functional. In Section \ref{sec:convergence}, we study the convergence of the reconstruction method, and prove that the sequence of \emph{estimates} (images and transformation parameters) converges to a critical point of the functional. The efficiency of the algorithm is then demonstrated experimentally in Section \ref{sec:results} for several problems. In the spirit of reproducible research, our code\footnote{We used the SPARCO toolbox available at \url{www.cs.ubc.ca/labs/scl/sparco/} \cite{berg07}.} and data needed to reproduce the results presented in this paper can be downloaded openly at \url{lts2www.epfl.ch/people/gilles/softwares}. We finally conclude in Section \ref{sec:conclusion}. Some results required to prove the convergence of the algorithm are gathered in the appendix.

Note that we already briefly studied the problem of reconstructing a set of images from multi-view measurements in \cite{puy12a}, where the measurement model \refeq{eq:measurement_model} was also used. However, the technique proposed to reconstruct the images was based on the Bregman iterative regularization procedure, which is different from the one used here. Furthermore, the convergence results presented in \cite{puy12a} essentially concern the decrease of the objective value and, unlike here, do not ensure the convergence of the sequence of estimated images and parameters.

\subsection{Notations and definitions}
\label{sec:notations}
The Euclidean scalar product of $\Rbb^n$ is denoted by $\scp{\vec{\cdot}}{\vec{\cdot}}$ and $\norm{\vec{\cdot}}_2$ is the corresponding $\ell_2$-norm. The $\ell_1$-norm of a vector $\vec{x} = (x_i)_{1 \leq i \leq n} \in \Rbb^n$ is defined as $\norm{\vec{x}}_1 = \sum_{i=1}^{n} \abs{x_i}$.

Let $C$ be a non-empty closed convex subset of $\Rbb^n$. The indicator function of $C$ is denoted by $\ind_C \colon \Rbb^n \rightarrow \Rbb \cup \{+ \infty\}$. It is a proper lower semicontinuous convex function that satisfies $\ind_C(\vec{x}) = 0$ if $\vec{x} \in C$, and $\ind_C(\vec{x}) = +\infty$ otherwise.

Let $f \colon \Rbb^n \rightarrow \Rbb \cup \{+ \infty\}$ be a proper lower semicontinuous function. The domain of $f$ is denoted and defined by $\dom f = \{\vec{x} \in \Rbb^n: f(\vec{x}) < + \infty\}$. The subdifferential of $f$ at $\vec{x} \in \dom f$ is denoted by $\partial f(\vec{x})$ (see, e.g., Section 2 of \cite{attouch10} for the definition). Note that $\partial f(\vec{x}) = \emptyset$ if $\vec{x} \notin \dom f$. A necessary (but not sufficient) condition for $\vec{x}^* \in \Rbb^n$ to be a minimizer of $f$ is $\partial f(\vec{x}^*) \ni \vec{0}$. A point that satisfies the last condition is called a critical point of $f$. 

Finally, $\set{C}^k$ denotes the set of functions that are $k$ times continuously differentiable.

\section{A non-convex optimization technique}
\label{sec:method}
In this section, we first identify the solution of the compressed multi-view imaging problem with the minimizer of a non-convex functional and propose an algorithm to minimize this functional.

\subsection{Solution as minimizer of a non-convex functional}
Let $\boundNoise \geq 0$ be an upper bound on the energy of the measurement noise, i.e., $\norm{\noise}_2 \leq \boundNoise$. Our objective is to find a set of parameters $\param^*$ and images $\sig^*$ that satisfy the constraint $\norm{\SenMa(\param^*)\,\sig^* - \meas}_2 \leq \boundNoise$, where 
\begin{align*}
\SenMa(\param) = 
\left[
\begin{array}{cccc}
\SenMa_1 \InterpMa(\param_1) & \SenMa_1 & \ldots & \ma{0}\\
\vdots & \vdots & \ddots & \vdots \\
\SenMa_\nbObs \InterpMa(\param_\nbObs) & \ma{0} & \ldots & \SenMa_\nbObs \\
\end{array}
\right]
\in
\Rbb^{\nbObs\dimMeas \times (\nbObs+1)\dimSig}.
\end{align*}
As infinitely many solutions satisfy this constraint, prior information is needed to restrict the set of admissible solutions. To regularize this ill-posed inverse problem, we can, for example, search for images sparse in a wavelet basis by minimizing the $\ell_1$-norm of their wavelet coefficients, or seek a solution that minimizes the Total Variation (TV) norm if the images are piecewise constant \cite{rudin92}. We may also require the transformation parameters $\param_j$, with $j=1,\ldots,\nbObs$, to belong to some convex sets $\ConsParam_j = \{\param_j \in \Rbb^{\nbParam} :\ \paramLow_j \leq \param_j \leq \paramUp_j\}$ where $\paramLow_j \in \Rbb^\nbParam$ and $\paramUp_j \in \Rbb^\nbParam$ are pre-defined upper and lower bounds\footnote{Let $\paramUp = (\bar{\theta}_i)_{1 \leq i \leq \nbParam} \in \Rbb^\nbParam$, $\param = (\theta_i)_{1 \leq i \leq \nbParam} \in \Rbb^\nbParam$, $\param \leq \paramUp$ means that $\theta_i \leq \bar{\theta}_i$ for all $i \in \{1, \ldots, \nbParam\}$.}. Therefore, an estimate of the images and transformation parameters can be obtained by solving a minimization problem of the form
\begin{align}
\label{eq:minimization_problem}
\min_{(\sig, \param)}
\prior(\sig) + \reg\ \norm{\SenMa(\param)\,\sig - \meas}_2^2 \;
\text{ subject to } \; \param \in \ConsParam,
\end{align}
where $\prior \colon \Rbb^{(\nbObs+1)\dimSig} \rightarrow \Rbb \cup \{+ \infty\}$ is a proper lower-semicontinuous convex function, $\reg^{-1} > 0$ is a regularizing parameter that should be adjusted with the noise level $\epsilon$, and $\ConsParam = \{\param = (\param_1^\adjoint, \ldots, \param_\nbObs^\adjoint)^\adjoint \in \Rbb^{\nbObs\nbParam} :\ \param_j \in  \ConsParam_j,\ j = 1, \ldots, \nbObs\}$. 

Problem \refeq{eq:minimization_problem} is non-linear and finding a global minimizer is not trivial. Nevertheless, one can still try to solve this problem using an alternating minimization technique such as the block coordinate descent method \cite{tseng01}. There is however no guarantee to converge to a critical point of Problem \refeq{eq:minimization_problem} with this technique. Recently, Attouch \emph{et al.} proposed in \cite{attouch10} a new type of algorithms for the minimization of non-convex functionals and, more importantly, managed to prove that the sequence generated by these algorithms converges to a critical point for a wide class of functionals. As stated by the authors, their algorithm can be interpreted as a proximal regularization of the Gauss-Seidel method where cost-to-move functions are added in the minimization procedure. These convergence results were then generalized to inexact descent methods that satisfy a sufficient decrease condition \cite{attouch11}. Based on this work, we develop a minimization method for problem \refeq{eq:minimization_problem} and prove that the generated sequence converges to a critical point $(\sig^*, \param^*)$ of the functional $\obj \colon \Rbb^{(\nbObs+1)\dimSig} \times \Rbb^{\nbObs\nbParam} \rightarrow \Rbb \cup \{+ \infty\}$ defined as
\begin{align}
\label{eq:objective_function} 
\obj(\sig, \param) =  \prior(\sig) + \reg\ \norm{\SenMa(\param)\,\sig - \meas}_2^2 + \ind_{\ConsParam}(\param).
\end{align}
{ \color{black}
Let us recall that a global minimizer of $\obj$ is also a critical point but that a critical point of $\obj$ is not necessarily a global minimizer. The point $(\sig^*, \param^*)$ might thus only be a local minimizer or a saddle point of Problem \refeq{eq:minimization_problem}.}

The proposed algorithm is a descent method. It generates a sequence of estimates $(\sig^{k}, \param^{k})_{k \in \Nbb}$ such that $\obj(\sig^{k+1}, \param^{k+1}) \leq \obj(\sig^k, \param^k)$ for all $k \in \Nbb$, and consists of two steps. We start by describing each of these steps and study the convergence of the algorithm in Section \ref{sec:convergence}.

\subsection{First step of the algorithm}
\label{sec:first_step}
Let $(\sig^k, \param^k) \in \Rbb^{(\nbObs+1)\dimSig} \times \ConsParam$ be the estimates obtained after $k$ iterations of the algorithm. The first step consists in finding a new estimate $\sig^{k+1} \in \Rbb^{(\nbObs+1)\dimSig}$ that decreases the value of the objective function $\obj$ while keeping $\param^k$ fixed. Let us choose $\sig^{k+1}$ as a solution of
\begin{align}
\label{eq:step1}
\min_{\sig \in \Rbb^{(\nbObs+1)\dimSig}} \prior(\sig) + \reg\ \norm{\SenMa(\param^k)\,\sig - \meas}_2^2 + \frac{\regMove_{\sig}}{2}\ g(\sig - \sig^{k}),
\end{align}
where $\regMove_{\sig} > 0$, and $g \colon \Rbb^{(\nbObs+1)\dimSig} \rightarrow \Rbb$ is proper lower-semicontinuous convex function such that $g(\sig) \geq 0$ for all $\sig \in \Rbb^{(\nbObs+1)\dimSig}$ {\color{black} and $g(\vec{0}) = 0$}. It is clear that \refeq{eq:step1} implies
\begin{align}
\label{eq:decrease_step_1}
\obj(\sig^{k+1}, \param^{k}) + \frac{\regMove_{\sig}}{2}\ g(\sig^{k+1} - \sig^{k}) \leq \obj(\sig^{k}, \param^{k}).
\end{align}
Hence $\obj(\sig^{k+1}, \param^{k}) \leq \obj(\sig^k, \param^k)$ with a decrease of $\regMove_{\sig}\, g(\sig^{k+1} - \sig^{k})/2$, at least. Note that this first minimization problem is convex in $\sig$ and can be solved using, e.g., the algorithms presented in \cite{beck09a, combettes11, raguet13}. 

In Problem \refeq{eq:step1}, the function $\regMove_{\sig}\, g(\bm{\cdot})$ acts as a proximal term. It ensures a sufficient decrease of the functional $\obj$ at each iteration and is essential to prove the convergence of the sequence $(\sig^k, \param^k)_{k \in \Nbb}$ to a critical point. This function also provides, to some extent, a control on the evolution of the generated sequence, and we realized that it must be chosen carefully to achieve good {\color{black} and stable results in all the scenarios tested}. Note that in the work \cite{attouch10} of Attouch \emph{et al.}, the function $g$ is a squared $\ell_2$-norm.

From the multi-view measurements $\meas_1, \ldots, \meas_\nbObs$, the algorithm should reconstruct the background image, separate the occlusions (foreground), and provide the transformation parameters that register the initial images between them. This problem is solved here by alternatively refining the estimation of the images $\sig$ (background and foreground) and of the transformation parameters $\param$ (image registration). We noticed that to improve the quality of the registration, it is better to have a reconstruction of the images in a mutiscale fashion, as in \cite{wakin09, park12}. In these papers, the authors use a dedicated measurement matrix that allows this type of reconstruction. In the current paper, this goal is achieved by choosing a function $g$ that favors the reconstruction of the coarse scales first and then of the fine scales. There is thus no constraint on the choice of the measurement matrices anymore.

In order to have a reconstruction of the images from coarse to fine scales, we use a wavelet tight-frame $\ma{W} \in \Rbb^{\dimSig \times \dimSpSig}$, with $\dimSpSig \geq \dimSig$, i.e., $\ma{W}\ma{W}^\adjoint$ is equal to $\ma{I}_{\dimSig} \in \Rbb^{\dimSig \times \dimSig}$, the identity matrix. One should remark that most of the largest wavelet coefficients of a natural image live at the coarsest scales and that the amplitude of these coefficients decreases when going to finer scales (except at the few singularities). To achieve our goal, we need a function $g$ that favors a reconstruction of the coarse scales first, by selecting the biggest wavelet coefficients, and then of the fine scales, by selecting the smallest wavelet coefficients. Let $\Dict \in \Rbb^{(\nbObs+1)\dimSig \times (\nbObs+1)\dimSpSig}$ be the block-diagonal matrix built by repeating $\nbObs + 1$ times the matrix $\ma{W}$ on the diagonal. We have $\Dict\Dict^\adjoint = \ma{I}_{(\nbObs + 1)\dimSig}$. Take $\sig^{k+1}$ as a minimizer of $\prior(\sig) + \reg \, \norm{\SenMa(\param^k)\,\sig - \meas}_2^2 + \regMove_{\sig} \norm{\Dict^\adjoint (\sig - \sig^k)}_1/2$, and remember that the $\ell_1$-norm favors solutions with few large coefficients. Hence, the function $\prior$ forces $\sig^{k+1}$ to fit our prior, the quadratic term forces $\sig^{k+1}$ to be consistent with the observations, and the cost-to-move function $\sig \mapsto \norm{\Dict^\adjoint (\sig - \sig^k)}_1$ imposes that $\sig^{k+1}$ differs from $\sig^{k}$ by only a few large wavelet coefficients. Starting from $\sig^0 = \vec{0} \in \Rbb^{(\nbObs+1)\dimSig}$ at $k=0$, the first estimated images $\sig^1$ have thus a sparse decomposition in the wavelet tight-frame $\ma{W}$. The sparsity increases with the parameter $\regMove_{\sig}$. As we are reconstructing natural images, these few large coefficients mainly live at large scales and $\sig^1$ is a coarse approximation of the solution. The estimation of the images is then refined at the next iteration, and the cost-to-move function favors the selection of the next biggest wavelet coefficients living at finer scales. Thanks to this procedure, we have a reconstruction of the images from coarse to fine scales. This behavior will be illustrated in Section \ref{sec:results}. 

Note that to be able to prove the convergence of the sequence $(\sig^k, \param^k)_{k \in \Nbb}$ to a critical point of $\obj$ with the results presented in \cite{attouch10, attouch11}, we actually substitute the Huber function for the $\ell_1$-norm as cost-to-move function. The Huber function is a smooth approximation of the $\ell_1$-norm which depends on a smoothing parameter $\mu > 0$. This parameter can be chosen small so that both functions are nearly identical. Let $\vec{\alpha} = (\alpha_i)_{1 \leq i \leq \dimSpSig} \in \Rbb^{(\nbObs+1)\dimSpSig}$, the Huber function $\huber: \Rbb^{(\nbObs+1)\dimSpSig} \rightarrow \Rbb$ satisfies
\begin{align*}
\huber(\vec{\alpha}) = \sum_{i=1}^{(\nbObs+1)\dimSpSig} h_i,
\quad \text{ where } \quad
h_i = \left\{
\begin{array}{ll}
\alpha_i^2/(2 \mu), & \text{if} \abs{\alpha_i} < \mu,\\
\abs{\alpha_i} + \mu/2, & \text{otherwise},
\end{array}
\right. 
\forall i \in \{1, \ldots, (\nbObs+1)\dimSpSig\}.
\end{align*}
From now on, the term $g(\sig - \sig^k)$ in Problem \refeq{eq:step1} is replaced by $\huber(\Dict^\adjoint(\sig - \sig^{k}))$ {\color{black} (including in our implementation of the algorithm).} 


\subsection{Second step of the algorithm}
The second step consists in updating the transformation parameters to further decrease the value of the objective function $\obj$. As the functions $\param \mapsto \norm{\SenMa(\param^k)\,\sig - \meas}_2^2 $ and $\ind_{\ConsParam}$ are separable in $\param_j$, with $j = 1, \ldots, \nbObs$, we optimize the transformation parameters separately for each observation. 

Let $(\sig^k, \param^k) \in \Rbb^{(\nbObs+1)\dimSig} \times \ConsParam$ be the estimates obtained after $k$ iterations of the algorithm and $\sig^{k+1} \in \Rbb^{(\nbObs+1)\dimSig}$ be the solution of problem \refeq{eq:step1}. To simplify the notations, we introduce $\nbObs$ new functions $\qua_j^{k+1} \colon \Rbb^\nbParam \rightarrow \Rbb$, with $j = 1, \ldots, \nbObs$, defined as 
\begin{align}
\label{eq:fidelity_term}
\qua^{k+1}_j(\param_j) = \norm{\SenMa_j\InterpMa(\param_j) \sig_0^{k+1} + \SenMa_j \sig_j^{k+1} - \meas_j}_2^2.
\end{align}
Note that $\norm{\SenMa(\param) \sig^{k+1} - \meas}_2^2 = \sum_{j=1}^\nbObs \qua^{k+1}_j(\param_j)$ and that $\ind_{\ConsParam}(\param) = \sum_{j=1}^\nbObs \ind_{\ConsParam_j}(\param_j)$. We choose to update the transformation parameters with the following projected Newton-like method \cite{bertsekas82, schmidt11}.
 
Let us assume that the entries of the matrix $\InterpMa(\param_j)$ are differentiable with respect to the transformation parameters. The first order Taylor expansion of $\InterpMa(\param_j)\sig_0^{k+1}$ at $\param_i^{k}$ is $\InterpMa(\param_i^{k})\sig_0^{k+1} + \Jacob(\param_j^{k}) (\param_j - \param_j^{k})$ with 
\begin{align*}
\Jacob(\param^{k}_j) = \left(\partial_{\theta_{1j}} \InterpMa(\param_j^k)\, \sig_0^{k+1}\ ,\ \ldots\ ,\ \partial_{\theta_{\nbParam j}} \InterpMa(\param_j^k)\, \sig_0^{k+1}\right) \in \Rbb^{\dimSig \times \nbParam}.
\end{align*}
Therefore, 
\begin{align*}
\qua_j^{k+1}(\param_j^{k}) + \scp{\nabla\qua_j^{k+1}(\param_j^{k})}{\param_j - \param_j^{k}} +\ \norm{\SenMa_j\Jacob(\param^{k}_j)(\param_j - \param_j^{k})}_2^2,
\end{align*}
with
\begin{align*}
\nabla\qua_j^{k+1}(\param_j^{k}) = 2 \left(\SenMa_j\Jacob(\param^{k}_j)\right)^\adjoint \left(\SenMa_j\InterpMa(\param_j^k)\sig_0^{k+1} + \SenMa_j\sig_j^{k+1} - \meas_j\right),
\end{align*}
is a second order approximation of $\qua_j^{k+1}$ at $\param_j^{k}$. The positive semi-definite matrix 
\begin{align}
\label{eq:hessian}
\Hess(\param^{k}_j) = 2 \left(\SenMa_j\Jacob(\param^{k}_j)\right)^\adjoint \left(\SenMa_j\Jacob(\param^{k}_j)\right),
\end{align}
can be viewed as an approximation of the Hessian of the function $\qua_j^{k+1}$. To update the transformation parameters, we choose to minimize this quadratic approximation to which we add another quadratic term that ensures a decrease of the objective function. We take as next estimate of the transformation parameters
\begin{align}
\label{eq:step2}
\param_j^{k+1} = \argmin_{\param_j \in \ConsParam_j}&\ \scp{\nabla\qua_j^{k+1}(\param_j^{k})}{\param_j - \param_j^{k}} + \inv{2} \scp{\param_j - \param_j^{k}}{\left[\Hess(\param_j^k) + 2^i \regMove_{\param} \ma{I}_{\nbParam} \right](\param_j - \param_j^{k})},
\end{align}
where $\ma{I}_{\nbParam} \in \Rbb^{\nbParam \times \nbParam}$ is the identity matrix, $\regMove_{\param}>0$, and $i$ is the smallest positive integer such that
\begin{align}
\label{eq:condition_lambda}
\qua_j^{k+1}(\param_j^{k+1}) \leq\ & \qua_j^{k+1}(\param_j^k) + \scp{\nabla\qua_j^{k+1}(\param_j^{k})}{\param_j^{k+1} - \param_j^{k}} \nonumber \\ 
& +\ \inv{2}\scp{\param_j^{k+1} - \param_j^{k}}{\left[\Hess(\param_j^k) + (2^i - 1) \regMove_{\param} \ma{I}_{\nbParam}\right](\param_j^{k+1} - \param_j^{k})}.
\end{align}
The existence of such an $i$ is discussed in the next section\footnote{Note that if one is able to compute the Lipschitz constant $\Lambda_j^{k+1}$ of $\nabla \qua_j^{k+1}$, then $\Lambda_j^{k+1}$ can be substituted for $2^i \regMove_{\param}$ and the algorithm also converges.}. One can remark that if the minimization \refeq{eq:step2} was performed over all $\Rbb^\nbParam$ instead of $\ConsParam_j$, the solution would be $\param_j^{k+1} = \param_j^{k} - [\Hess(\param_j^k) + 2^i \regMove_{\param} \ma{I}_{\nbParam}]^{-1}\nabla\qua_j^{k+1}(\param_j^{k})$, as in a Newton method with Hessian $\Hess(\param_j^k) + 2^i \regMove_{\param} \ma{I}_{\nbParam}$. One can also note that if we had chosen $\Hess(\param_j^k) = \vec{0}$ then $\param_j^{k+1}$ would be obtained with a simple projected gradient update. We noticed however that the Newton-like update \refeq{eq:step2}, which can, for example, be solved using the algorithms presented in \cite{beck09a}, yields more accurate results.

To check that our choice of new transformation parameters decreases the value of the objective function $\obj$, one just has to combine \refeq{eq:step2} and \refeq{eq:condition_lambda} to realize that
\begin{align*}
\qua_j^{k+1}(\param_j^{k+1}) & + \frac{\regMove_{\param}}{2} \norm{\param_j^{k+1} - \param_j^{k}}_2^2\
\leq \nonumber \\
& \qua_j^{k+1}(\param_j^{k}) + \scp{\nabla\qua_j^{k+1}(\param_j^{k})}{\param_j - \param_j^{k}} + \inv{2}\scp{\param_j - \param_j^{k}}{\left[\Hess(\param_j^k) + 2^i \regMove_{\param} \ma{I}_{\nbParam}\right](\param_j - \param_j^{k})},
\end{align*}
for any $\param_j \in \ConsParam_j$. Choosing $\param_j = \param_j^k$ in the last inequality, multiplying it by $\reg$, and summing all the inequalities obtained for $j$ from $1$ to $\nbObs$ yields
\begin{align}
\label{eq:decrease_step_2}
\obj(\sig^{k+1}, \param^{k+1}) + \frac{\reg\,\regMove_{\param}}{2} \norm{\param^{k+1} - \param^{k}}_2^2 \leq \obj(\sig^{k+1}, \param^{k}).
\end{align}
Therefore $\obj(\sig^{k+1}, \param^{k+1}) \leq \obj(\sig^{k+1}, \param^{k})$ with a decrease of $\kappa \,\regMove_{\param}\norm{\param^{k+1} - \param^{k}}_2^2/2$, at least.

\section{Convergence analysis}
\label{sec:convergence}
In this section, we analyze the convergence of the optimization method described in Section \ref{sec:method}. We first present sufficient conditions under which the generated sequence converges to a critical point of $\obj$. Then, we describe several cases where these conditions are satisfied. In particular, we study different types of geometric transformations which meet these requirements. Finally, we discuss the influence of the cost-to-move parameters on the final reconstruction quality.

\begin{algorithm}
\caption{}
\begin{algorithmic}
\STATE {\bfseries Inputs:} measurements $\vec{y}\in \Rbb^{\nbObs \dimMeas}$, wavelet tight-frame $\Dict \in \Rbb^{(\nbObs+1)\dimSig \times (\nbObs+1)\dimSpSig}$, regularization parameter $\reg > 0$, parameters $(\regMove_{\sig}^k)_{k\in \Nbb}, \regMove_{\param} > 0$, and bounds $\paramUp \geq \paramLow$. 
\vspace{1mm} 
\STATE {\bfseries Initializations:} set $k=0$, $\sig^0 = \vec{0} \in \Rbb^{(\nbObs+1)\dimSig}$, and $\param^0 \in \ConsParam$. 
\vspace{1mm} 
\REPEAT 
\STATE 1) Set
\begin{align*}
\sig^{k+1} \gets \argmin_{\sig \in \Rbb^{(\nbObs+1)\dimSig}} \; \obj(\sig, \param^{k}) + \frac{\regMove_{\sig}^k}{2}\ \huber\left(\Dict^\adjoint(\sig - \sig^{k})\right).
\end{align*}	
\STATE 2) For all $j=1, \ldots, \nbObs$, set
\begin{align*}
\param_j^{k+1} \gets \argmin_{\param_j \in \ConsParam_j}\ \scp{\nabla\qua_j^{k+1}(\param_j^{k})}{\param_j - \param_j^{k}} + \inv{2}\scp{\param_j - \param_j^{k}}{\left[\Hess(\param_j^k) + 2^i \regMove_{\param} \ma{I} \right](\param_j - \param_j^{k})},
\end{align*}
where $\qua_j^{k+1}$ is defined in \refeq{eq:fidelity_term}, $\Hess(\param_j^k)$ is defined in \refeq{eq:hessian}, and $i$ is the smallest positive integer such that \refeq{eq:condition_lambda} is satisfied.
\vspace{1mm} 
\STATE 3) $k \gets {k+1}$ 
\vspace{1mm} 
\UNTIL{convergence or $k \geq k_{\rm max}$}
\vspace{1mm} 
\STATE {\bfseries Outputs:} Estimated images $\sig^*$ and transformation parameters $\vec{\param}^*$.
\end{algorithmic}
\end{algorithm}

\subsection{General convergence result}
Let us consider Algorithm $1$ which is a summary of the optimization method described in the previous section. We are now in position to state our main convergence result.
\begin{theorem}
\label{th:main}
Let $\obj$ be the objective function defined in \refeq{eq:objective_function} with $\reg > 0$. Assume that $\obj$ is bounded below, that the entries of the matrix $\InterpMa$ defined in \refeq{eq:interpolation_matrix} are twice continuously differentiable with respect to the transformation parameters, that $\Dict \in \Rbb^{(\nbObs+1)\dimSig \times (\nbObs+1)\dimSpSig}$ satisfies $\Dict\Dict^\adjoint = \ma{I}_{(\nbObs + 1)\dimSig}$, and that $0 < \regMoveMin \leq \regMove_{\sig}^k, \regMove_{\param} \leq \regMoveMax$ for all $k \in \Nbb$. Then, the sequence of estimates $(\sig^k, \param^k)_{k \in \Nbb}$ generated by Algorithm $1$ is correctly defined and the following statements hold.
\begin{enumerate}
\item For all $k \geq 0$,
\begin{align}
\label{eq:sufficent_decrease}
\obj(\sig^{k+1},\ & \param^{k+1})  + \frac{\regMoveMin}{2} \bigg[\reg\ \norm{\param^{k+1} - \param^{k}}_2^2 +  \huber(\Dict^{\adjoint}(\sig^{k+1}  - \sig^{k})) \bigg] \leq \obj(\sig^{k}, \param^{k}). 
\end{align}
Hence $L(\sig^k, \param^k)$ does not increase.
\item The sequences $(\sig^{k+1} - \sig^{k})_{k \in \Nbb}$ and $(\param^{k+1} - \param^{k})_{k \in \Nbb}$ converge. Indeed,
\begin{align}
\lim_{k \rightarrow + \infty} \norm{\sig^{k+1} - \sig^{k}}_2 + \norm{\param^{k+1} - \param^{k}}_2 = 0.
\end{align}
\item Assume that $\obj$ has the \KL property (see Definition $3.2$ in \cite{attouch10}). Then, if the sequence $(\sig^k)_{k \in \Nbb}$ is bounded, the sequence $(\sig^k, \param^k)_{k \in \Nbb}$ converges to a critical point $(\sig^*, \param^*)$ of $\obj$.
\end{enumerate}
\end{theorem}
\vspace{2mm}
\begin{proof}
Let us start by showing that the sequence generated by Algorithm $1$ is well-defined. For the first step of the algorithm, note that, for all $(\tilde{\sig}, \tilde{\param}) \in  \Rbb^{(\nbObs+1)\dimSig} \times \Rbb^{\nbObs\nbParam}$ and $\tilde{\regMove}>0$, $\sig \mapsto L(\sig, \tilde{\param})$ is a proper lower semicontinuous convex function which is bounded below and that $\sig \mapsto \tilde{\regMove}\, \huber(\Dict^{\adjoint}(\sig - \tilde{\sig}))/2$ is a proper lower semicontinuous convex function which is coercive. Therefore, the function $\sig \mapsto L(\sig, \tilde{\param}) + \tilde{\regMove}\, \huber(\Dict^{\adjoint}(\sig - \tilde{\sig}))/2$ has a minimizer (Corollary $11.15$, \cite{bauschke11}). For the second step of the algorithm, remark that, for all $(\tilde{\sig}_0, \tilde{\sig}_j) \in  \Rbb^{\dimSig} \times \Rbb^{\dimSig}$, the function $\qua_j \colon \param_j \mapsto \norm{\SenMa_j\InterpMa(\param_j) \tilde{\sig}_0 + \SenMa_j \tilde{\sig}_j - \meas_j}_2^2$ is $\set{C}^1$ with Lipschitz continuous gradient on the closed and bounded convex set $\ConsParam_j$. Let $\Lambda_j$ be this Lipschitz constant, whose value \emph{a priori} depends on $(\tilde{\sig}_0, \tilde{\sig}_j)$. Using the descent lemma (Lemma $3.1$, \cite{attouch11}), we have
\begin{align*}
\qua_j(\param_j^1) \leq \qua_j(\param_j^2) + \scp{\nabla\qua_j(\param_j^2)}{\param_j^1 - \param_j^2} + \frac{\Lambda_{j}}{2}\ \norm{\param_j^1 - \param_j^2}_2^2,
\end{align*}
for any two points $(\param_j^1, \param_j^2) \in \ConsParam_j^2$. This proves the existence of an integer $i$ such that \refeq{eq:condition_lambda} is satisfied at each second step of Algorithm $1$. The existence of a minimizer in Problem \refeq{eq:step2} is then proved by using, e.g., Proposition $11.14$ in \cite{bauschke11}. An induction finally shows that sequence $(\sig^k, \param^k)_{k \in \Nbb}$ is well-defined.

Inequality $\refeq{eq:sufficent_decrease}$ follows by combination of \refeq{eq:decrease_step_1} and \refeq{eq:decrease_step_2} and by using the fact that $\regMove_{\sig}^k, \regMove_{\param} \geq \regMoveMin$ for all $k \in \Nbb$. Then, summing these inequalities from $k=0$ to $K$ yields
\begin{align*}
\frac{\regMoveMin}{2} \sum_{k=0}^K \reg\ \norm{\param^{k+1} - \param^{k}}_2^2\ +\ & \huber(\Dict^\adjoint(\sig^{k+1} - \sig^{k}))
\leq \obj(\sig^{0}, \param^{0}) - \obj(\sig^{K+1}, \param^{K+1})
\end{align*}
As $L$ is bounded below and $\obj(\sig^{0}, \param^{0}) = \reg \norm{\meas}_2^2$, it is clear that the righthand side of the previous inequality is also bounded. Consequently,
\begin{align*}
\sum_{k=0}^{+ \infty} \huber(\Dict^\adjoint(\sig^{k+1} - \sig^{k})) + \reg\ \norm{\param^{k+1} - \param^{k}}_2^2 < + \infty,
\end{align*}
and, using the definition of $\huber$ and the fact that $\Dict\Dict^\adjoint = \ma{I}_{(\nbObs + 1)\dimSig}$, the second point of the theorem holds.

The proof of the third point make use of results established by Attouch \emph{et al.} in \cite{attouch11} and can be found in Appendix \ref{app:proof}. 
\end{proof}

{\color{black} Note that to study the convergence of Algorithm $1$, we considered, for simplicity, that the minimization problems \refeq{eq:step1} and \refeq{eq:step2} can be minimized exactly. However, the results presented by Attouch \emph{et al.} in \cite{attouch11} also hold with inexact descent methods. This suggests that Algorithm $1$  also convergences with inexact minimizations of \refeq{eq:step1} and \refeq{eq:step2}, upon the introduction of few additional requirements on the ``quality'' of these minimizations.}

The third point of Theorem \ref{th:main} applies if $\obj$ has the \KL property. As explained in \cite{attouch10}, this property is satisfied by several types of functions including semi-algebraic ones. A function $g:\Rbb^n \rightarrow \Rbb \cup \{+ \infty\}$ is semi-algebraic if its graph $\{(\sig, t) \in \Rbb^n \times \Rbb : g(\sig) = t\}$ is a semi-algebraic subset of $\Rbb^{n+1}$, i.e., if it can be written as a finite union of sets of the form $\{\sig \in \Rbb^{n+1}: p_i(\sig) = 0,\ q_i(\sig) < 0,\ i = 1, \ldots, r\}$, where $p_i$ and $q_i$ are polynomials. The $\ell_1$-norm is thus an example of such a function. Note that the indicator function of a semi-algebraic set is semi-algebraic, and that the set of semi-algebraic functions is stable under basics operations such as sums, products or compositions (Section $4.3$, \cite{attouch10}). 

In the definition of the objective function $\obj$, the set $\ConsParam$ is semi-algebraic. Consequently, $\obj$ satisfies the \KL property if, e.g., $\prior$ and the fidelity term $(\param, \sig) \mapsto \norm{\SenMa(\param) \sig - \meas}_2^2$ are semi-algebraic. For $\prior$, one can for example use $\prior(\sig) = \norm{\ma{\Phi} \sig}_1$ with any real matrix $\ma{\Phi}$. This choice also ensures that $\obj$ is bounded below. Also note that if $\ma{\Phi}$ has full column rank, then, using condition \refeq{eq:sufficent_decrease}, one can show that the sequence $(\sig_k)_{k \in \Nbb}$ is bounded. For the fidelity term, its semi-algebraicity depends on the properties of the generating function $\kernel$ and the type of geometric transformations. In the next section, we present examples of functions $\kernel$ and geometric transformations such that this term is semi-algebraic and such that the entries of the matrix $\InterpMa$ are twice continuously differentiable, as required by Theorem \ref{th:main}.

\subsection{Examples of admissible geometric transformations}
Let us start by studying the simple case of translations. This type of transformation can be represented by $2$ parameters $\param = (\theta_1, \theta_2)^\adjoint \in \Rbb^2$ and a function $\trans_{\param}$ defined as
\begin{align*}
\tau_{\param}(\pos) = (u_1 + \theta_1,\ u_2 + \theta_2)^\adjoint.
\end{align*}
Combined, for example, with the cubic interpolator defined in \cite{keys81} as generating function $\kernel(\pos) = \phi(u_1) \phi(u_2)$ where
\begin{align}
\label{eq:generating_function}
\phi(u) = 
\left\{
\begin{array}{ll}
\frac{3}{2} \abs{u}^3 - \frac{5}{2} \abs{u}^2 + 1, & \text{ if } 0 \leq \abs{u} < 1,\\
-\inv{2} \abs{u}^3 +  \frac{5}{2} \abs{u}^2 - 4 \abs{u} + 2, & \text{ if } 1 \leq \abs{u} < 2,\\
0, & \text{ otherwise},
\end{array}
\right. 
\end{align}
the entries of the matrix $\InterpMa$ are twice continuously differentiable with respect to the transformation parameters, and the function $(\param, \sig) \mapsto \norm{\SenMa(\param) \sig - \meas}_2^2$ is semi-algebraic. Theorem \ref{th:main} thus applies in this situation. Remark that with this choice of generating function the matrix $\InterpMa$ is sparse, which permits us to transform rapidly the background image. Note that, instead of the cubic interpolator, we could also have chosen any B-spline interpolators of degree larger than $3$ \cite{unser00}.

Then, with the same choice of generating function, we can also handle the case of affine transformations. Indeed, such transformations can be represented by $6$ parameters  $\param = (\theta_1, \ldots, \theta_6)^\adjoint \in \Rbb^6$ with 
\begin{align}
\label{eq:affine}
\tau_{\param}(\pos) = (\theta_1 u_1 + \theta_2 u_2 + \theta_3,\ \theta_4 u_1 + \theta_5 u_2 + \theta_6)^\adjoint,
\end{align}
and the requirements to apply Theorem \ref{th:main} are also met.

Next, let us show how to handle the case of homographies. An homography is usually represented using $8$ parameters  $\param = (\param_1, \ldots \param_8)^\adjoint \in \Rbb^8$ with $\trans_{\param}$ satisfying
\begin{align*}
\tau_{\param}(\pos) = 
\left(
\frac{\theta_1 u_1 + \theta_2 u_2 + \theta_3}{\theta_7 u_1 + \theta_8 u_2 + 1},\
\frac{\theta_4 u_1 + \theta_5 u_2 + \theta_6}{\theta_7 u_1 + \theta_8 u_2 + 1}
\right)^\adjoint.
\end{align*}
Unfortunately, the function $(\param, \sig) \mapsto \norm{\SenMa(\param) \sig - \meas}_2^2$ is not semi-algebraic in this case. However, if $\abs{\theta_7 u_1 + \theta_8 u_2} \ll 1$, the transformation function can be approximated by
\begin{align}
\label{eq:homography}
\tau_{\param}(\pos) = 
\left(
\begin{array}{l}
\theta_1 u_1 + \theta_2 u_2 + \theta_3\\
\theta_4 u_1 + \theta_5 u_2 + \theta_6
\end{array}
\right) (1 - \theta_7 u_1 - \theta_8 u_2),
\end{align}
using a first order Taylor expansion. With this approximation, all the conditions are now fulfilled to apply Theorem \ref{th:main}. Note that the condition $\abs{\theta_7 \pos_1 + \theta_8 \pos_2} \ll 1$ can be enforced during the reconstruction by using appropriate bounds $\paramUp$ and $\paramLow$. 

{\color{black} Beyond these examples, let us mention that if the generating function $\kernel$ and the transformation function $\tau_{\param}$ are twice continuously differentiable, Algorithm $1$ can also be applied and the first two points of Theorem \ref{th:main} still hold.}

\subsection{Influence of the algorithm parameters}
\label{sec:discussion}
Let us highlight that, even though the sequence $(\sig^k, \param^k)_{k \in \Nbb}$ generated by Algorithm $1$ converges to a critical point of $\obj$ for all strictly positive parameter $\regMove_{\param}$ and strictly positive bounded sequence $(\regMove_{\sig}^k)_{k \in \Nbb}$, different values of these parameters yield different results. The choice of these parameters is thus important. 

{\color{black}
For the sequence $(\regMove_{\sig}^k)_{k \in \Nbb}$, we start from large values to constrain $\sig^{k+1}$ to stay ``close'' to $\sig^k$ when the estimated images are still inaccurate. Starting with large values is also essential to obtain images $\sig^k$ made of coarse wavelets only during the first iterations. As a rule of thumb, we have noticed that the best reconstructions were obtained when $\regMove_{\sig}^0$ is chosen such that $\norm{\SenMa(\param^0)\,\sig^1 - \meas}_2^2/\norm{\meas}_2^2 \in [0.1, 0.2]$, after the first step of the first iteration. Then, we slightly decrease the value of $\regMove_{\sig}^k$ at each iteration with the recursion rule $\regMove_{\sig}^{k+1} = \max(0.9 \regMove_{\sig}^k\, , 0.1)$. The parameter $\regMove_{\param}$ applying on the transformation parameters seems to have less influence on the final reconstruction quality, and we keep it fixed at $0.1$ in all the experiments of the next section.

Finally, concerning the bounds $\paramLow$ and $\paramUp$ constraining the transformation parameters, we simply choose them large enough to ensure that the actual transformation parameters between images are in the allowed range, while making sure that uninteresting solutions, such as translations larger than the size of the images or scaling parameters tending to zero, are prevented.
}

\subsection{Limitations of our algorithm}
{\color{black} We now conclude this section by discussing some limitations of our algorithm.

In this work, we considered geometric transformations globally described by \emph{few} parameters. This restricts the number of applications where the algorithm can be applied. For example, when the observed scene is not planar, the geometric transformations between images are usually more complex and depend on the structure of the scene. In this situation, more general transformation models have to be used. One should note however that the following two properties of the proposed method may help to handle such a situation. First, the requirements of Theorem \ref{th:main} hold for a large class of transformation models $\trans_{\param}$. We thus can use transformation models with more degrees of freedom to approximate more complex transformations. Second, the presence of foreground images in the measurement model can render our method robust to small transformation model mismatches by identifying the parts of the scene that cannot be aligned with the assumed model as occlusions. 

Finally, another limitation concerns the number of transformation parameters that the algorithm can efficiently handle. Indeed, if the number $\nbParam$ of transformation parameters is large, the minimization problem at the second step of Algorithm $1$ might become intractable. First, the computation of the matrix $\Hess \in \Rbb^{\nbParam \times \nbParam}$, defined in \refeq{eq:hessian}, will become time consuming (unless the transformation model used exhibits special structure). Second, solving the minimization problem \refeq{eq:step2} is computationally expensive for unstructured large matrix $\Hess$. Note however that one way to circumvent this issue is to approximate the matrix $\Hess$ by computing only its most ``significant'' entries and setting the remaining ones to zero.
}

\section{Experiments}
\label{sec:results}
{\color{black}
In this section, we demonstrate numerically the efficiency of the algorithm in three different applications: robust image alignment, compressed sensing, and super-resolution. The code and data needed to reproduce the results presented in this section are available at \url{lts2www.epfl.ch/people/gilles/softwares}. }

\subsection{Robust image alignment}

\begin{figure}
\centering
\centering \scriptsize
\begin{sideways} \hspace{5mm} \scriptsize Input images \end{sideways}
\includegraphics[width=15.15cm, keepaspectratio]{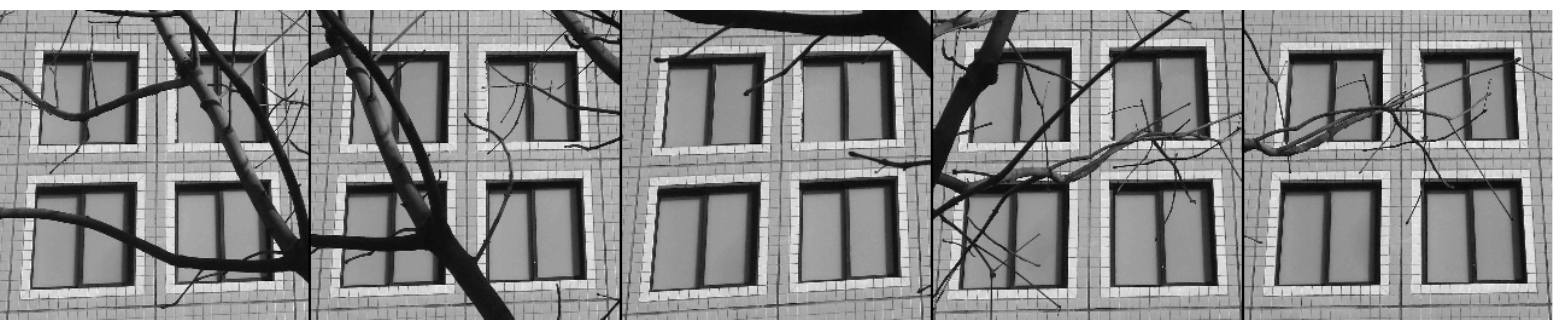}\\
\hspace{+0.4cm} $\nbObs = 6$ \hspace{2.1cm} $\nbObs = 8$
\hspace{2.1cm} $\nbObs = 10$ \hspace{2.1cm} $\nbObs = 12$ \hspace{2.1cm} $\nbObs = 16$\\
\begin{sideways} \hspace{.3cm} Proposed method \end{sideways}
\includegraphics[width=15.15cm, keepaspectratio]{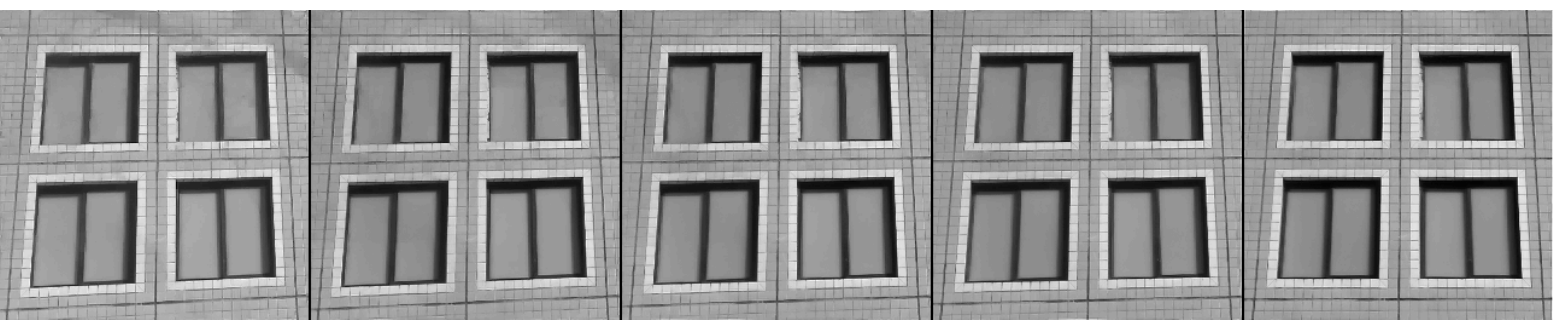}\\
\begin{sideways} \hspace{1.1cm} \scriptsize RASL \end{sideways}
\includegraphics[width=15.15cm, keepaspectratio]{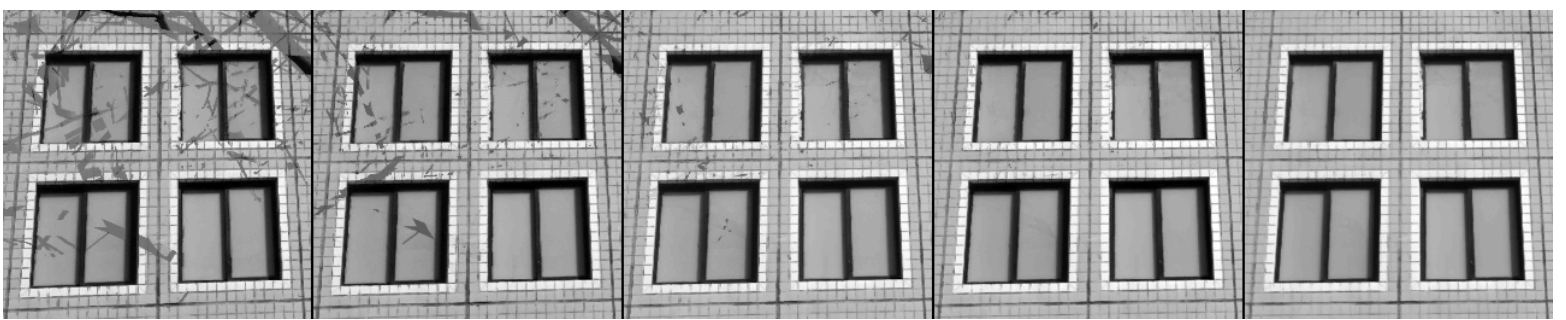}
\caption{\label{fig:alignment1} Top panel: first $5$ input images. Bottom panels: estimated background images with the proposed method and RASL using $\nbObs = 6, 8, 10, 12$ or $16$ input images (from left to right).}
\end{figure}
%

\begin{figure}
\centering 
\scriptsize \hspace{.3cm} $\nbObs = 8$ \hspace{6.9cm} $\nbObs = 16$\\
\begin{sideways} \hspace{.7cm} Proposed method \end{sideways}
\includegraphics[width=.48\linewidth, keepaspectratio]{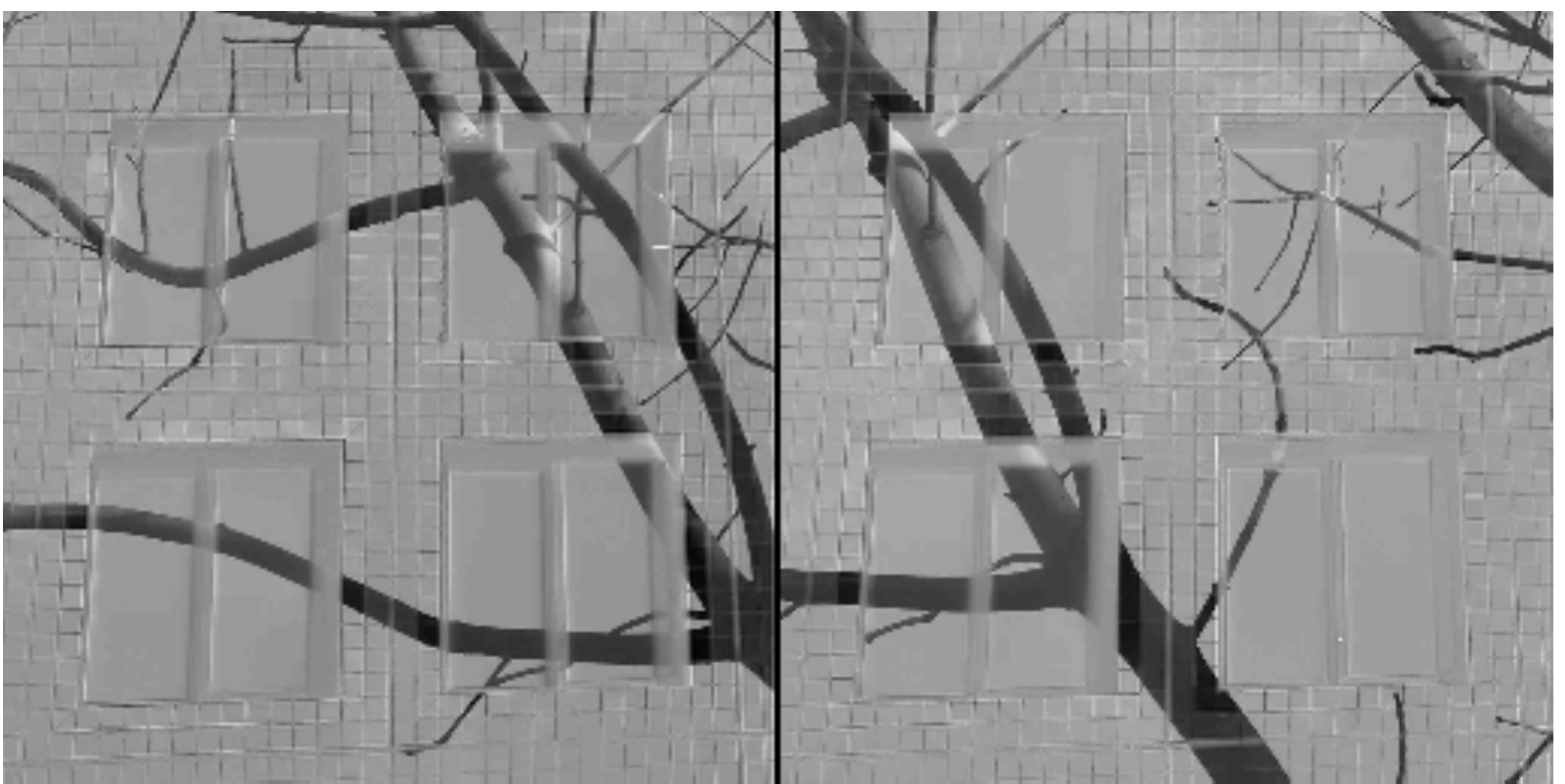}
\hfill
\includegraphics[width=.48\linewidth, keepaspectratio]{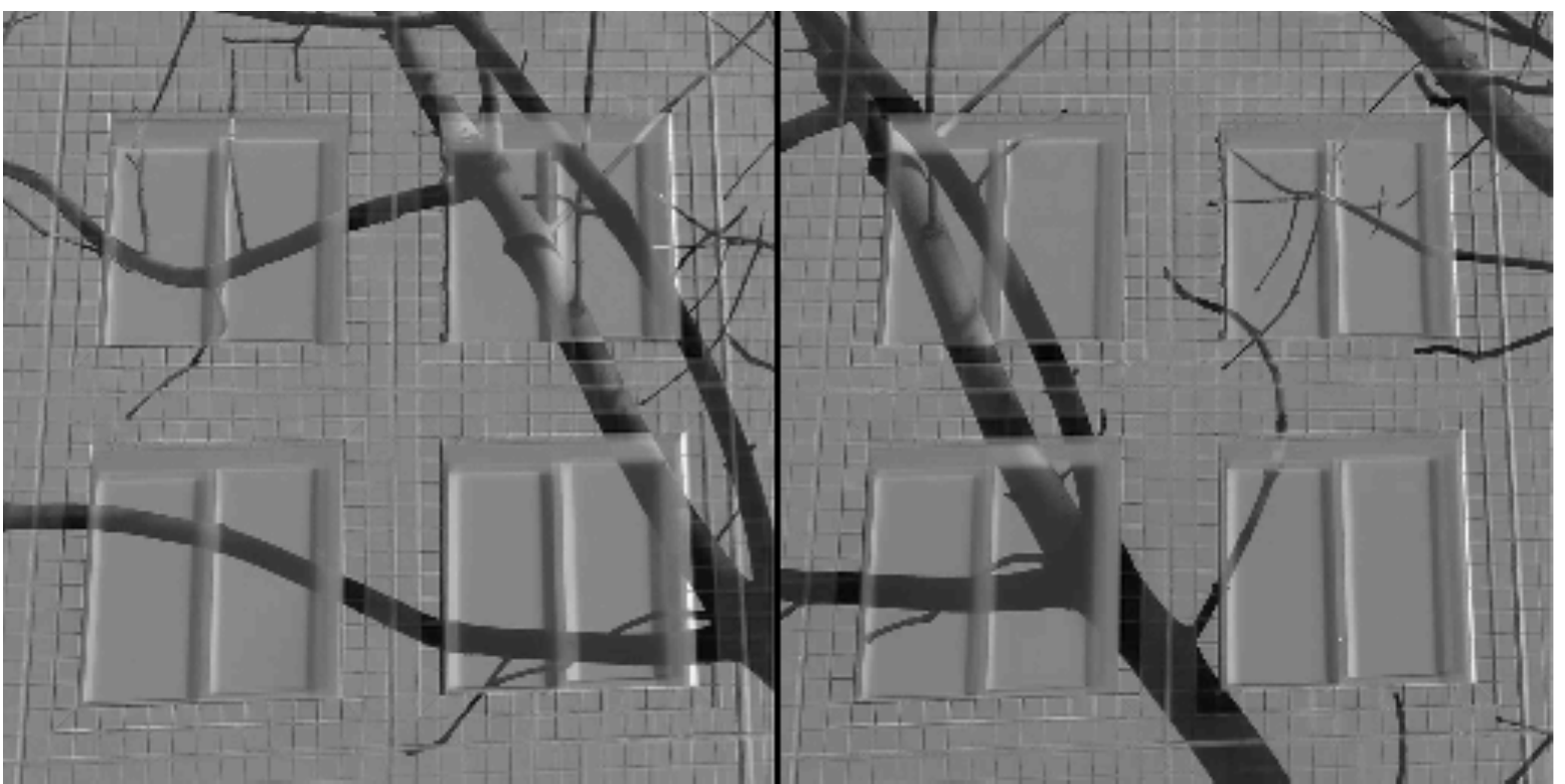}\\
\begin{sideways} \hspace{1.5cm} \scriptsize RASL \end{sideways} \hspace{-0.4mm}
\includegraphics[width=.48\linewidth, keepaspectratio]{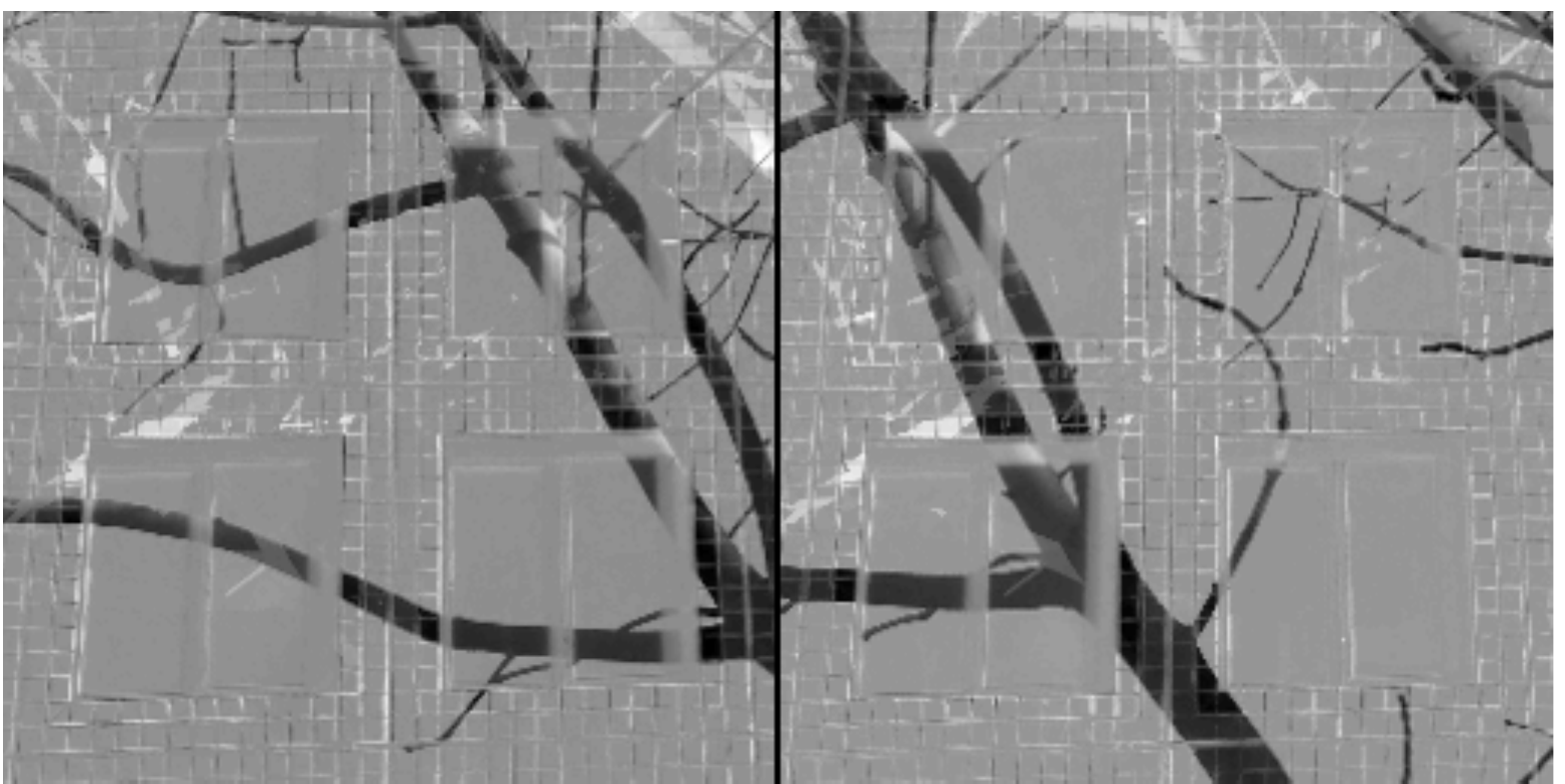}
\hfill
\includegraphics[width=.48\linewidth, keepaspectratio]{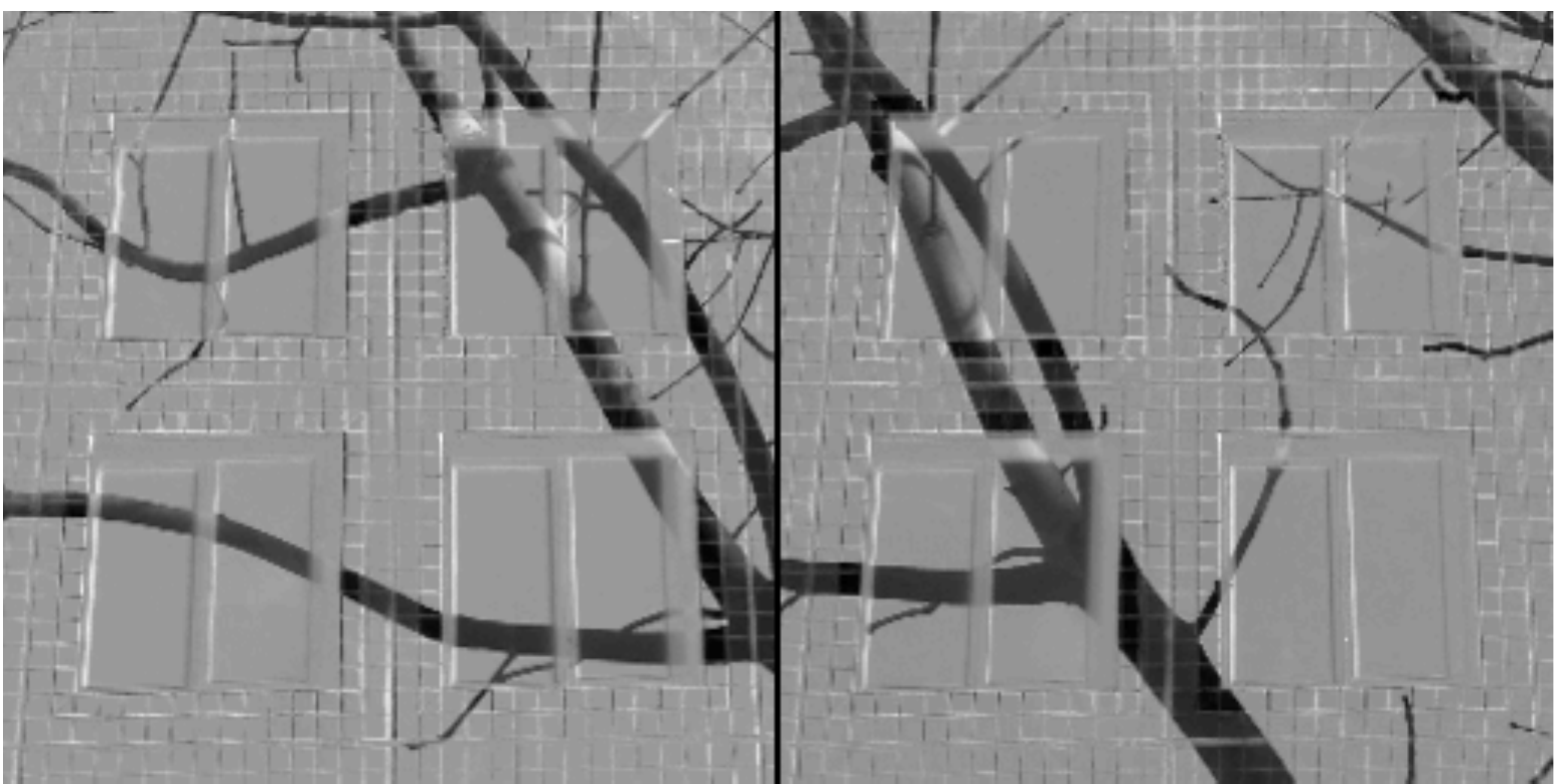}
\caption{\label{fig:alignment2} First $2$ estimated foreground images with the proposed method and RASL using $8$ (left) or $16$ (right) input images.}
\end{figure}
%

\begin{figure}
\centering
\begin{minipage}{0.325\linewidth} \centering \scriptsize Not registered \\
\includegraphics[height=0.85\linewidth, keepaspectratio]{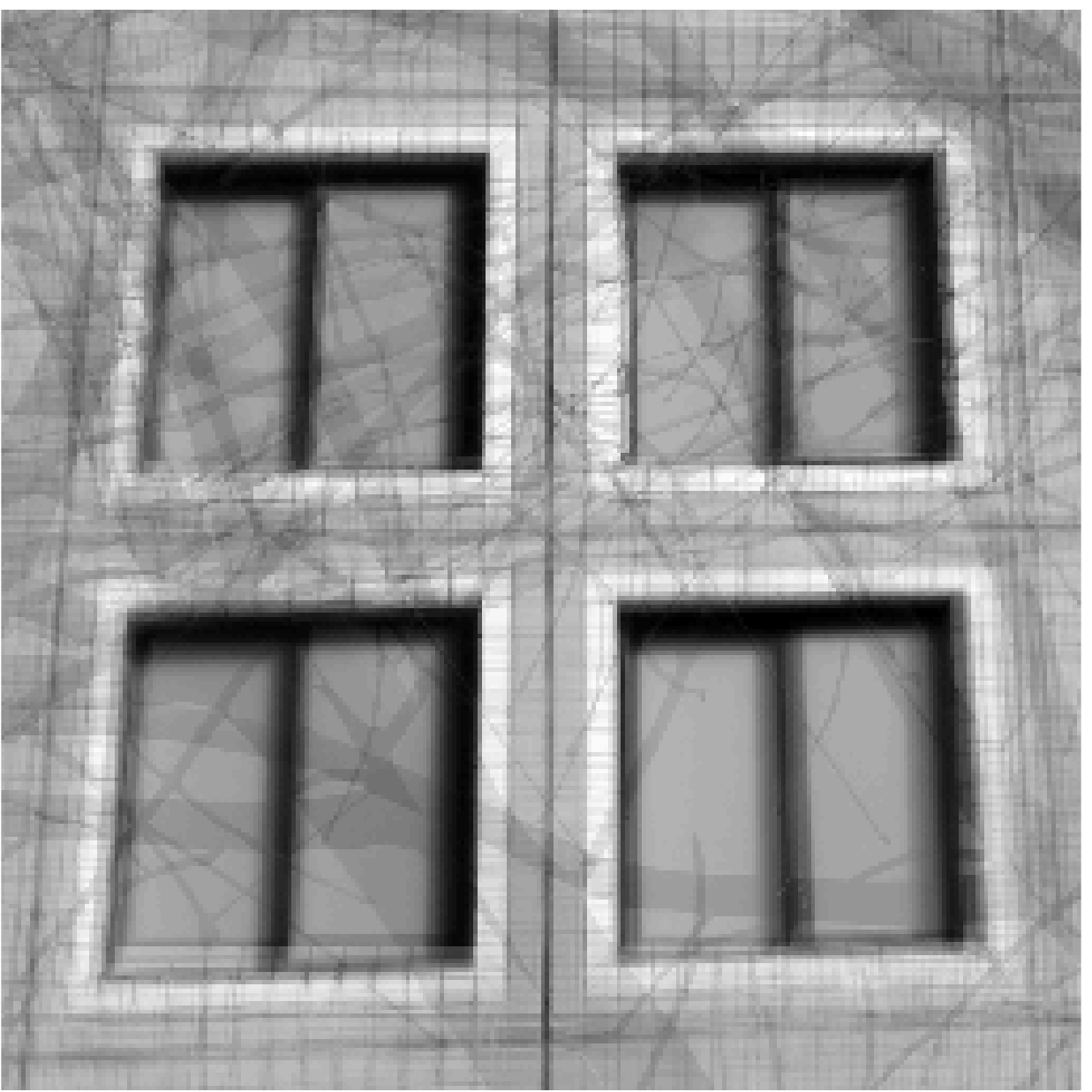}
\end{minipage}
\begin{minipage}{0.325\linewidth} \centering \scriptsize Registered - Proposed method\\
\includegraphics[height=0.85\linewidth, keepaspectratio]{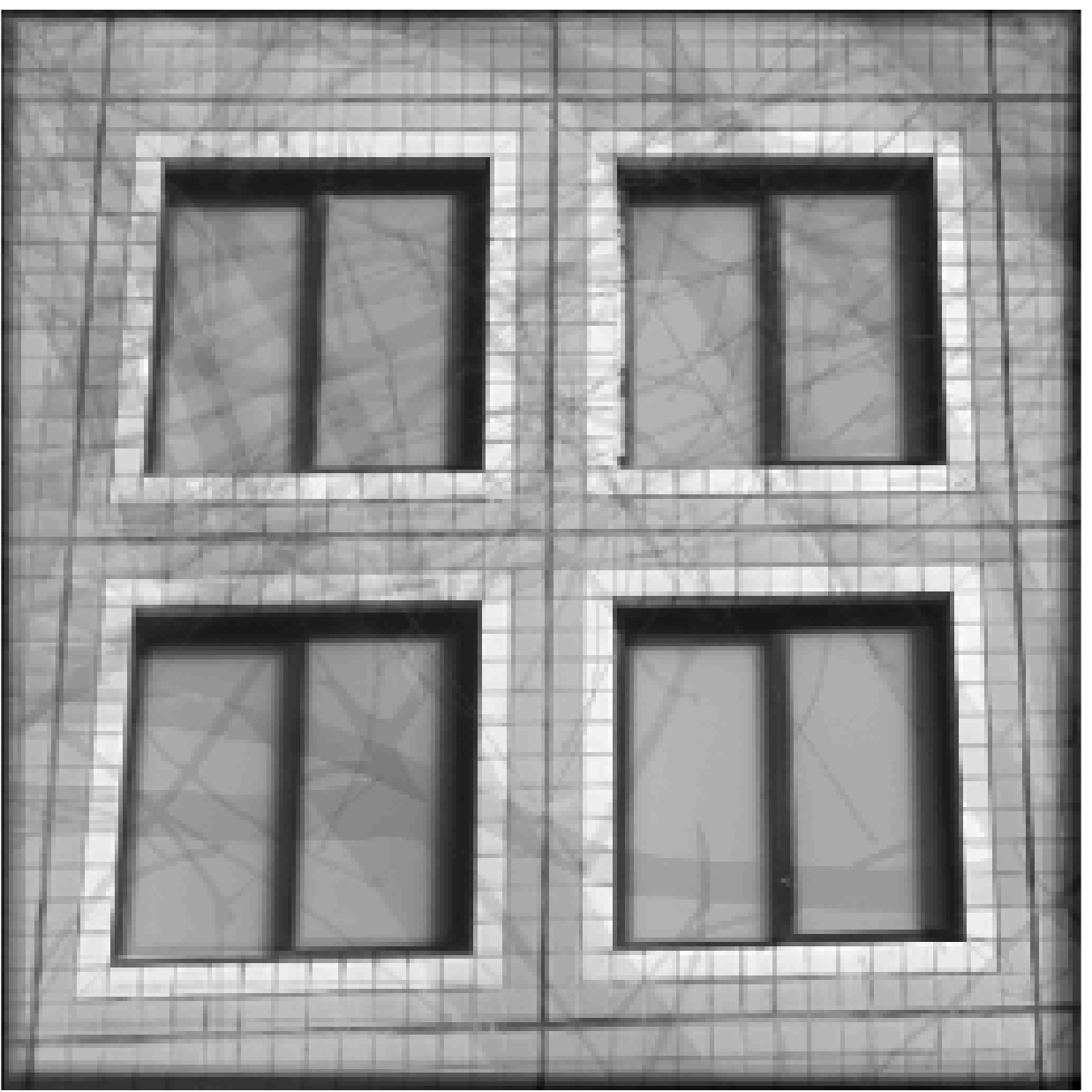}
\end{minipage}
\begin{minipage}{0.325\linewidth} \centering \scriptsize Registered - RASL\\ 
\includegraphics[height=0.85\linewidth, keepaspectratio]{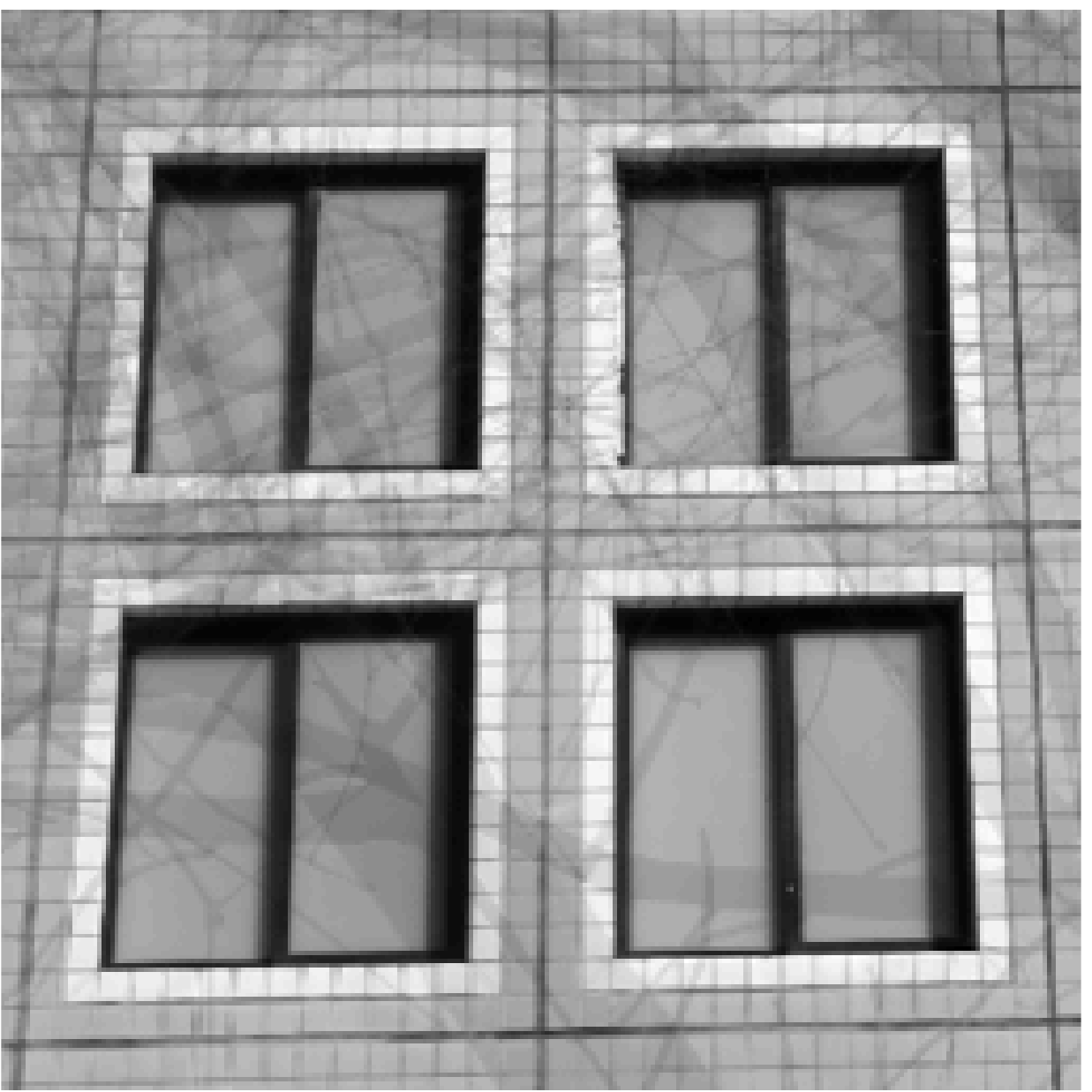}
\end{minipage}
\caption{\label{fig:alignment3} Left panel: superposition of the first $10$ input images. Middle panel: superposition of the same $10$ images after registration with our method. Right panel: superposition of the same $10$ images after registration with RASL.}
\end{figure}
%

\begin{figure}
\centering
\begin{minipage}{0.243\linewidth} \centering \scriptsize $\sig_0^1$ \\
\includegraphics[height=\linewidth, keepaspectratio]{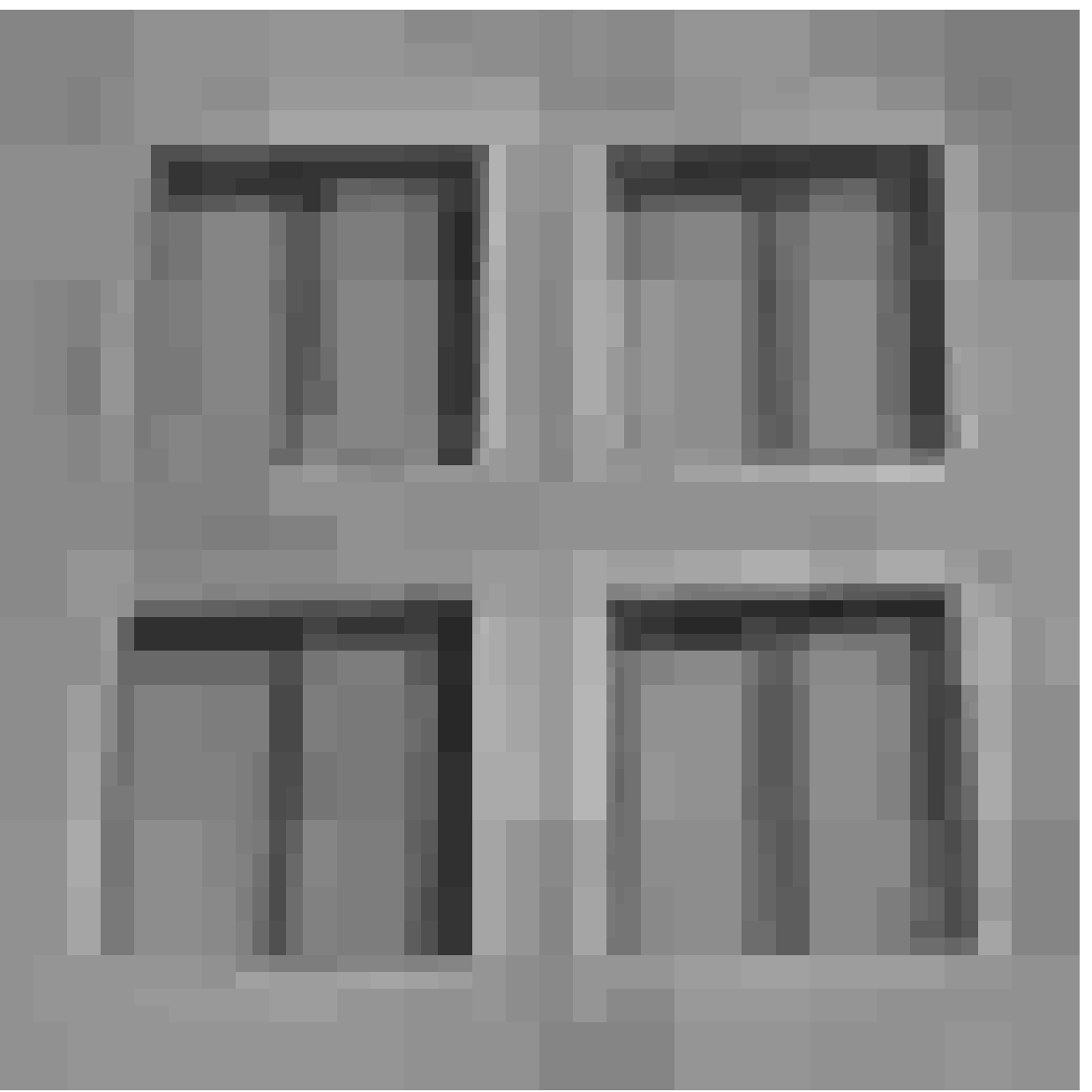}
\end{minipage}
\begin{minipage}{0.243\linewidth} \centering \scriptsize $\sig_0^{11}$ \\
\includegraphics[height=\linewidth, keepaspectratio]{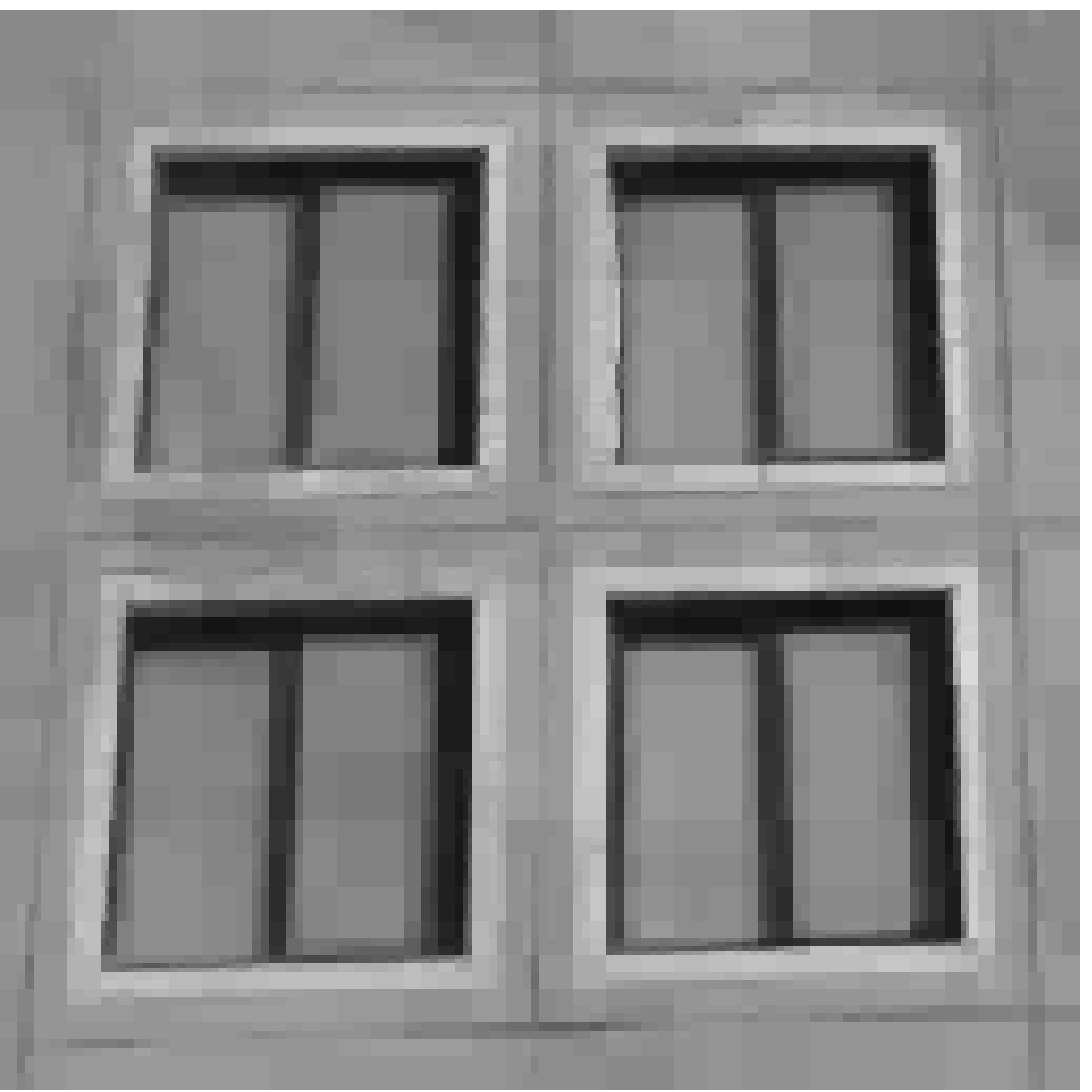}
\end{minipage}
\begin{minipage}{0.243\linewidth} \centering \scriptsize $\sig_0^{21}$ \\
\includegraphics[height=\linewidth, keepaspectratio]{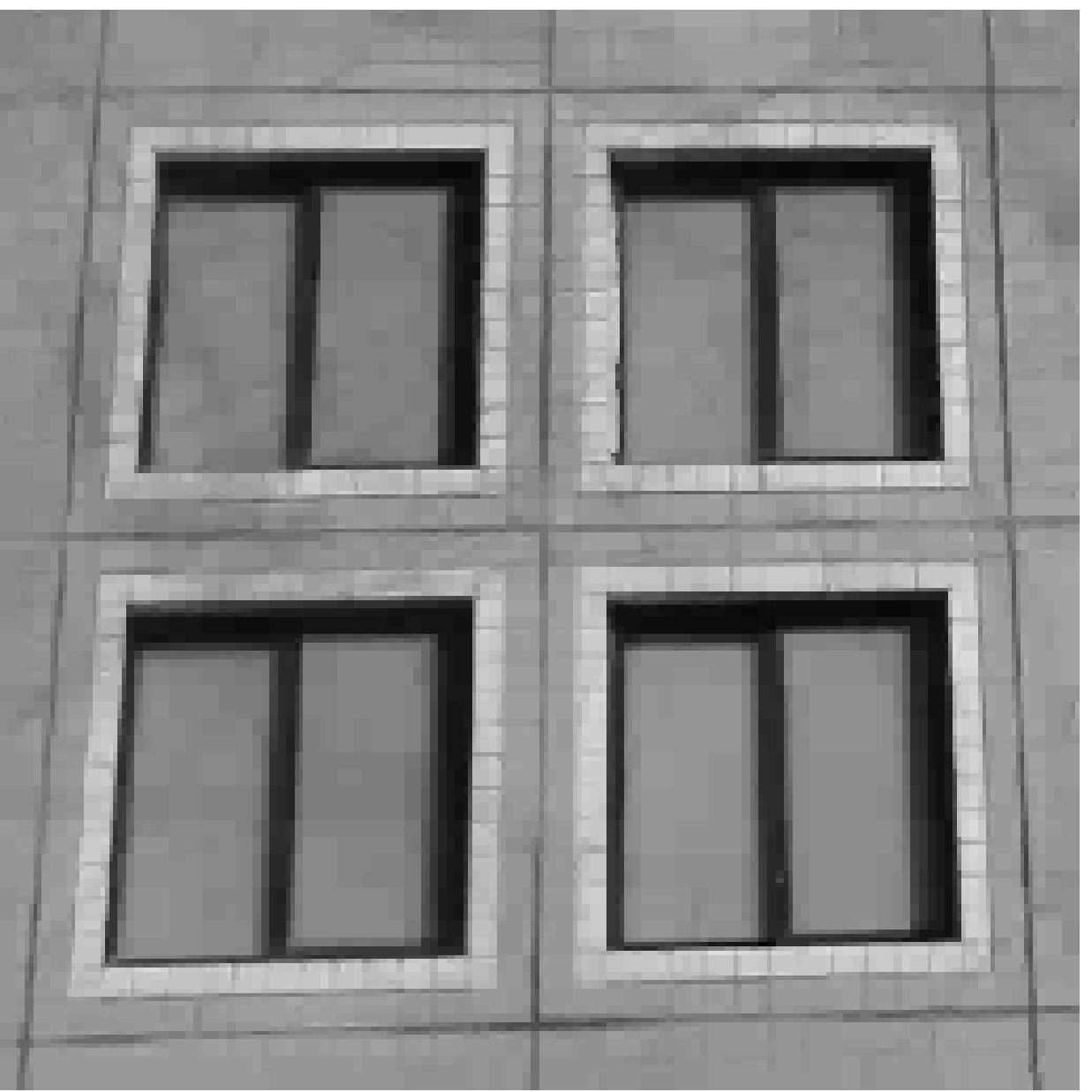}
\end{minipage}
\begin{minipage}{0.243\linewidth} \centering \scriptsize $\sig_0^*$ \\
\includegraphics[height=\linewidth, keepaspectratio]{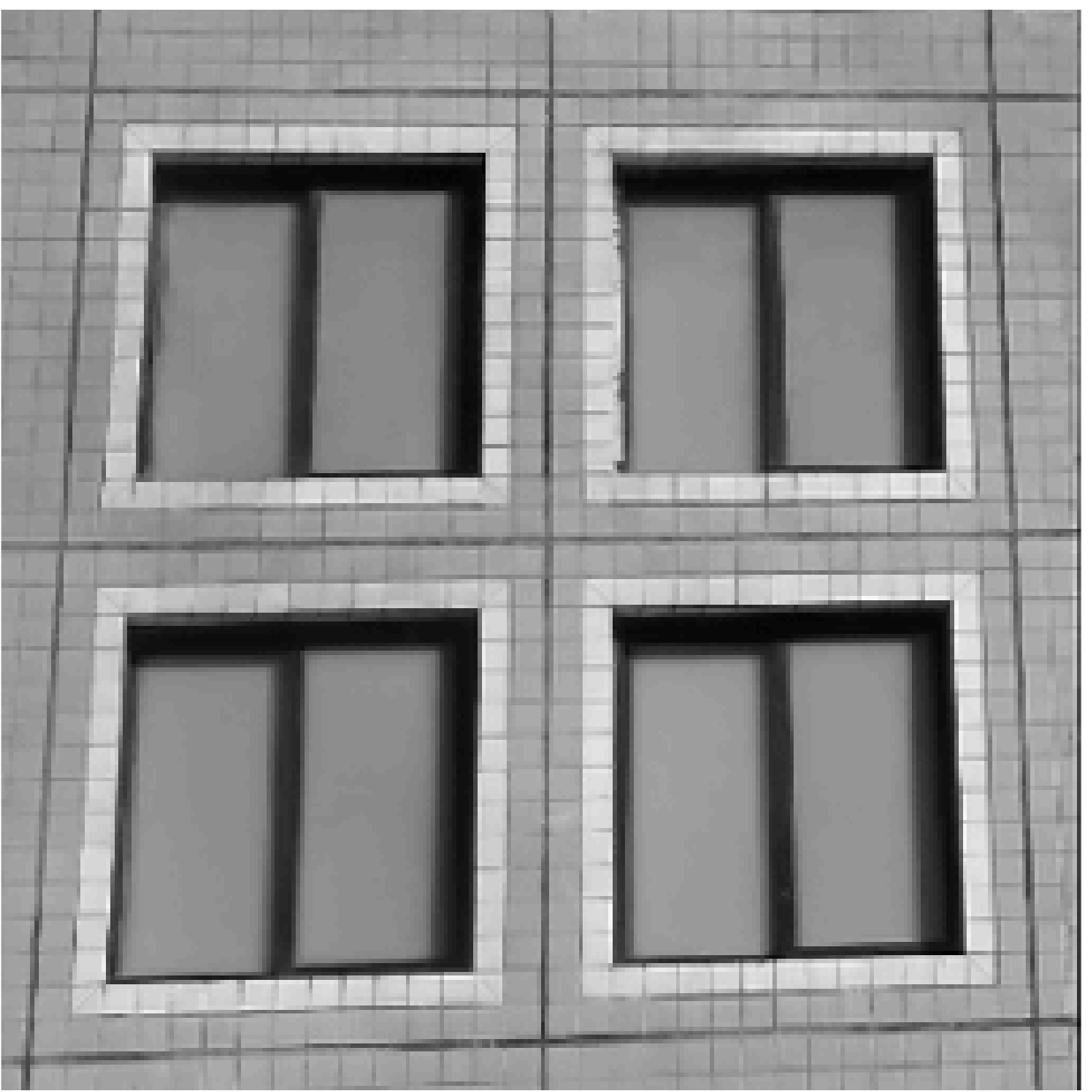}
\end{minipage}
\caption{\label{fig:alignment4} Sequence of estimated background images $\sig_0^{k+1}$ at $k=0$ (first from the left), $k=10$ (second), $k=20$ (third), and at convergence (fourth).}
\end{figure}

{\color{black} Robust image alignment consists in aligning several images of a given scene despite the presence of large occlusions, as, for example, in the images presented in the top panel\footnote{Dataset available at \url{perception.csl.illinois.edu/matrix-rank/rasl.html}.} of Fig.~\ref{fig:alignment1}. 

We propose here to solve this problem by using the measurement model defined in \refeq{eq:measurement_model} and Algorithm $1$ to minimize \refeq{eq:objective_function}. In the measurement model, the matrices $\SenMa_1, \ldots, \SenMa_\nbObs$, are set to the identity $\ma{I}_{\dimSig} \in \Rbb^{\dimSig \times \dimSig}$, and $\meas_1, \ldots, \meas_\nbObs$, contain the occluded observed images. Ideally our algorithm should estimate the background image $\sig_0$, i.e., the wall and the windows in Fig.~\ref{fig:alignment1} (top panel), extract the occlusions $\sig_1, \ldots, \sig_\nbObs$, i.e., the branches, and provide the transformation parameters $\param_1, \ldots, \param_\nbObs$, which align the background image on the observed images.

To show the efficiency of our technique, we compare our results with the ones obtained using RASL \cite{peng10, peng12}. This method is based on low rank and sparse approximations. More precisely, the authors in \cite{peng10, peng12} concatenate the occluded input images in a matrix $\ma{Y} = [\meas_1, \ldots, \meas_{\nbObs}] \in \Rbb^{\dimMeas \times \nbObs}$ and propose to solve
\begin{align}
\label{eq:RASL}
(\tilde{\ma{B}}, \tilde{\ma{F}}, \tilde{\param}) \in \argmin_{\ma{B}, \ma{F}, \param} \norm{\ma{B}}_{*} + \frac{1}{\sqrt{m}} \norm{\ma{F}}_{1, 1}\quad \text{ subject to }\quad \ma{Y} \circ \tau_{\param} = \ma{B} + \ma{F},
\end{align}
where $\ma{B} \in \Rbb^{\dimMeas \times \nbObs}$, $\ma{F} \in \Rbb^{\dimMeas \times \nbObs}$, $\norm{\ma{B}}_{*}$ denotes the nuclear norm of $\ma{B}$, $\norm{\ma{F}}_{1, 1}$ is the sum of the absolute value of the entries of $\ma{F}$, and $\ma{Y} \circ \tau_{\param}$ are the input images transformed according to the parameters $\param$ and the transformation model $\tau_{\param}$. This optimization problem decomposes the initial images $\ma{Y}$ in a low-rank matrix $\ma{B}$ which contains the background elements, and a sparse matrix $\ma{F}$ which contains the occlusions. The transformation parameters $\param$ are adjusted so that the matrix $\ma{F}$ is as sparse as possible and the rank of the matrix $\ma{B}$ is as small as possible.

One can remark that the data constraint $\ma{Y} \circ \tau_{\param} = \ma{B} + \ma{F}$ is non-linear in $\param$ and minimizing \refeq{eq:RASL} is thus not trivial. To get around this difficulty, the authors linearize the term $\ma{Y} \circ \tau_{\param}$ with respect to the transformation parameters. This approximation is only valid for small transformations but the optimization problem becomes convex and easier to solve. Then, to align images with large transformations between them, the authors propose to repeatedly solve this convex problem and linearize the data constraint around the new estimated parameters to improve, little by little, the registration. Note that the authors do not provide a rigorous proof of convergence of this procedure. 

Finally, it is interesting to note that, surprisingly, no constraints are imposed on the transformation parameters in \refeq{eq:RASL}. Therefore, if there exist a set of parameters $\param_0 \in \Rbb^{\nbObs\nbParam}$ such that $\ma{Y} \circ \tau_{\param_0} = \ma{0}$, then it is easy to prove that the global minimum of \refeq{eq:RASL} is $0$ and $\tilde{\ma{B}} = \tilde{\ma{F}} = \ma{0}$. In such a situation, finding a global minimizer of \refeq{eq:RASL} obviously does not yield to the desired solution, and, if one correctly aligns the images, it is only because the procedure described in \cite{peng10, peng12} converges to a local minima of problem \refeq{eq:RASL}. Note that as soon as the class of transformations considered allows us to scale the images, we already have $\ma{Y} \circ \tau_{\param} \rightarrow 0$ when the scaling parameter tends to zero. Let us however highlight that this issue can trivially be corrected by adding a constraint on the transformation parameters in the original formulation \refeq{eq:RASL} and modify accordingly the procedure described in \refeq{eq:RASL}.

\subsubsection{Background/foreground separation quality}

In this first set of experiments, we use $\nbObs = 6, 8, 10, 12$, and $16$ occluded images of $\dimSig = 256 \times 256$ pixels as inputs. The first $5$ images are presented in the top panels of Fig.~\ref{fig:alignment1}. Then, for each $l$, we run Algorithm $1$ and RASL to align the images and to estimate the background and foreground images. 

In our algorithm, we set $\reg=100$, $\regMove_{\param} = 0.1$, $\regMove_{\sig}^k = \max(0.9^k\,(20\reg), 0.1)$, and $\mu = 10^{-10}$. The matrix $\Dict$ is built using the Haar wavelet basis, and $\prior(\sig)$ is set to $\norm{\Dict^\adjoint \sig}_1$. We also assume that the transformations between images are well modeled by homographies and thus use model \refeq{eq:homography} for $\trans_{\param}$. The parameters are constrained as follows: $\abs{1-\theta_1}, \abs{1-\theta_5} \leq 0.2$, $\abs{\theta_2}, \abs{\theta_4} \leq 0.2$, $\abs{\theta_3}, \abs{\theta_6} \leq 20$, $\abs{\theta_7}, \abs{\theta_8} \leq 0.01$, with $u_1$ and $u_2$ belonging to $\{-127, -126, \ldots, 128\}$.

In Fig.~\ref{fig:alignment1}, we present for each $\nbObs$ the background images\footnote{For RASL, the background image is obtained by averaging the estimated background images in the matrix $\tilde{\ma{B}}$.} estimated with our method and RASL. One can first remark that with a small number of input images, i.e., $\nbObs=6, 8$ and $10$, the background images estimated with our method contains much less artifacts due to the occlusions than the ones estimated with RASL. Finally, our method only needs $10$ input images to obtain a background image free of occlusion artifacts while RASL needs more input images.

In Fig.~\ref{fig:alignment2}, we present the first two foreground images estimated with both methods from $\nbObs = 8$ and $16$ input images. One can notice that, for both methods, the foreground images essentially contain the occlusions, i.e., the branches, present in the initial input images. However, at $\nbObs = 8$, the foreground images estimated with RASL are contaminated by other objects not initially present in the corresponding input images. With a small number of input images, our method separates the background from the foreground better than RASL does. At $\nbObs = 16$, the foreground images estimated by both methods are quite similar. We note however that the windows are slightly less visible in the foreground images estimated by RASL.

To visualize the quality of the estimated transformation parameters, we present in Fig.~\ref{fig:alignment3} the superposition of $\nbObs = 10$ input images before and after registration with our method and RASL. One can easily remark that both methods are able to accurately register the set of input images.

Finally, we present in Fig.~\ref{fig:alignment4} the evolution of the estimated background image $\sig_0^{k+1}$ at iterations $k=0, 10, 20, 30$, and at convergence with our method. This sequence of images illustrates the fact that, as explained in Section \ref{sec:first_step}, the images are reconstructed in a multiscale fashion.

\subsubsection{Registration quality} We now compare quantitatively the registration quality that one can obtain with RASL or our method. 

In this second set of experiments, we use a ``scaling+translational'' model represented by $3$ parameters  $\param = (s, t_x, t_y)^\adjoint \in \Rbb^3$ with 
\begin{align}
\label{eq:scaling+translation}
\tau_{\param}(\pos) = (s\, u_1 + t_x,\ s\, u_2 + t_y)^\adjoint,
\end{align}
where $s$ is a scaling factor, and $(t_x, t_y)$ are translation parameters in the first and second dimensions of the images, respectively. To compare the registration performance, we run two different types of simulations on images of $\dimSig = 128 \times 128$ pixels. The spatial coordinates $u_1$ and $u_2$ belong to $\{-63, -62, \ldots, 64\}$. Let us mention that with this transformation model, all the requirements needed to apply Theorem \ref{th:main} are met.

For the first type of simulations, we draw independently and uniformly at random $\nbObs$ translation parameters, $\{ (t_x)_1, \ldots, (t_x)_\nbObs \}$ and $\{ (t_y)_1, \ldots, (t_y)_\nbObs \}$, in the intervals $[-\Delta t_x/2,\, \Delta t_x/2]$ and $[-\Delta t_y/2,\, \Delta t_y/2]$, respectively, while keeping $s$ fixed at $1$. Then, we transform a reference image\footnote{The image used is part of the castle-R20 dataset available at \url{cvlab.epfl.ch/~strecha/multiview/rawMVS.html} \cite{strecha08}.}  according to these parameters to create $\nbObs$ different input images. Finally, we register these images using RASL and our algorithm. This procedure is repeated for different sizes of the intervals, $\Delta t_x \in \{2, 8, 16, 24, 32\}$ and $\Delta t_y \in \{2, 8, 16, 24, 32, 40, 48\}$, and $30$ independent simulations are performed for each pair of intervals.

For the second type of simulations, we draw independently and uniformly at random $\nbObs$ translation parameters $\{ (t_x)_1, \ldots, (t_x)_\nbObs \}$ in the intervals $[-\Delta t_x/2,\, \Delta t_x/2]$ and $\nbObs$ scaling parameters $\{ s_1, \ldots, s_\nbObs \}$ in the interval $[1-\Delta s/2,\, 1+\Delta s/2]$, while keeping $t_y$ fixed at $0$. This procedure is repeated with $\Delta t_x \in \{2, 8, 16, 24, 32\}$ and $\Delta s \in \{0, 0.13, 0.25, 0.38, 0.5\}$, and $30$ independent simulations are also performed for each pair of intervals.

In our algorithm, we set $\reg=100$, $\regMove_{\param} = 0.1$, $\regMove_{\sig}^k = \max(0.9^k\,(50\reg), 0.1)$, and $\mu = 10^{-10}$. The matrix $\Dict$ is built using the Daubechies $8$ wavelet basis, and $\prior(\sig)$ is set to $\norm{\Dict^\adjoint \sig}_1$. The parameters are constrained as follows: $\abs{1 - s} \leq 0.5$, and $\abs{t_x}, \abs{t_y} \leq 64$.

To compare the registration quality of both methods, we compute the normalized root-mean-square-error $\sigma_{i \rightarrow j}$ between the ground truth parameters $\param_{i \rightarrow j} \in \Rbb^3$ that register the $i^\th$ image onto the $j^\th$ one and the corresponding parameters $\param_{i \rightarrow j}^* \in \Rbb^3$ estimated by either RASL or our method. These values are then averaged over all possible combinations of images to obtain the registration error
\begin{align*}
\sigma = \frac{1}{\nbObs(\nbObs-1)}\ \sum_{i = 1}^{\nbObs} \sum_{j \neq i} \sigma_{i \rightarrow j}\  \text{ with }\ \sigma_{i \rightarrow j} =  \frac{\norm{\param_{i \rightarrow j}^*-\param_{i \rightarrow j}}_2}{\norm{\param_{i \rightarrow j}}_2}.
\end{align*}

Fig.~\ref{fig:alignment6} presents the registration error $\sigma$ averaged over the $30$ simulations obtained with RASL and our method for each type of simulations with $\nbObs=10$. One can remark that for small transformations RASL is slightly more precise than our method. However, for large transformations, we obtain a smaller registration error. Our method is thus more robust to large displacements. Note that we also performed these simulations with $\nbObs = 5$ and the same conclusions hold.

\begin{figure*}
\includegraphics[width=\linewidth, keepaspectratio]{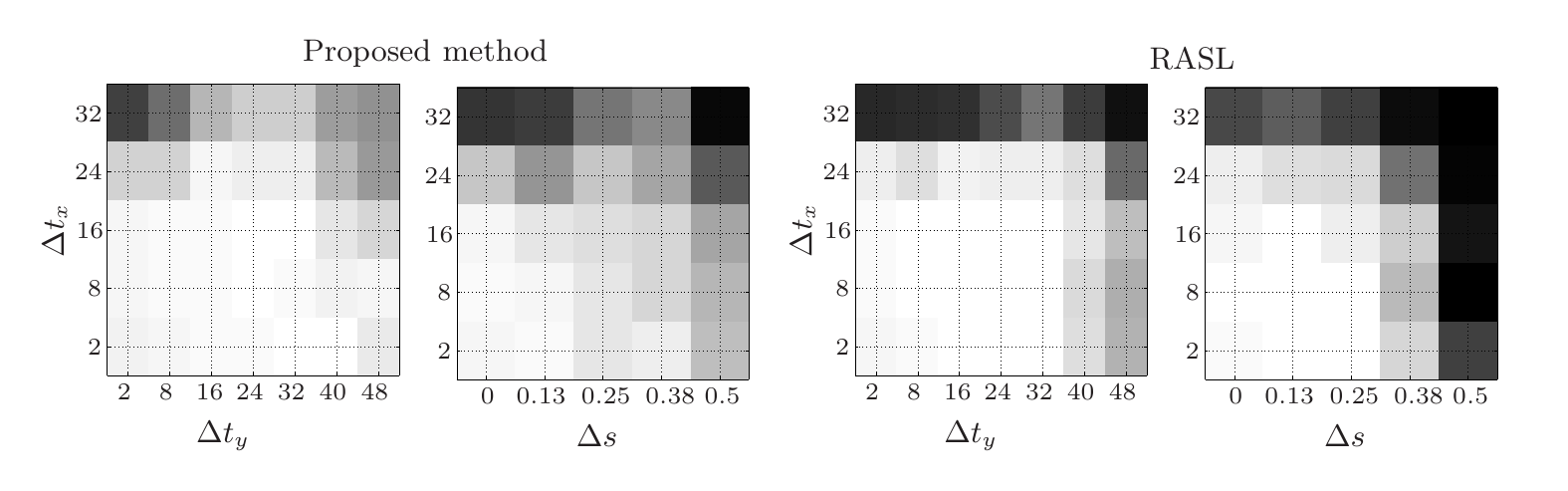}
\caption{\label{fig:alignment6} Registration error $\sigma$ averaged over $30$ simulations for the proposed method (left panels) and RASL (right panels) as a function of $(\Delta t_x, \Delta t_y)$ and of $(\Delta t_x, \Delta s)$. The colorbar goes from white for $\sigma = 0$ to black for $\sigma = 0.5$.}
\end{figure*}
}

\subsection{Compressed sensing}
\begin{figure*}
\centering
\begin{sideways} \hspace{5mm} \scriptsize Input images \end{sideways}
\includegraphics[width=15.15cm, keepaspectratio]{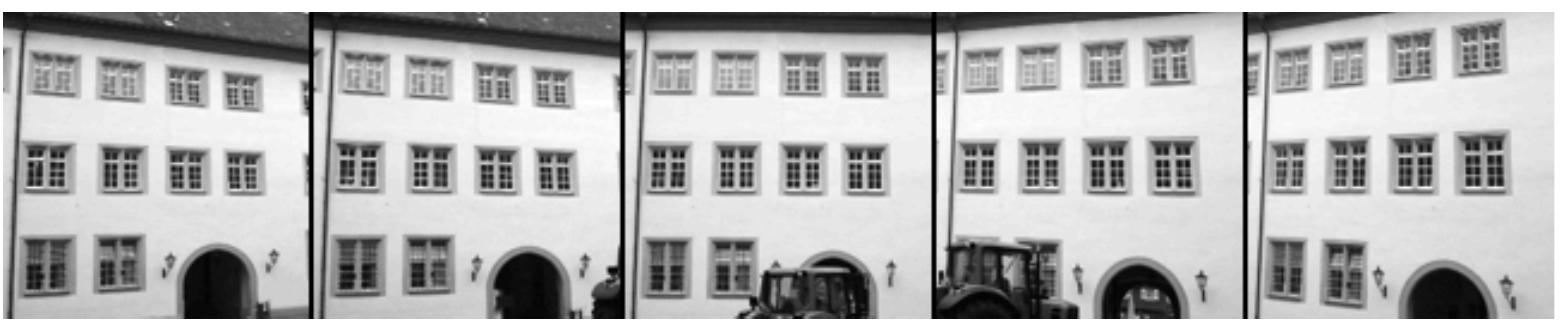}\\
\begin{sideways} \hspace{4mm} \scriptsize Method \refeq{eq:BPDN} \end{sideways}
\includegraphics[width=15.15cm, keepaspectratio]{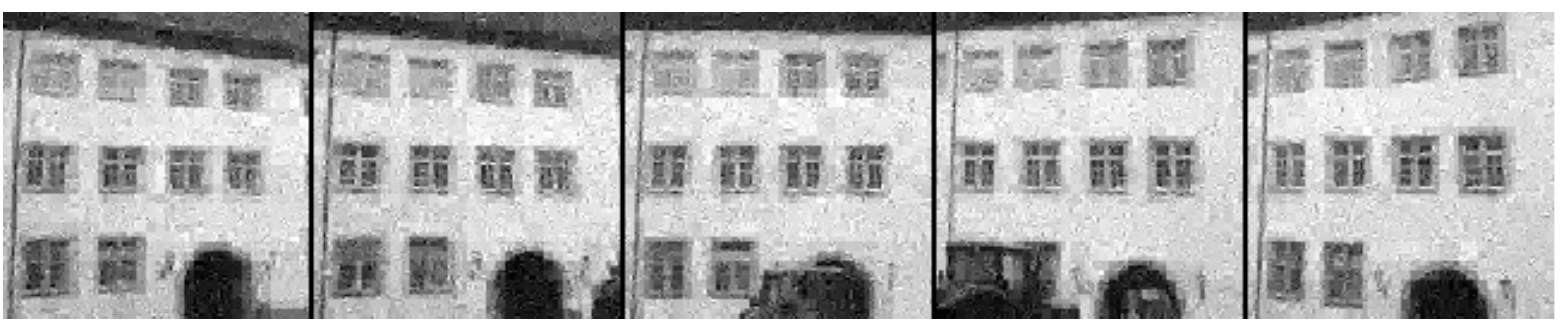}\\
\begin{sideways} \hspace{4mm} \scriptsize Method \refeq{eq:group_sparse} \end{sideways}
\includegraphics[width=15.15cm, keepaspectratio]{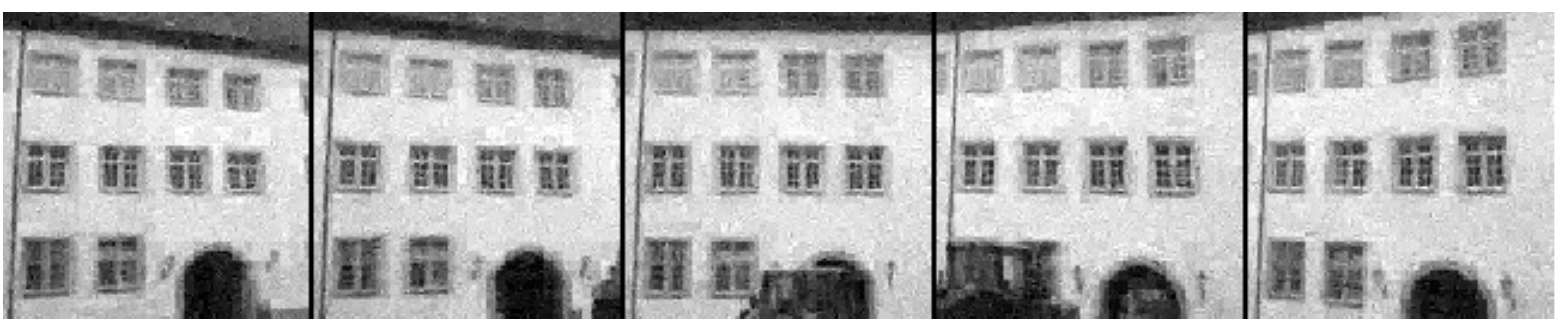}\\
\begin{sideways} \hspace{2mm} \scriptsize Proposed method \end{sideways}
\includegraphics[width=15.15cm, keepaspectratio]{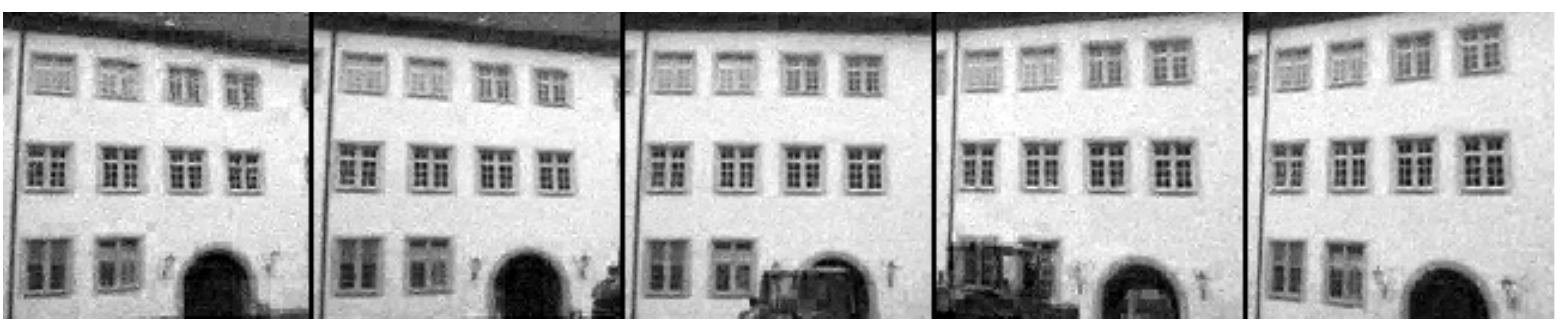}\\
\caption{\label{fig:CS1}Top panel: $5$ initial images. Middle-top panel: reconstructed images with method \refeq{eq:BPDN}. Middle-bottom panel: reconstructed images with method \refeq{eq:group_sparse}. Bottom panel: reconstructed images with the proposed method.}
\end{figure*}
\begin{figure}
\centering
\includegraphics[width=.7\linewidth, keepaspectratio]{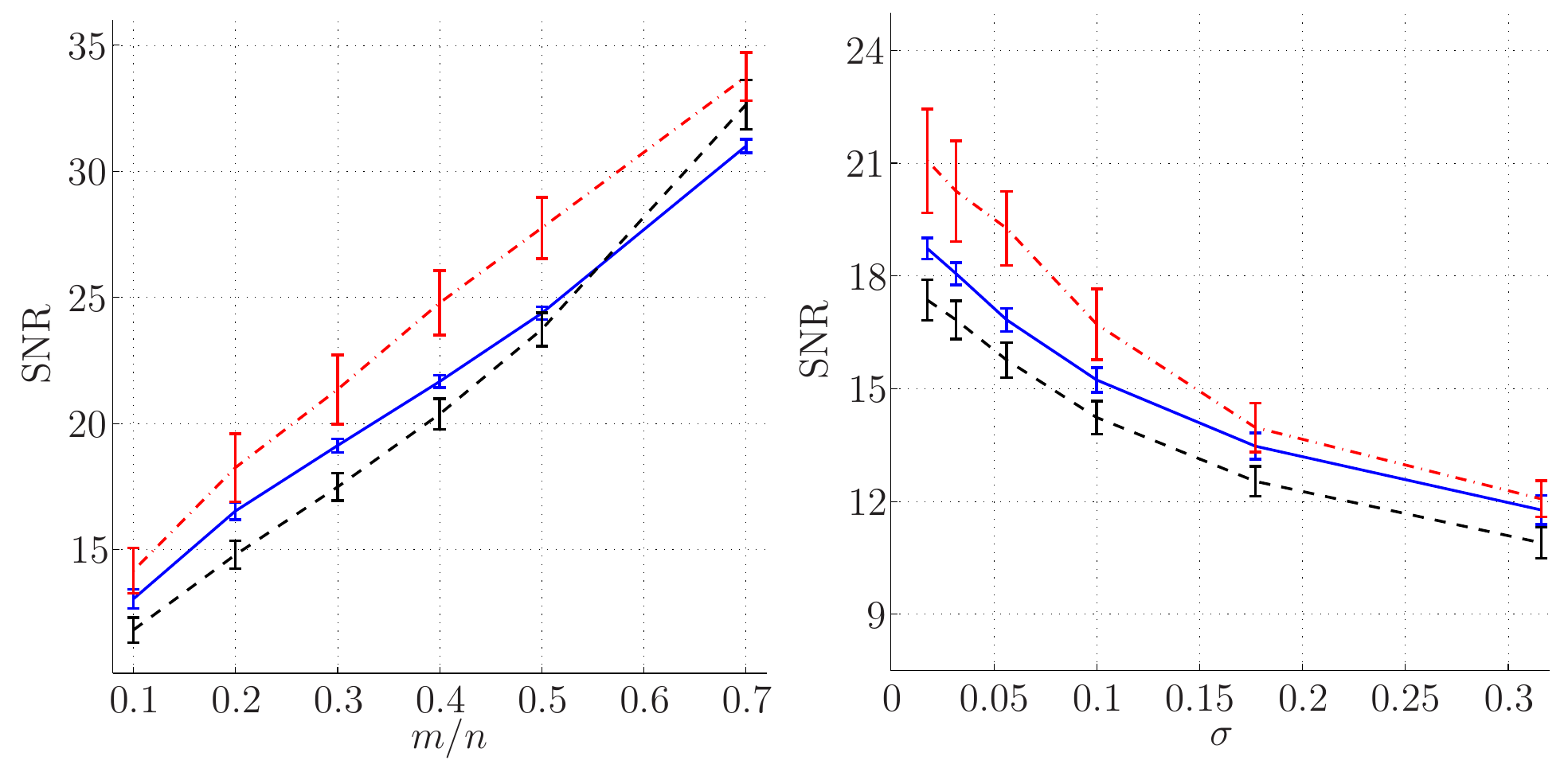}
\caption{\label{fig:CS3}Left panel: Reconstruction SNR as a function of $\dimMeas/\dimSig$. Right panel: Reconstruction SNR as a function of the noise level $\sigma$. All curves represent the mean SNR over the $30\times 5$ reconstructed images. The vertical lines show the variation of the SNR at one standard deviation. The dashed black, continuous blue, and dot-dashed red curves represent the reconstruction SNR for method \refeq{eq:BPDN}, \refeq{eq:group_sparse}, and our algorithm, respectively.}
\end{figure}
\begin{figure}
\centering
\begin{minipage}{0.325\linewidth} \centering \scriptsize Not registered \\
\includegraphics[height=0.85\linewidth, keepaspectratio]{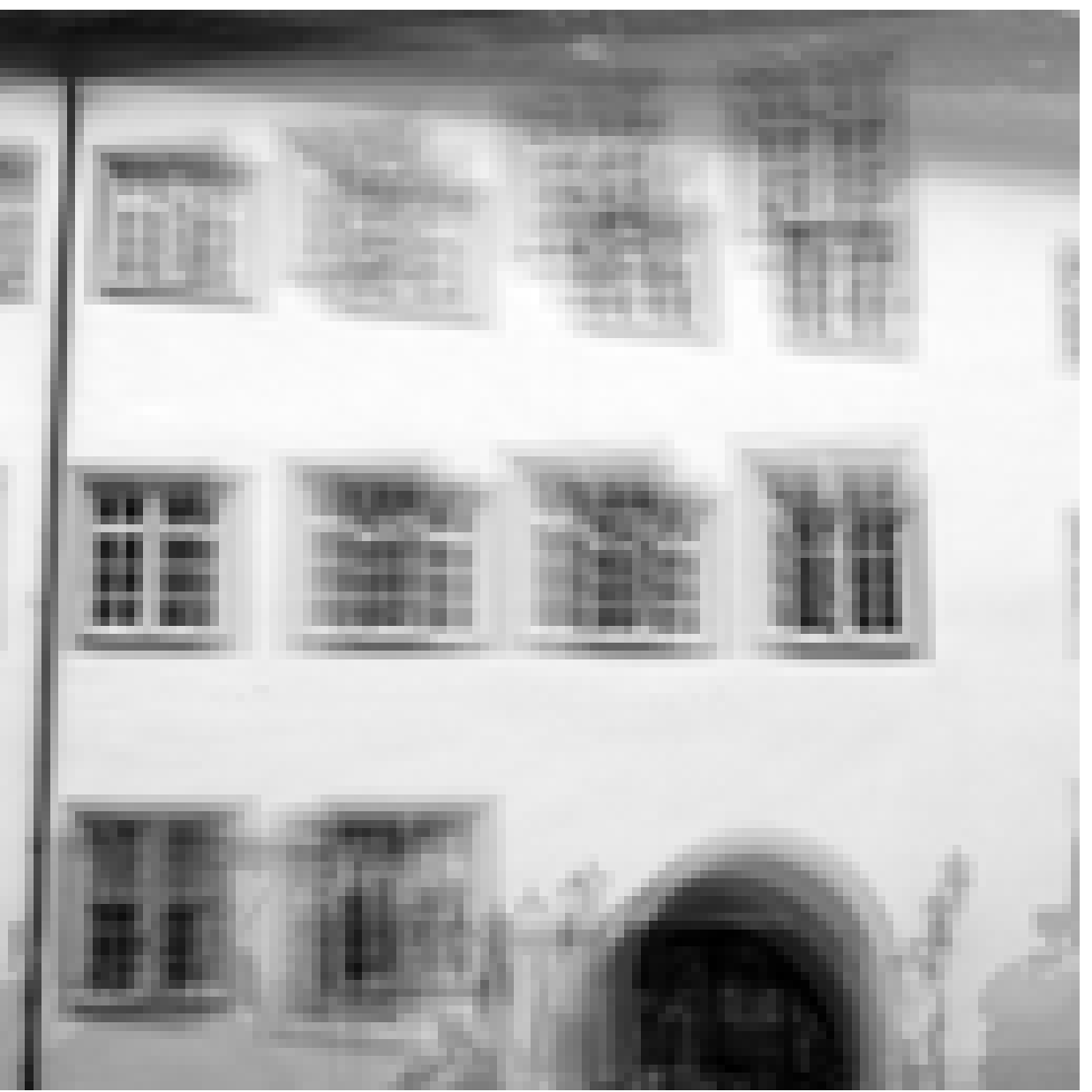}
\end{minipage}
\begin{minipage}{0.325\linewidth} \centering \scriptsize Registered - Proposed method \\
\includegraphics[height=0.85\linewidth, keepaspectratio]{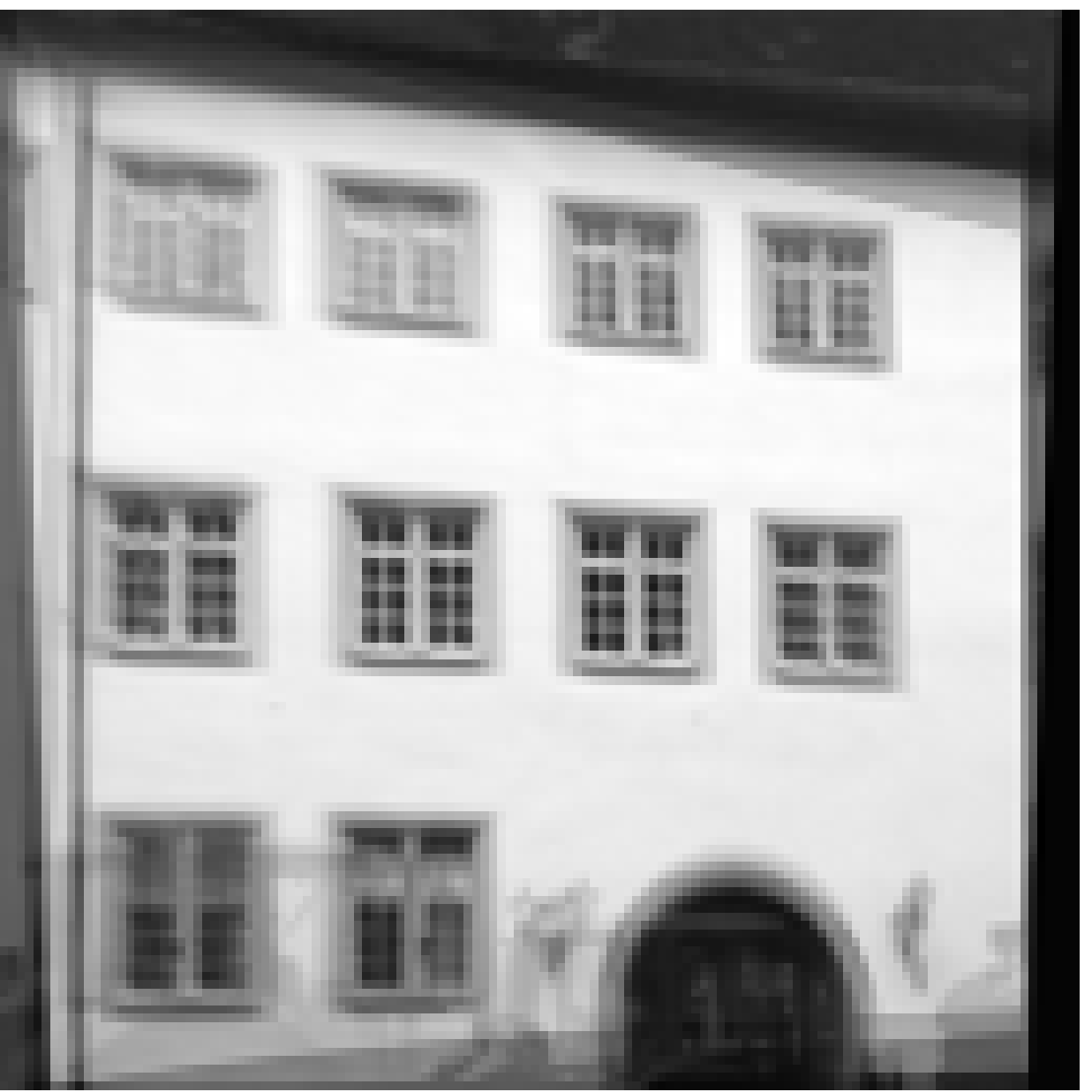}
\end{minipage}
\begin{minipage}{0.325\linewidth} \centering \scriptsize Estimated background image \\
\includegraphics[height=0.85\linewidth, keepaspectratio]{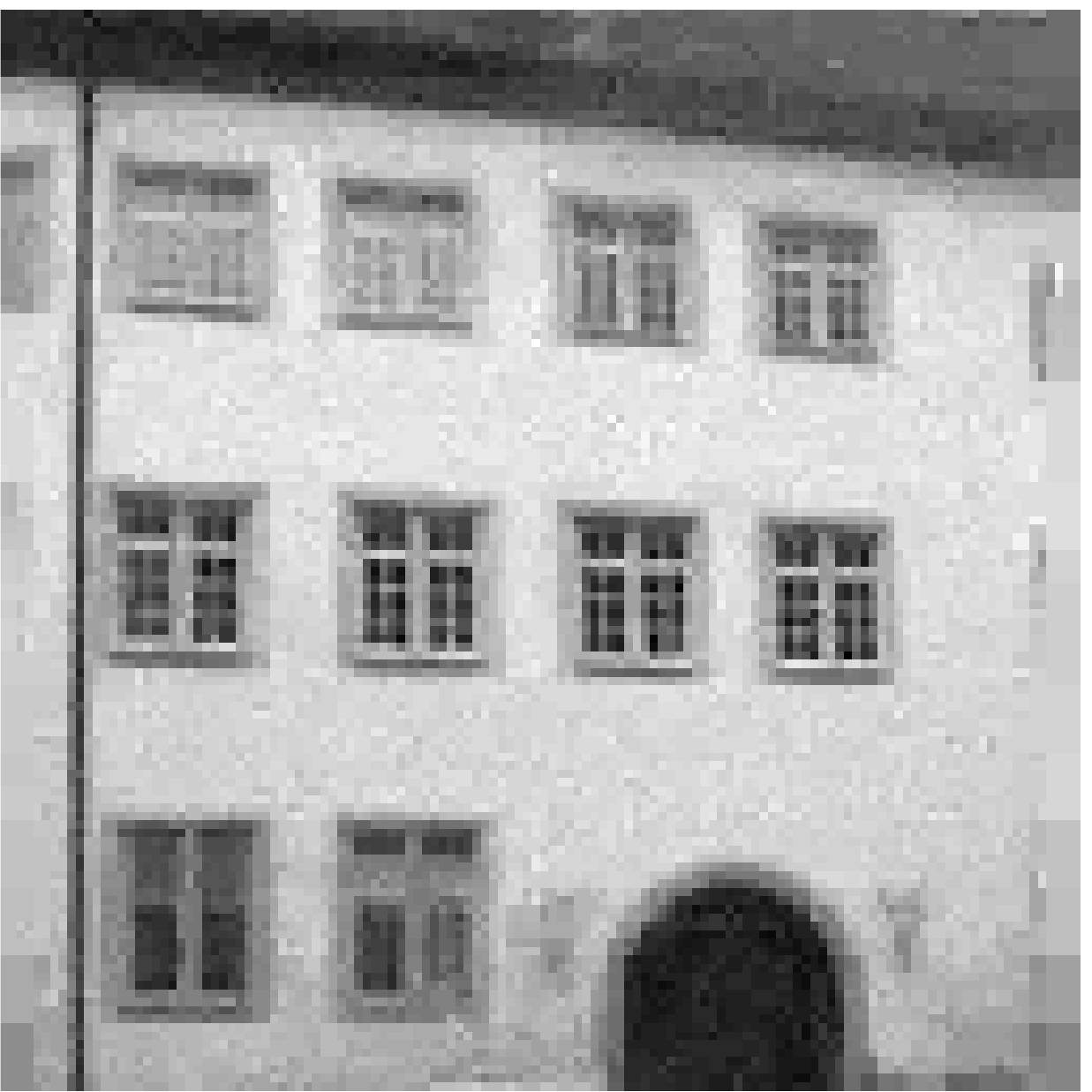}
\end{minipage}
\caption{\label{fig:CS2}Left panel: superposition of the $5$ initial images without registration. Middle panel: superposition of the $5$ initial images after registration with the estimated parameters $\param_1^*, \ldots, \param_{5}^*$. Right panel: estimated background image $\sig_0^*$.}
\end{figure}

In this section, we study the performance of our algorithm in a compressed sensing setting in comparison with other common algorithms. 

We use $\nbObs = 5$ different images\footnote{All images are in the castle-R20 dataset available at \url{cvlab.epfl.ch/~strecha/multiview/rawMVS.html} \cite{strecha08}.} of the same scene to generate $5$ different measurement vectors $\meas_1, \ldots, \meas_{\nbObs}$. These images, presented in the top panels of Fig.~\ref{fig:CS1}, contain $n = 128 \times 128$ pixels. Note that parts of the scene are sometimes occluded. The measurement vectors are obtained using the spread spectrum technique of \cite{puy12}, which consists of pre-modulating each image with a random $\pm 1$ sequence before randomly selecting $\dimMeas$ Fourier coefficients. Let $\bm{c} = (c_1, \ldots, c_\dimSig)^\adjoint \in \Rbb^{\dimSig}$ be a Rademacher sequence, $\ma{F} \in \Cbb^{\dimSig \times \dimSig}$ be the matrix that implements the $\rm 2D$ discrete Fourier transform, and {\color{black} $\Omega_j \in \Nbb^\dimMeas$ be $\nbObs$ sets of $\dimMeas$ indices selected independently and uniformly at random in $\{1, \ldots, \dimSig\}$. As we are dealing with images living in $\Rbb^\dimSig$, we can restrict the selection of the Fourier coefficients to one half of the Fourier plane. We denote $\tilde{\ma{F}} \in \Cbb^{\dimSig/2 \times \dimSig}$ the restriction of $\ma{F}$ to this half plane. The observation matrices satisfy
\begin{equation*}
\SenMa_j  = \left(
\begin{array}{c}
\tilde{\ma{F}}^{\rm r}\\
\tilde{\ma{F}}^{\rm i}
\end{array}
\right)_{\Omega_j}
\ma{C} \in \Rbb^{\dimMeas \times \dimSig},\quad j = 1, \ldots, \nbObs,
\end{equation*}
where $\tilde{\ma{F}}^{\rm r}, \tilde{\ma{F}}^{\rm i} \in \Rbb^{\dimSig/2 \times \dimSig}$ are respectively the real and imaginary parts of $\tilde{\ma{F}}$, $\ma{C} \in \Rbb^{\dimSig \times \dimSig}$ is the diagonal matrix containing the sequence $\bm{c}$, and $\left(\cdot \right)_{\Omega_j}$ restricts an $\dimSig \times \dimSig$ matrix to its rows indexed by $\Omega_j$.}

We assume here that the transformations between images can be represented by homographies and use model \refeq{eq:homography} for $\trans_{\param}$. For the prior term $\prior$, we assume that the images are sparse in the Haar wavelet basis $\ma{W} \in \Rbb^{\dimSig \times \dimSig}$. We build the block-diagonal matrix $\Dict$ by repeating $\nbObs + 1$ times the matrix $\ma{W}$ on the diagonal, and set $\prior(\sig) = \norm{\Dict^\adjoint \sig}_1$. {\color{black} The same matrix $\Dict$ is used in the cost-to-move term $\huber$}. In all simulations, we use $\regMove_{\param} = 0.1$, $\regMove_{\sig}^k = \max(0.9^k\,(20\reg), 0.1)$, $\mu = 10^{-10}$, and the following constraints apply on the transformation parameters: $\abs{1-\theta_1}, \abs{1-\theta_5} \leq 0.2$, $\abs{\theta_2}, \abs{\theta_4} \leq 0.2$, $\abs{\theta_3}, \abs{\theta_6} \leq 20$, $\abs{\theta_7}, \abs{\theta_8} \leq 0.01$, with $u_1$ and $u_2$ belonging to $\{-63, -62, \ldots, 64\}$.

In a first set of simulations, we study the performance of Algorithm $1$ in the absence of noise for different number of measurements: $\dimMeas \in \{0.1\dimSig, 0.2\dimSig, 0.3\dimSig, 0.4\dimSig, 0.5\dimSig, 0.7\dimSig\}$. In a second set of simulations, we study the performance of Algorithm $1$ at $\dimMeas = 0.3\dimSig$ for different noise levels. The noise vector $\noise$ follows an i.i.d zero-mean Gaussian distribution of standard deviation $\sigma \in \{0.01, 0.0177, 0.0316, 0.0562, 0.1, 0.177, 0.316\}$. The squared $\ell_2$-norm of the noise vector $\noise/\sigma$ follows a chi-square distribution with $\nbObs\dimMeas$ degrees of freedom and we compute the bound on the noise level $\norm{\noise}_2^2 \leq \epsilon^2$ using the $99^\th$ percentile of this distribution. Then, $\kappa$ is set to $100$ in the noiseless case and is chosen so that $\norm{\SenMa(\param^*)\sig^* - \meas}_2 \approx \epsilon$ in the presence of noise\footnote{For each noise level, we use one simulation to estimate a value of $\reg$ such that, at converge of the algorithm, $\norm{\SenMa(\param^*)\sig^* - \meas}_2 \in [0.99\epsilon, 1.01 \epsilon]$. We then use the same value of $\reg$ for all subsequent simulations.}.

For comparison, we also reconstruct the images by solving the following convex optimization problems:
\begin{equation}
\label{eq:BPDN}
\min_{\ma{X}} \norm{\ma{W}^\adjoint \ma{X}}_{1}\ \text{ subject to }\ \norm{\ma{Y} - \SenMa \ma{X}}_F \leq \boundNoise,
\end{equation}
and
\begin{equation}
\label{eq:group_sparse}
\min_{\ma{X}} \norm{\ma{W}^\adjoint \ma{X}}_{2,1}\ \text{ subject to }\ \norm{\ma{Y} - \SenMa \ma{X}}_F \leq \boundNoise,
\end{equation}
where $\ma{X} = (\sig_1, \ldots, \sig_{\nbObs}) \in \Rbb^{\dimSig \times \nbObs}$, $\ma{Y} = (\meas_1, \ldots, \meas_{\nbObs}) \in \Rbb^{\dimMeas \times \nbObs}$, $\norm{\cdot}_F$ is the Frobenius norm, $\norm{\ma{X}}_{1} = \sum_{i=1}^\dimSig \sum_{j=1}^\nbObs \abs{x_{ij}}$, and $\norm{\ma{X}}_{2,1} = \sum_{i=1}^\dimSig (\sum_{j=1}^\nbObs x_{ij}^2)^{1/2}$, with $x_{ij}$ the entry on the $i^\th$ line and $j^\th$ column of $\ma{X}$. In the first problem, the images $\sig_1, \ldots, \sig_{\nbObs}$ are reconstructed with a sparse prior in the Haar wavelet basis. This is an extension of the Basis Pursuit problem, advocated by the compressed sensing theory \cite{candes06}, to multiple signals. In the second problem, the prior term favors reconstructions which have a similar sparsity pattern in the Haar wavelet basis \cite{bach12}, therefore imposing some correlation between images.

Fig.~\ref{fig:CS3} shows the quality of the reconstructed images in term of SNR. Let $(\sig^*, \param^*) \in \Rbb^{(\nbObs+1)\dimSig} \times \Rbb^{\nbObs\dimSig}$ be the images and transformation parameters recovered with our algorithm and $\sig \in \Rbb^{\nbObs\dimSig}$ be the $\nbObs$ initial images. The reconstruction SNR satisfies $-20\log_{10}(\norm{\InterpMa(\param^*_j) \sig_0^*+ \sig_j^* - \sig_j}_2/\norm{\sig_j}_2)$ in this case. For the two other methods \refeq{eq:BPDN} and \refeq{eq:group_sparse}, the reconstruction SNR satisfies $-20\log_{10}(\norm{\sig_j^* - \sig_j}_2/\norm{\sig_j}_2)$ where $\ma{X}^* = (\sig_1^*, \ldots, \sig_{\nbObs}^*) \in \Rbb^{\dimSig \times \nbObs}$ are the reconstructed images with these methods. The first graph of Fig.~\ref{fig:CS3} shows the reconstructions SNR as a function of $\dimMeas/\dimSig$ in the absence of noise, and the second graph shows the reconstructions SNR as a function of the noise level $\sigma$ for $\dimMeas/\dimSig = 0.3$. In each case, we perform $30$ simulations with independent noise realization and choice of $\Omega_j$. The curves show the mean SNR over the $30 \times 5$ reconstructed images. The vertical lines indicate the variation of the SNR at one standard deviation. One can deduce from these curves that the best reconstructions are obtained with our method.

In Fig.~\ref{fig:CS1}, we present the ground truth images as well as reconstructions obtained with the three different methods for $\dimMeas/\dimSig = 0.3$ in the absence of noise. On can remark that the reconstructions obtained with our algorithm exhibits much finer details. Fig.~\ref{fig:CS2} shows the estimated background image, next to the superposed initial images before and after registration with the estimated parameters. One can also note that the estimated background image is free of occlusions, and that the initial images are better aligned after registration with the estimated parameters.

\subsection{Super-resolution}
\begin{figure*}
\centering
\includegraphics[height=7cm, keepaspectratio]{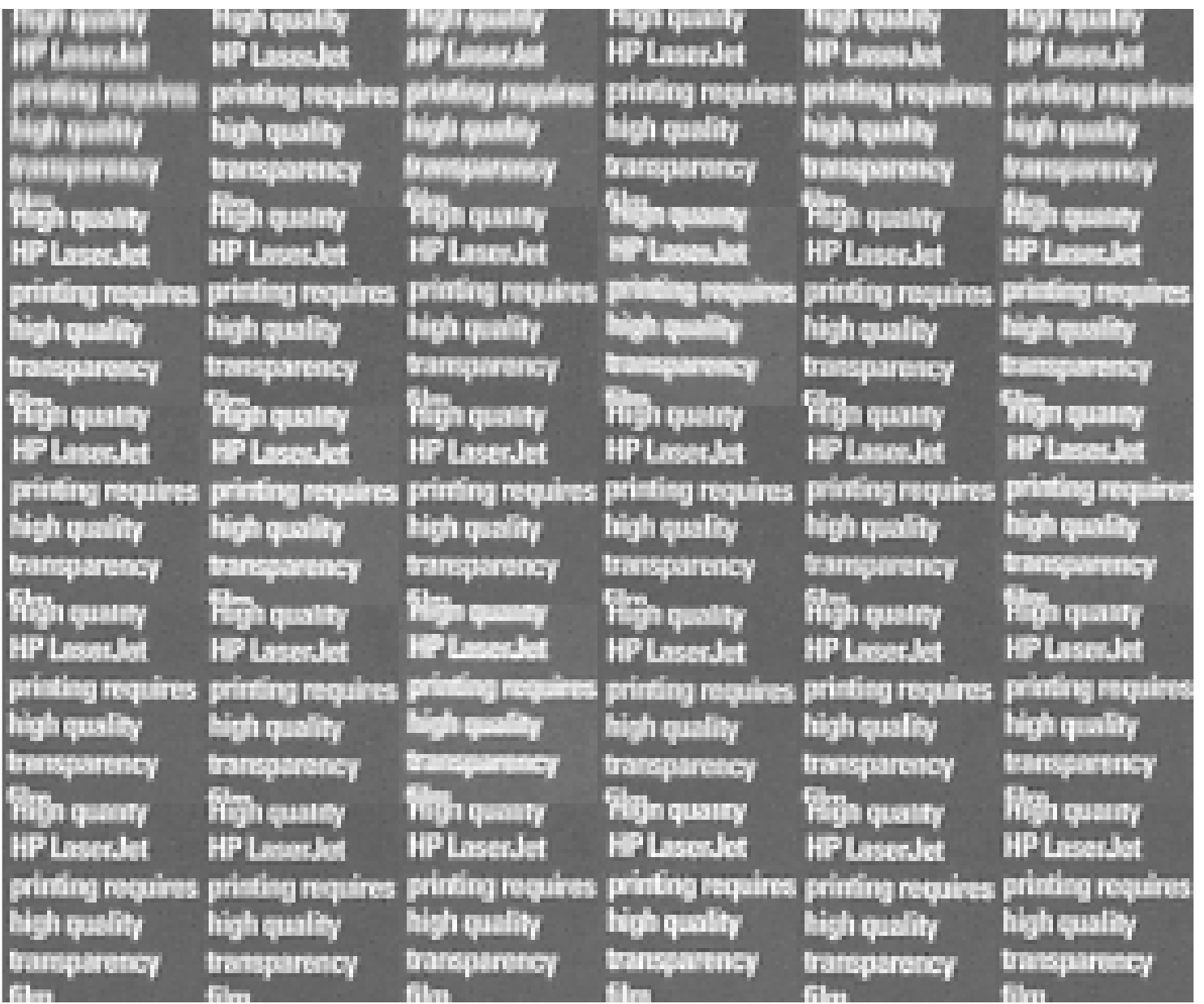}
\includegraphics[height=7cm, keepaspectratio]{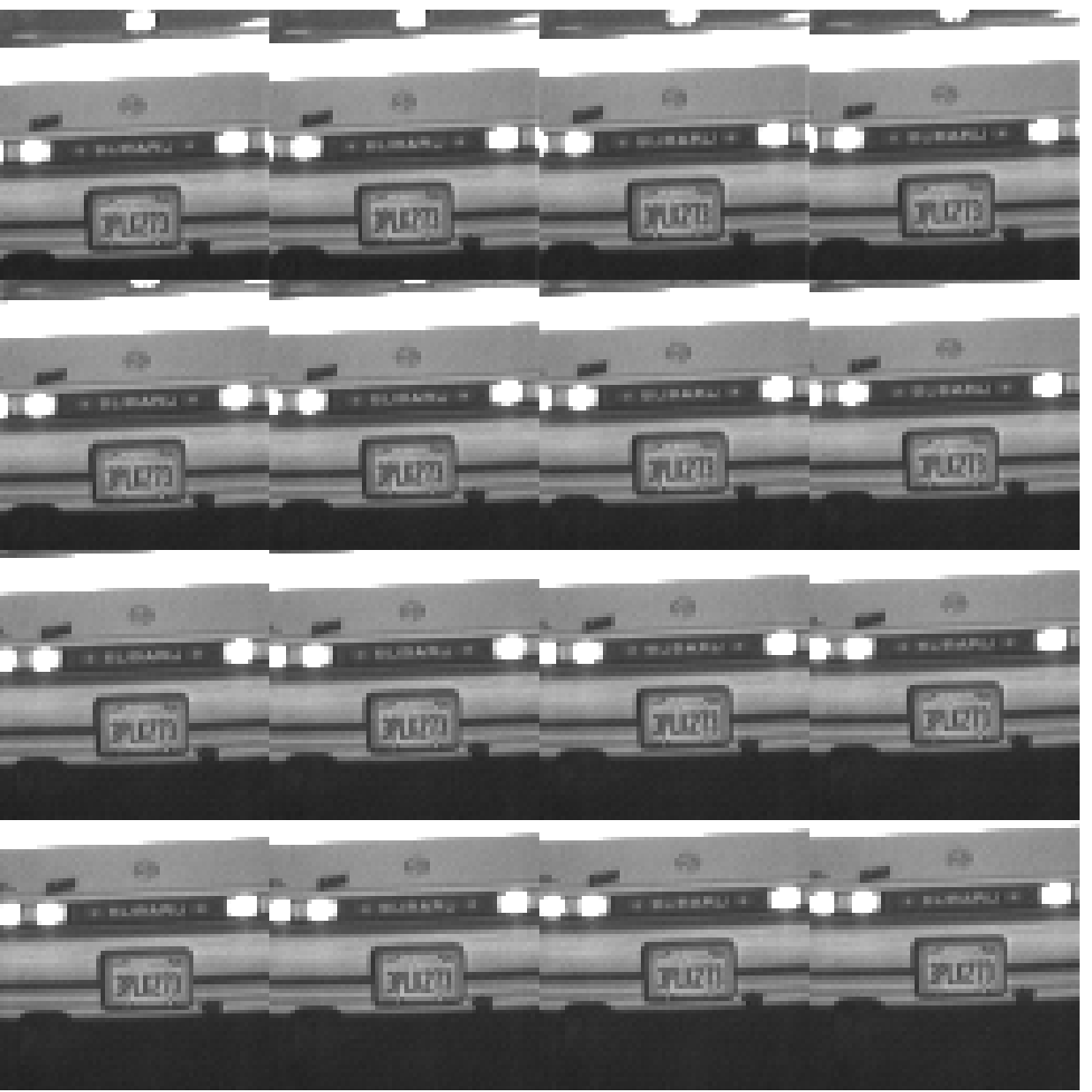}
\caption{\label{fig:sr1} Multi-frame datasets used to test the efficiency of our algorithm for super-resolution. The dataset on the right contains $30$ low-resolution images of $\dimMeas = 48 \times 48$ pixels. The dataset on the left contains $16$ low-resolution images of $\dimMeas = 64 \times 64$ pixels.}
\end{figure*}
\begin{figure*}
\centering
\begin{minipage}{3.75cm} \centering Bicubic interp. \end{minipage}
\begin{minipage}{3.75cm} \centering \hspace{-3mm} \cite{vandewalle06} +  Bicubic interp. \end{minipage}
\begin{minipage}{3.75cm} \centering \cite{vandewalle06} + \cite{pham06} \end{minipage}
\begin{minipage}{3.75cm} \centering Proposed method \end{minipage}\\
\includegraphics[width=3.75cm, keepaspectratio]{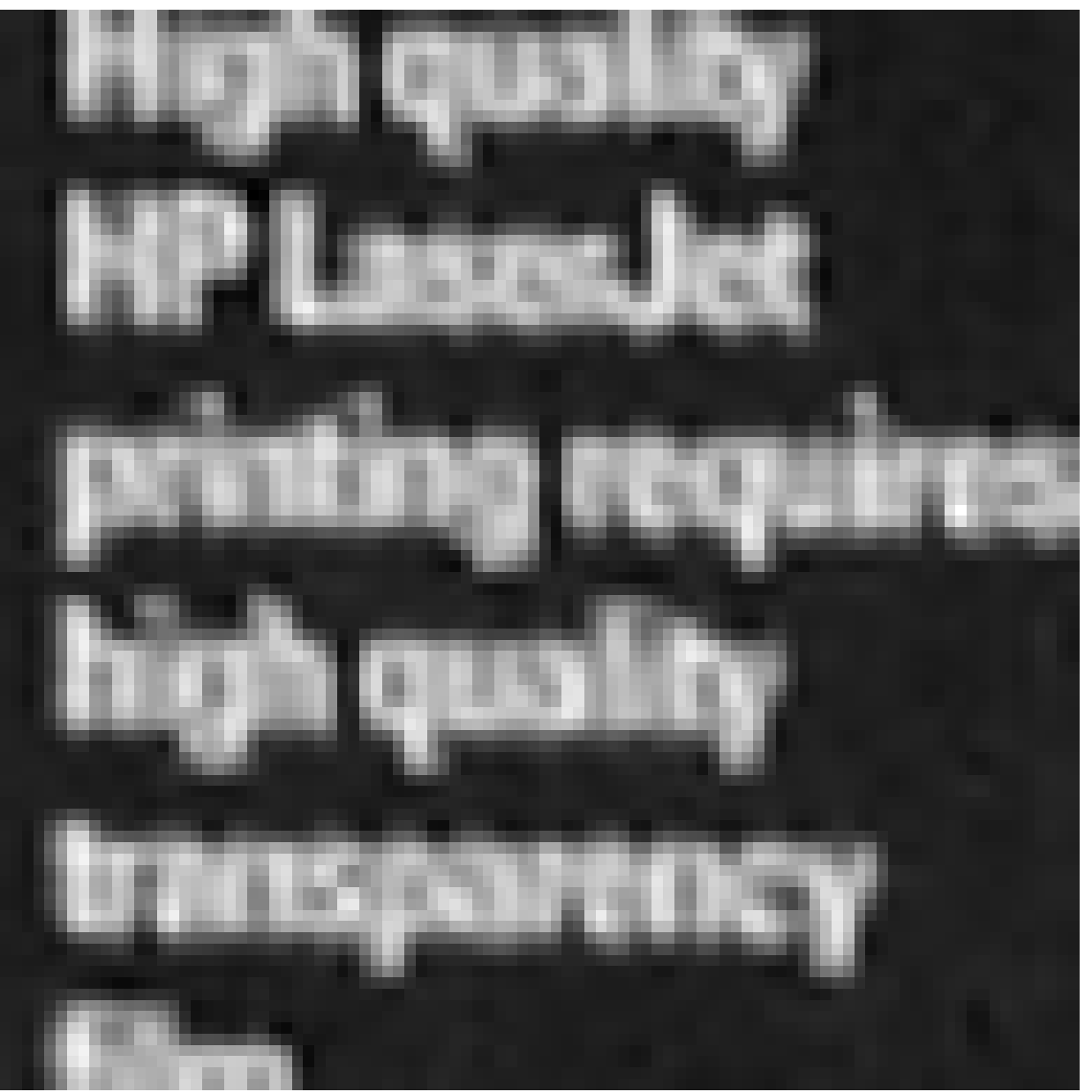}
\includegraphics[width=3.75cm, keepaspectratio]{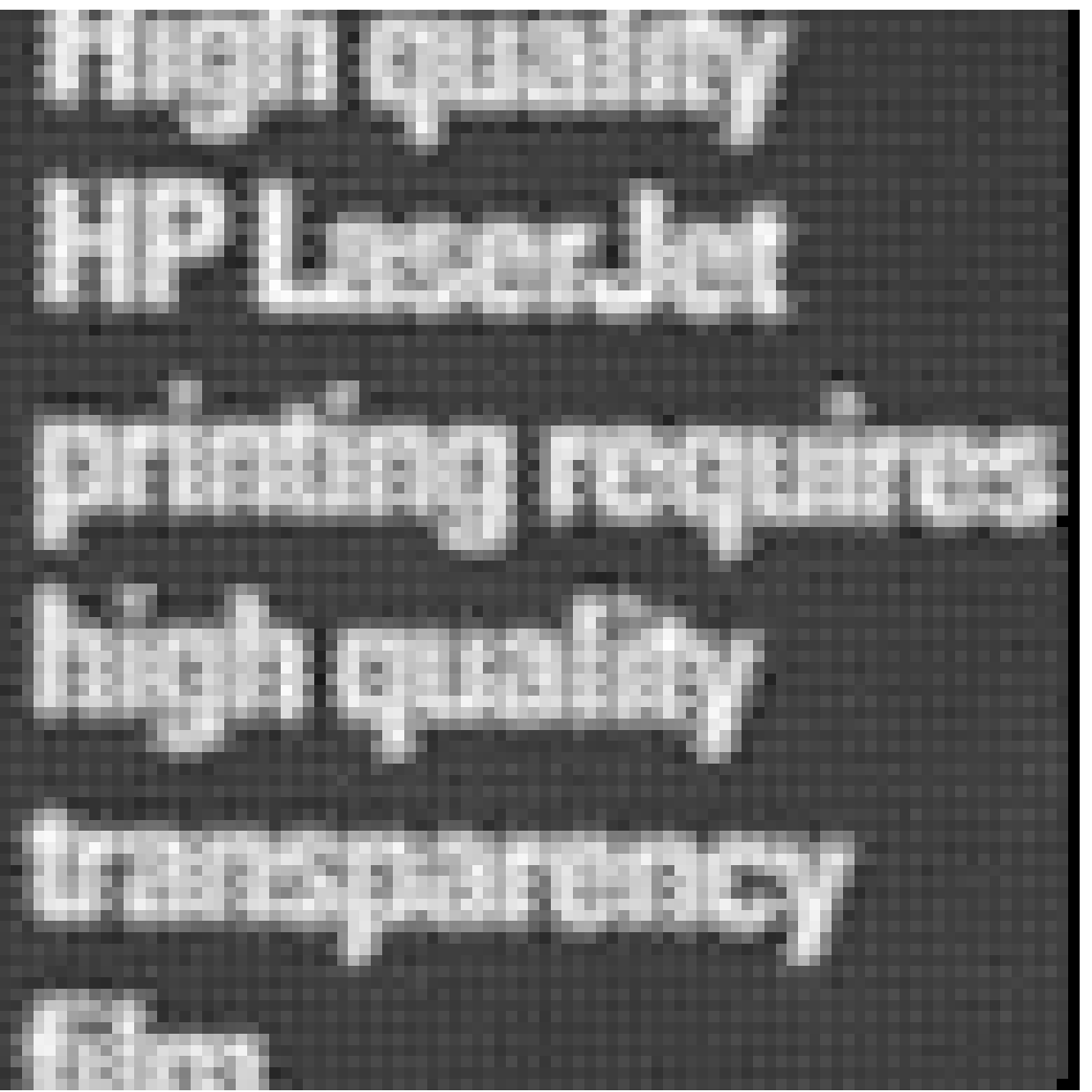}
\includegraphics[width=3.75cm, keepaspectratio]{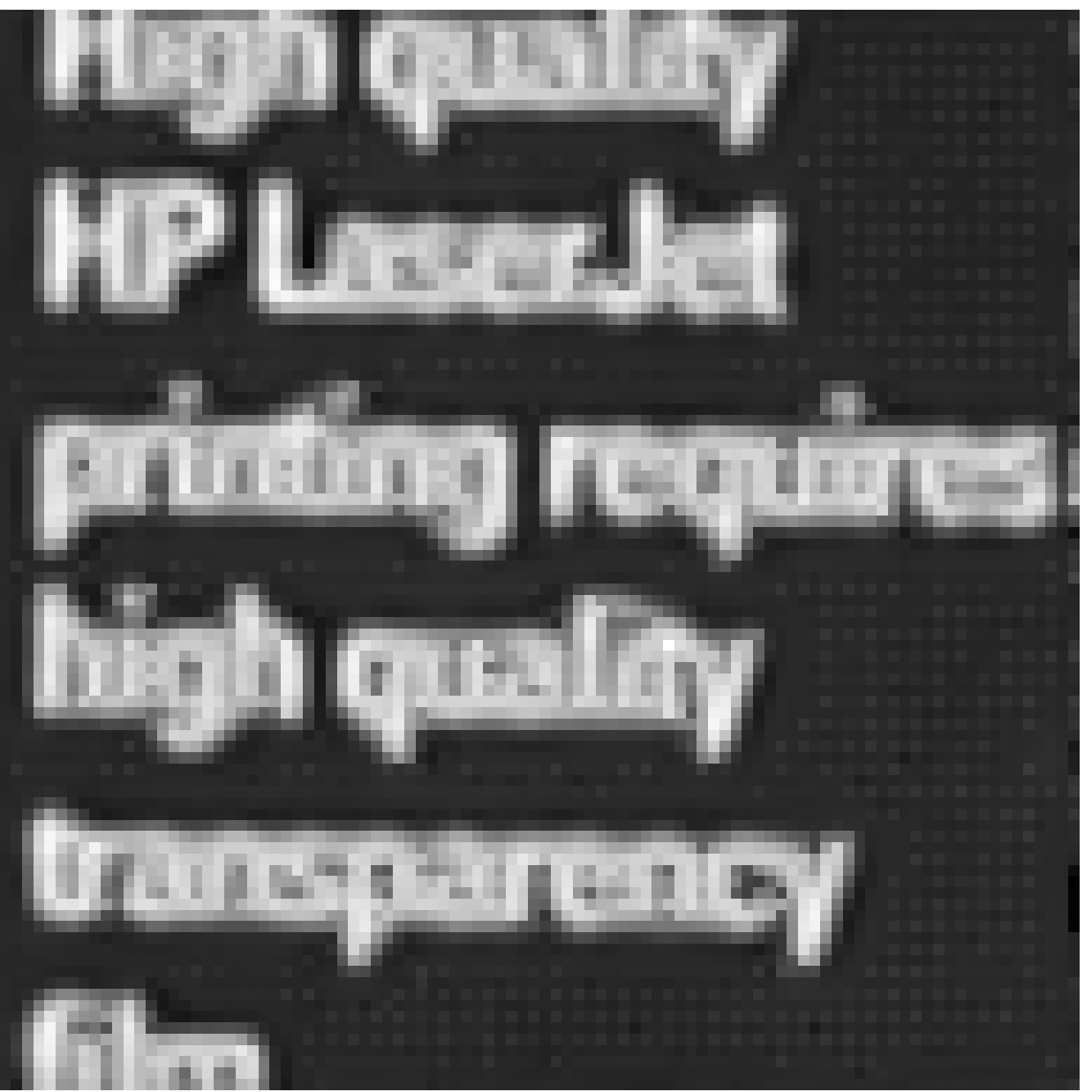}
\includegraphics[width=3.75cm, keepaspectratio]{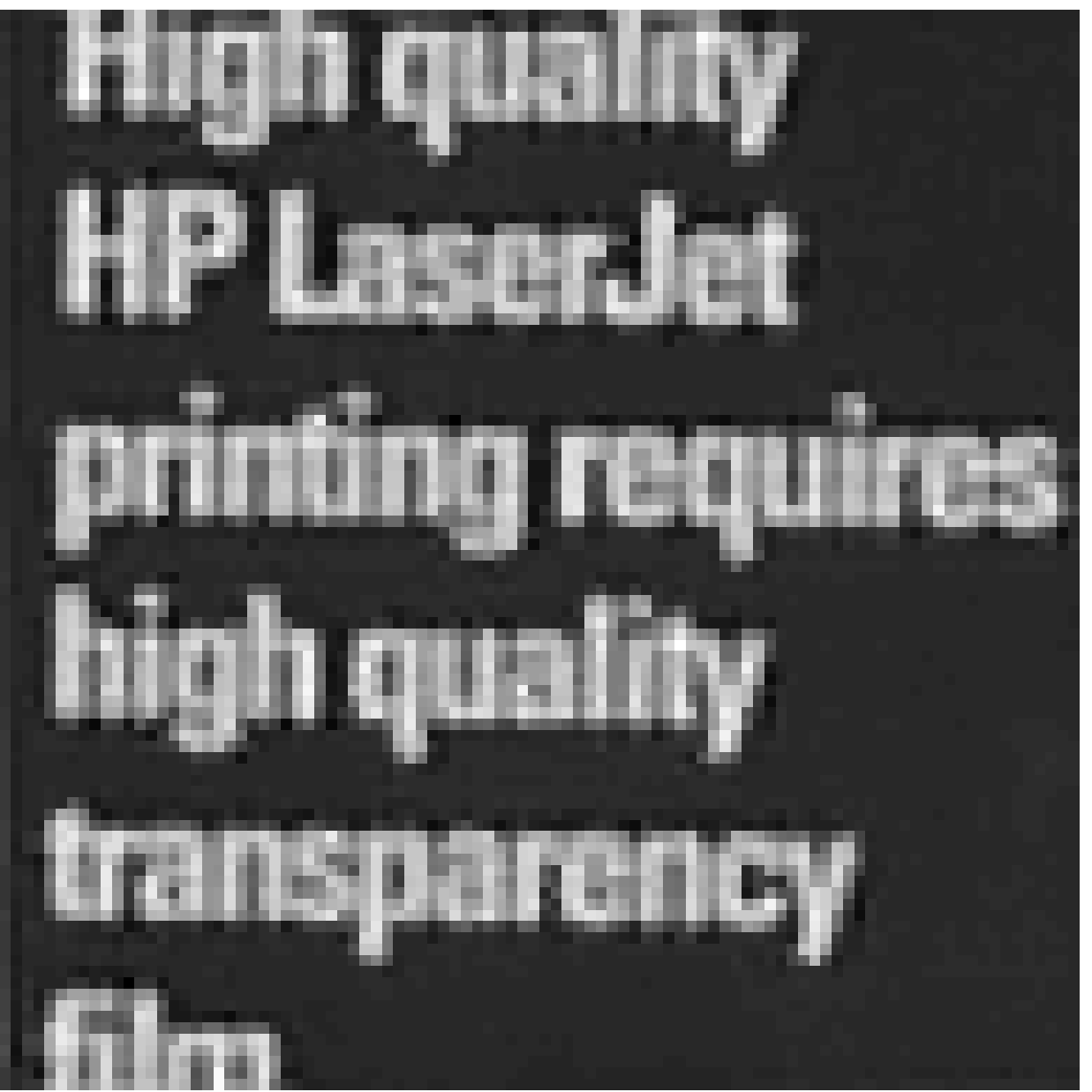}\\
\includegraphics[width=3.75cm, keepaspectratio]{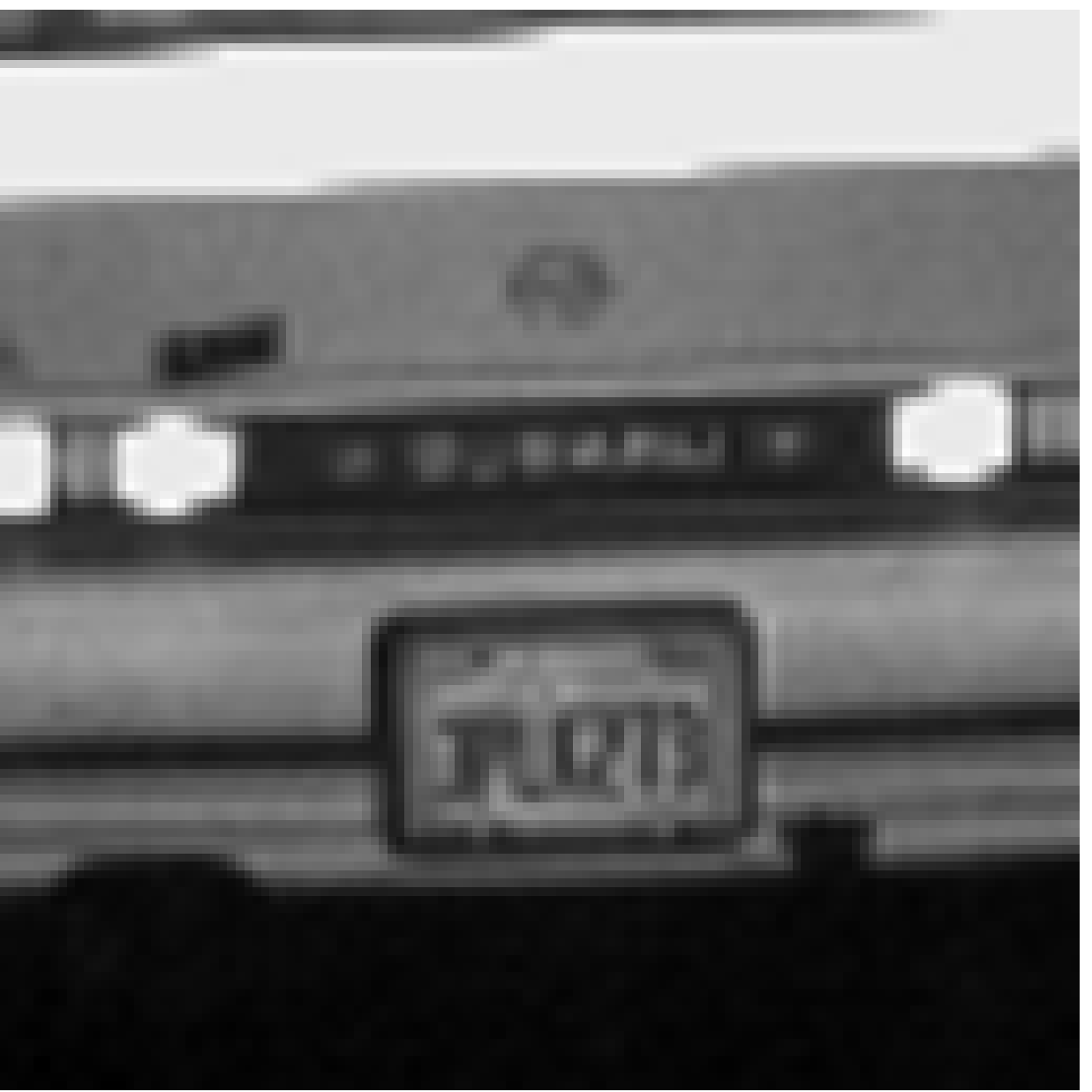}
\includegraphics[width=3.75cm, keepaspectratio]{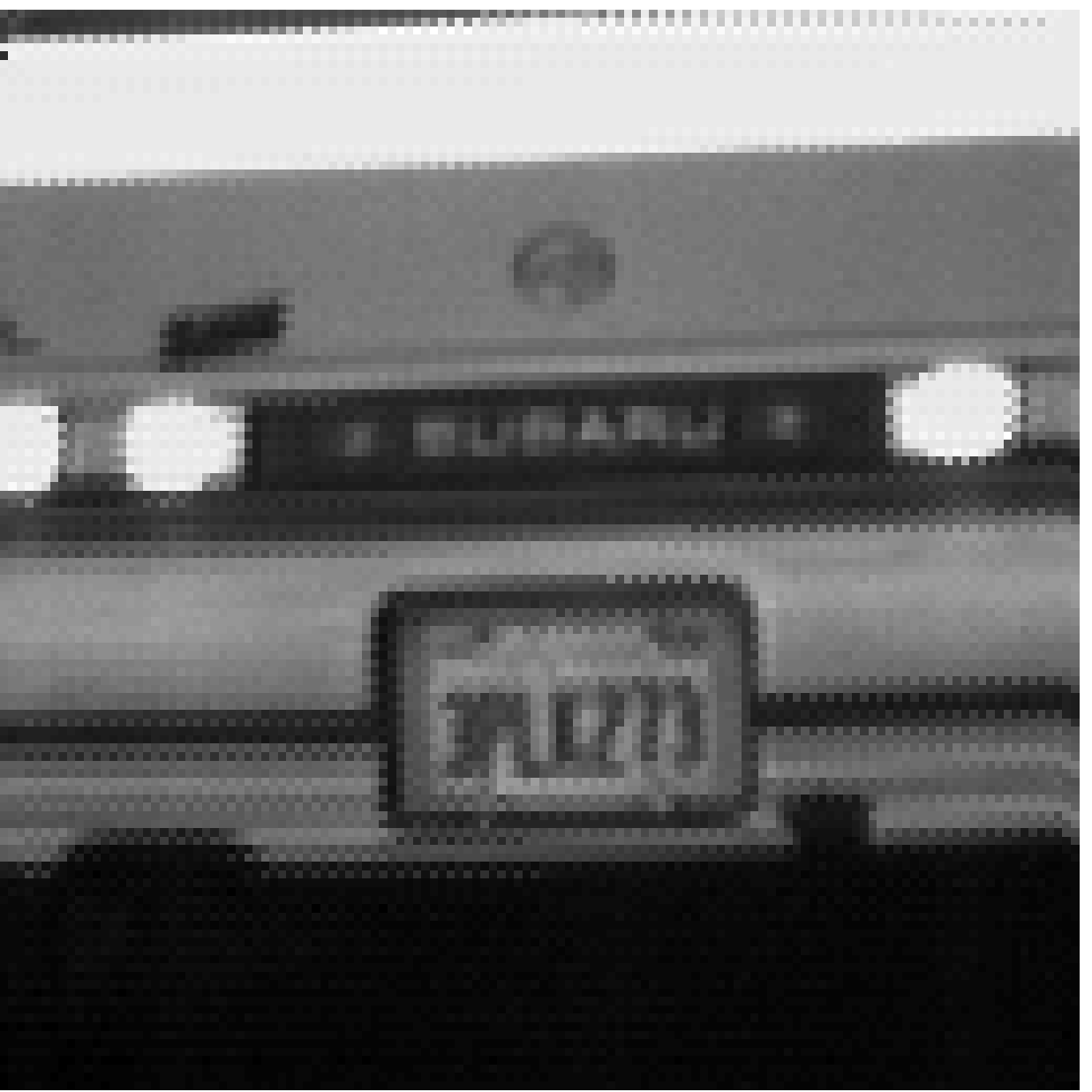}
\includegraphics[width=3.75cm, keepaspectratio]{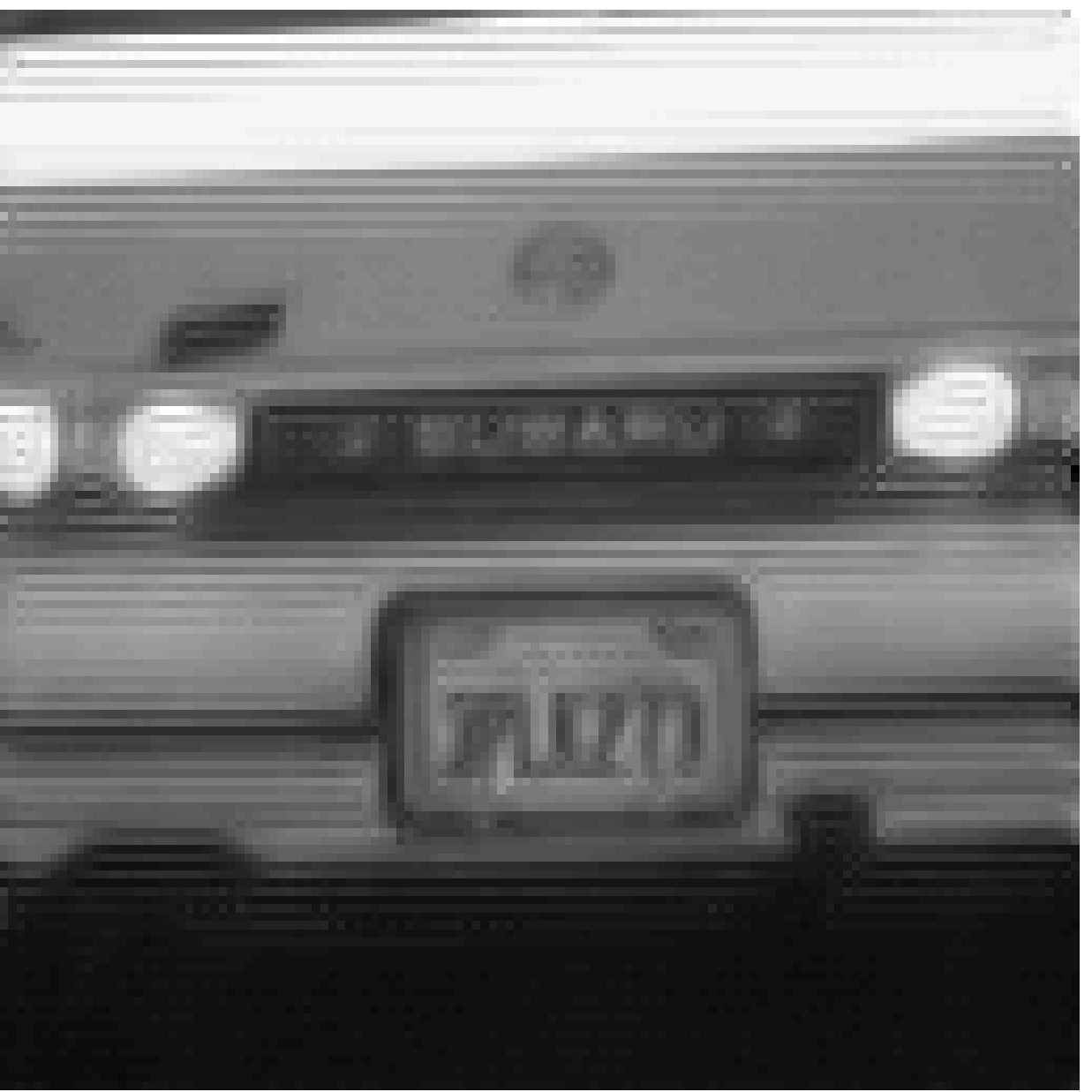}
\includegraphics[width=3.75cm, keepaspectratio]{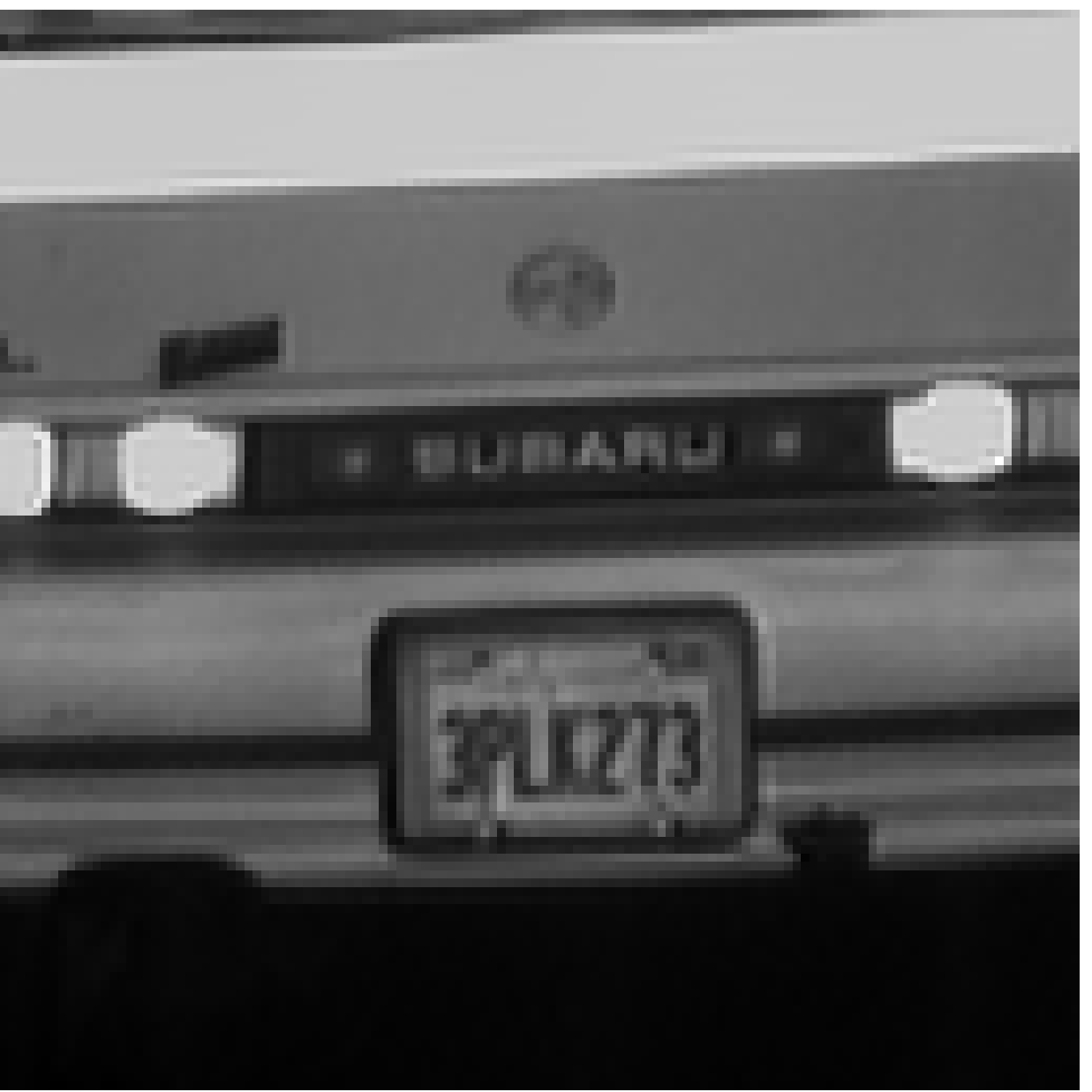}
\caption{\label{fig:sr2} From left to right: one input image of the original dataset upsampled independently using a bicubic interpolation; super-resolved image using the method described in \cite{vandewalle06} in combination with a bicubic interpolation; super-resolved image using the method described in \cite{vandewalle06} in combination with \cite{pham06}; super-resolved background image estimated by our algorithm.}
\end{figure*}

\begin{figure*}
\centering
\includegraphics[width=5cm, keepaspectratio]{figures/super_resolution/car_me-eps-converted-to.pdf}
\includegraphics[width=5cm, keepaspectratio]{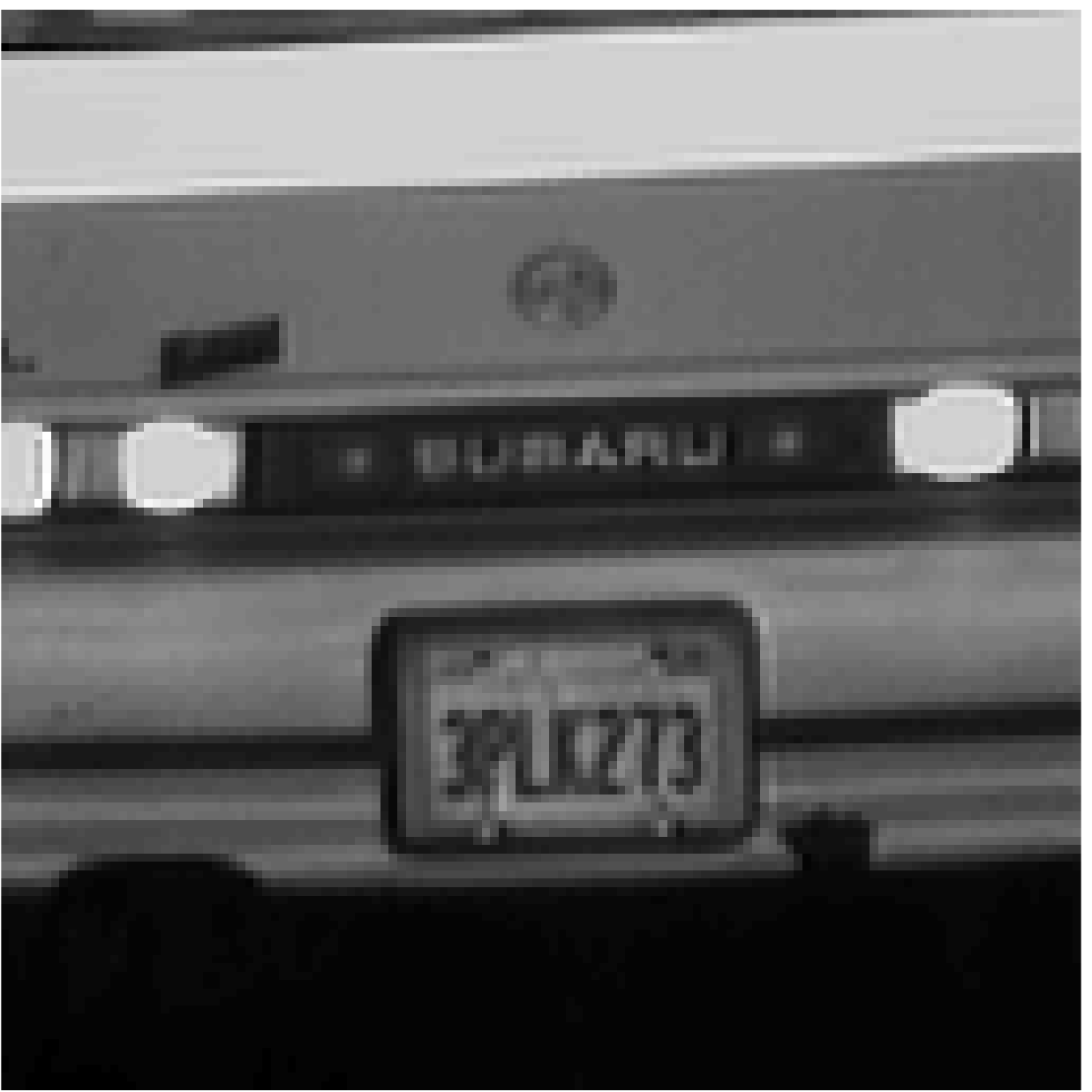}
\caption{\label{fig:sr3} Super-resolved background image estimated by our algorithm using the Haar wavelets (left) or the Daubechies wavelets (right) for $\Dict$ in the cost-to-move function $\huber$.}
\end{figure*}

{\color{black}
In this last experiment, we demonstrate the performance of our algorithm for super-resolution from multiple frames in comparison with the method described in \cite{vandewalle06}, whose code is available at \url{lcav.epfl.ch/software/superresolution}. We test both algorithms on the datasets\footnote{Datasets available at \url{users.soe.ucsc.edu/~milanfar/software/sr-datasets.html} (credit: Peyman Milanfar)} presented in Fig.~\ref{fig:sr1}.

To increase the resolution of the available images by a factor of $2$ in both dimensions with our method, we use the measurement model \refeq{eq:measurement_model} where all the sensing matrices $\SenMa_j$, with $j = 1, \ldots, \nbObs$, model the same blurring and downsampling operator that averages all the pixels in a rectangular window of $2 \times 2$ pixels and uniformly downsamples the resulting blurred image by a factor of $2$ in both dimensions. The vectors $\meas_1, \ldots, \meas_{\nbObs}$ are initialized with the low-resolution images. The transformations between images are assumed to be affine, and we use model \refeq{eq:affine} for $\trans_{\param}$. The transformation parameters are constrained as follows: $\abs{1-\theta_1}, \abs{1-\theta_5} \leq 0.2$, $\abs{\theta_2}, \abs{\theta_4} \leq 0.2$, $\abs{\theta_3}, \abs{\theta_6} \leq 30$, with $u_1$ and $u_2$ belonging to $\{-\dimSig/2+1, \ldots, \dimSig/2\}$. The matrix $\Dict$ is built using the Haar wavelet basis. We also assume that the high-resolution images $\sig_0, \ldots, \sig_{\nbObs}$ are sparse in the gradient domain and set $\prior(\sig) = \sum_{j=0}^\nbObs g(\sig_j)$ in \refeq{eq:objective_function}, where $g \colon \Rbb^{\dimSig} \rightarrow \Rbb$ is the isotropic Total Variation (TV) norm. Let us recall that the vector $\sig_j$ contains the samples $\csig_j(\pos_k)$, with $k = 1, \ldots, \dimSig$ and $\pos_k = (u_1^k, u_2^k)$ (see Section \ref{sec:problem_formulation}). The isotropic TV norm $g$ satisfies
\begin{align*}
g(\sig_j) = \sum_{k=1}^\dimSig \sqrt{\abs{\nabla_1\,\csig_j(\pos_k)}^2 + \abs{\nabla_2\,\csig_j(\pos_k)}^2},
\text{ where }
\left\{
\begin{array}{l}
\nabla_1\,\csig_j(\pos_k) = \csig_j(u_1^{k+1}, u_2^{k}) - \csig_j(u_1^{k}, u_2^{k}),\\
\nabla_2\,\csig_j(\pos_k) = \csig_j(u_1^{k}, u_2^{k+1}) - \csig_j(u_1^{k}, u_2^{k}),\\
\end{array}\right.
\end{align*}
assuming that $\csig(u_1^{\dimSig+1},\, \cdot) = \csig(\cdot\,, u_2^{\dimSig+1}) = 0$. Note that $g$ and, consequently, $\prior$ are semi-algebraic. Finally, we run Algorithm $1$ to minimize \refeq{eq:objective_function} with $\reg=100$, $\regMove_{\param} = 0.1$, $\regMove_{\sig}^k = \max(0.9^k\,(20\reg), 0.1)$, and $\mu = 10^{-10}$. 

The first dataset contains $\nbObs = 30$ low-resolution images of $\dimMeas = 48 \times 48$ pixels. The super-resolved images thus contain $\dimSig = 96 \times 96$ pixels. In the top panel of Fig.~\ref{fig:sr2}, we present one super-resolved image obtained by upsampling independently one of the original images using a bicubic spline interpolation, our super-resolved background image, and two super-resolved images obtained with the code of \cite{vandewalle06}. These last two images are obtained by first registering the low resolution images with the technique described in \cite{vandewalle06}. The first high-resolution image is then obtained by using a bicubic interpolation to combine the information of all registered low-resolution images. The second high-resolution image is obtained using the technique described in \cite{pham06}. On this first dataset, one can easily remark that our super-resolved image exhibits much shaper details than the images obtained with the other methods. In particular, the contours of the letters are better defined and the text is easier to read.

The second dataset contains $\nbObs = 16$ low-resolution images of $\dimMeas = 64 \times 64$ pixels. The super-resolved images thus contain $\dimSig = 128 \times 128$ pixels. In Fig.~\ref{fig:sr2}, we present one super-resolved image obtained by upsampling independently one of the original images using a bicubic spline interpolation, our super-resolved background image, and two super-resolved images obtained with the code of \cite{vandewalle06}, as described before. On this second dataset, one can notice that the brand of the car and license plate number are much better reconstructed with our method. The details are sharper, confirming again the superiority of our technique.

Let us mention the fact that the registration quality obtained with the method described in \cite{vandewalle06} is very sensitive to the choice of the reference frame. The results presented here are the best we were able to obtain after a careful choice of the reference frame. For most of the other choices of reference frame, this method was not able to register properly the low-resolution images and the subsequent super-resolved images were suffering of strong artifacts. Note that our method does not suffer of this issue as the algorithm automatically selects a reference coordinate system within the allowed range $\ConsParam$.

Finally, to show the effect of the choice of $\Dict$ in the cost-to-move function $\huber$, we run again Algorithm $1$ by building $\Dict$ with the Daubechies $8$ wavelets instead of the Haar wavelets. We present in Fig.~\ref{fig:sr3} the reconstructions obtained with both wavelet bases. We do not observe any significant differences between both reconstructions. The choice of $\Dict$ in the cost-to-move term $\huber$ thus does not seem to affect significantly the quality of the final reconstruction.
}

%

\subsection{Running time of the algorithm}
{\color{black}
We finally conclude this section by mentioning the computation times required to minimize the non-convex functional \refeq{eq:minimization_problem} using our Matlab implementation of Algorithm $1$. The running times mentioned have been obtained using a desktop computer (Windows $7$, Intel Core i$7$ $3.50$Ghz, $16$GB of RAM) and the parallel toolbox of Matlab to estimate the transformation parameters simultaneously. 

For the robust image alignment problem, our implementation of the algorithm has been running for about $14$ minutes to align $6$ images of $256 \times 256$ pixels. For the compressed sensing problem, around $7$ minutes have been necessary to reconstruct $5$ images of $128 \times 128$ pixels from $30$ percent of measurements. Finally, for the super-resolution problem, it has taken about $21$ minutes to estimate a high resolution image of the car with $128 \times 128$ pixels from the $16$ low-resolution input images.

Compared to the other methods we tested, the running time of our algorithm is longer. However, we would like to highlight that our implementation is general to be able to handle all these examples. We have not used the specificities of each problem to optimize the code and reduce the computation time. Furthermore, we have noticed that only a few iterations of the algorithm are necessary to estimate the transformation parameters. During the last iterations, only the estimation of the background and foreground images are refined. Therefore, an approximate solution can be obtained by stopping the algorithm after a few iterations and then solve one last time problem \refeq{eq:step1} with $\regMove_{\sig} = 0$ to estimate the images. Note however that the sequence of estimates does not converge to a critical point of $\obj$ with such an approach.
}

\section{Conclusion}
\label{sec:conclusion}
We presented a method to solve multi-view imaging problems where one has to reconstruct several images of a scene from only few linear observations made at different viewpoints. Each observed image is modeled as the sum of a geometrically transformed background image and of a foreground image, modeling possible occlusions. We considered here global geometric transformations represented by few parameters, such as translations, affine transformations, or homographies. The proposed reconstruction method jointly estimates the images (background and foreground) and the transformation parameters by minimizing a non-convex functional. We studied the convergence of the proposed algorithm and showed that the generated sequence of estimates converges to a critical point of the non-convex functional for commonly used priors and transformation models. {\color{black} Numerical experiments show that the proposed algorithm is competitive with well-known methods in robust image alignment, compressed sensing, and super-resolution. The method is able to robustly register a set of images in the presence of large displacements, and the separation background/foreground is also very accurate, even when only few random measurements of the images are accessible.}

Finally, we would like to conclude by highlighting the potential interest of the proposed technique for free breathing coronary magnetic resonance imaging (MRI) \cite{stehning05}. One major challenge in this application is to handle respiratory motion properly. The low acquisition speed in MRI forces researchers to invent new techniques to compensate for the inevitable motions occurring during the acquisition. To avoid motions due to heart contractions, an ECG signal is acquired to ensure that the Fourier measurements are always taken at the same instant in the heart cycle. A few measurements are then taken at each cycle. As the number of measurements acquired during one cycle is too small to accurately reconstruct an image of the heart, one should combine the measurements of several cycles to gather enough information. It is thus mandatory to properly compensate for the respiratory motion occurring between heart beats in the reconstruction method. Note that simple geometric transformations, such as translations, are already sufficient to reach good image quality \cite{bonanno12}. Consequently, as the technique developed in this paper automatically compensates for such geometric transformations, it is a promising method for such an application. {\color{black} Note that this hope is confirmed by the preliminary results presented in \cite{puy13} on real MRI data.}

\appendix \titleformat{\section}[hang]{\color{blue1}\large\bfseries\centering}{Appendix \thesection}{0mm}{}[]

\section{}
\label{app:proof}
To establish the third point of Theorem \ref{th:main}, we need the following theorem establish by Attouch \emph{et al.} in \cite{attouch11}.
\begin{theorem}[Theorem 6.2, \cite{attouch11}]
\label{th:attouch}
Let $p$ be a positive integer larger than $2$, $\sig = (\sig_1, \ldots, \sig_p)$ be a vector in  $\set{H} = \Rbb^{n_1} \times \ldots \times \Rbb^{n_p}$, and $g \colon \set{H} \rightarrow \Rbb \cup \{+\infty\}$ a function of the form $g(\sig) = Q(\sig) + \sum_{i=1}^p g_i(\sig_i)$, where $Q \colon \set{H} \rightarrow \Rbb$ is a $\set{C}^1$ function with locally Lipschitz continuous gradient and $g_i : \Rbb^{n_i} \rightarrow \Rbb \cup \{+\infty\}$ is a proper lower semicontinuous function, $i = 1, \ldots, p$. Assume that $g$ is bounded from below and satisfies the \KL property. Let $(\sig^k)_{k \in \Nbb}$ and $(\vec{\nu}^k)_{k \in \Nbb}$ be sequences such that
\begin{align}
\label{eq:conditions_attouch}
&g_i(\sig_i^{k+1}) + Q(\sig_1^{k+1}, \ldots, \sig_{i-1}^{k+1}, \sig_{i}^{k+1}, \ldots, \sig_{p}^{k})\ + a \norm{\sig_i^{k+1} - \sig_i^k}_2^2\ \nonumber \\
& \hspace{5mm} \leq\ \, g_i(\sig_i^{k}) + Q(\sig_1^{k+1}, \ldots, \sig_{i-1}^{k+1}, \sig_{i}^{k}, \ldots, \sig_{p}^{k}),\nonumber \\
&\vec{\nu}_i^{k+1} \in \partial g_i(\sig_i^{k+1}), \\
&\norm{\vec{\nu}_i^{k+1} + \partial_{x_i} Q(\sig_1^{k+1}, \ldots, \sig_{i}^{k+1}, \sig_{i+1}^{k}, \ldots, \sig_{p}^{k})}_2 
\leq b \norm{\sig_i^{k+1} - \sig_i^k}_2, \nonumber
\end{align}
where $i$ ranges over $\{1, \ldots, p\}$ and $a, b > 0$. Then, if the sequence $(\sig^k)_{k \in \Nbb}$ is bounded, it converges to some critical point $\sig^*$ of $g$.
\end{theorem}

The objective function $\obj$ defined in \refeq{eq:objective_function} is equal to 
\begin{align*}
\obj(\sig, \param) = \reg\ \qua (\sig, \param) + \prior(\sig) + \ind_{\ConsParam} (\param), 
\end{align*}
where $\qua \colon \Rbb^{(\nbObs+1)\dimSig} \times \Rbb^{\nbObs\nbParam} \rightarrow \Rbb$ satisfies $\qua (\sig, \param) = \norm{\SenMa(\param) \sig - \meas}_2^2$. The objective function has the the form required by Theorem \ref{th:attouch}. Furthermore, $\prior$ and $\ind_{\ConsParam_i}$ are proper lower semicontinuous functions, and, using the assumptions of Theorem \ref{th:main}, $\qua$ is twice continuously differentiable and thus have locally Lipschitz continuous gradient. We also note that $\obj$ is bounded below and satisfies the \KL property. To prove the third point of Theorem \ref{th:main}, it remains to show that the sequence generated by Algorithm $1$ is bounded and satisfies conditions \refeq{eq:conditions_attouch}. 

First, due to the constraints that apply on $\param$, it is obvious that the sequence $(\param^k)_{k \in \Nbb}$ is bounded. Therefore, assuming that $(\sig^k)_{k \in \Nbb}$ is bounded, as in the third point of Theorem \ref{th:main}, is enough to ensure the boundedness condition of the entire sequence of estimates.

Second, $\norm{\sig^{k+1} - \sig^k}_2$ tends to $0$ as $k \rightarrow  + \infty$, as established in the second point of Theorem \ref{th:main}. Therefore there exists $k_0 \in \Nbb$ such that for all $k \geq k_0$, $\huber(\Dict^\adjoint(\sig^{k+1} - \sig^k)) = \norm{\Dict^\adjoint(\sig^{k+1} - \sig^k)}_2^2/(2\mu) = \norm{\sig^{k+1} - \sig^k}_2^2/(2\mu)$, as $\Dict \Dict^\adjoint = \ma{I}_{(\nbObs+1)\dimSig}$. Let $k$ be larger than $k_0$ in the following. Inequality \refeq{eq:decrease_step_1} shows that
\begin{align*}
\prior(\sig^{k+1}) + \reg\ \qua(\sig^{k+1}, \param^{k}) + \frac{\regMoveMin}{4 \mu} \norm{\sig^{k+1} & - \sig^k}_2^2 \leq \prior(\sig^{k}) + \reg\ \qua(\sig^{k}, \param^{k}).
\end{align*}
Furthermore, the first order optimality condition of Problem \refeq{eq:step1} shows that there exists $\vec{\nu}_{\sig}^{k+1} \in \partial \prior(\sig^{k+1})$ such that
\begin{align*}
\vec{\nu}_{\sig}^{k+1} + \reg\ \partial_{\sig} \qua(\sig^{k+1}, \param^k) + \frac{\regMove_{\sig}^k}{2\mu}\, (\sig^{k+1} - \sig^{k}) = 0,
\end{align*}
which implies, 
\begin{align*}
\norm{\vec{\nu}_{\sig}^{k+1} + \reg\ \partial_{\sig} \qua(\sig^{k+1}, \param^k)}_2 &\leq \frac{\regMoveMax}{2\mu}\, \norm{\sig^{k+1} - \sig^{k}}_2.
\end{align*}

Finally, we deduce from equation \refeq{eq:decrease_step_2} that
\begin{align*}
\ind_{\ConsParam}(\param^{k+1}) + \reg\ \qua(\sig^{k+1}, \param^{k+1}) + \frac{\reg\regMove_{\param}}{2} & \norm{\param^{k+1} - \param^{k}}_2^2 \leq \ind_{\ConsParam}(\param^{k}) + \reg\ \qua(\sig^{k+1}, \param^{k}).
\end{align*}
And, using the first order optimality condition of the problems \refeq{eq:step2}, we conclude that there exists $\vec{\nu}_{\param_j}^{k+1} \in \partial \ind_{\ConsParam_j}(\param_j^{k+1})$ such that
\begin{align*}
\vec{\nu}_{\param_j}^{k+1} + \reg\ \nabla \qua^{k+1}_j(\param_j^k) + \reg\ [\Hess(\param_j^k) + 2^i \regMove_{\param} \ma{I}_{\nbParam}] (\param_j^{k+1} - \param_j^{k}) = 0,
\end{align*}
for all $j = 1, \ldots, \nbObs$. Let $\Gamma$ denote the Lipschitz constant of $\nabla Q$ on a product of balls $\Ball_{\sig} \times \Ball_{\param}$ containing the sequence $(\sig_k, \param_{k})_{k \in \Nbb}$. By construction of $i$ in Algorithm $1$, we necessarily  have $2^i \regMove_{\param} \leq \eta = \max(2\Gamma, \regMove_{\param})$. Remark also that, on the same products of balls,
their exists $\Lambda$ such that the singular values of the matrices $\Hess(\param^k_j)$, $j = 1, \ldots, \nbObs$, are bounded above by $\Lambda$. Therefore,
\begin{align*}
\norm{\vec{\nu}_{\param_j}^{k+1} + \reg\ \nabla \qua^{k+1}_j(\param_j^{k+1})}_2
&=
\norm{\vec{\nu}_{\param_j}^{k+1} +  \reg \nabla \qua^{k}_j(\param_j^{k}) +  \reg \nabla \qua^{k+1}_j(\param_j^{k+1}) - \reg \nabla \qua^{k+1}_j(\param_j^{k})}_2\\
&\leq
\norm{\vec{\nu}_{\param_j}^{k+1} + \reg \nabla \qua^{k+1}_j(\param_j^{k})}_2 + \reg \norm{\nabla \qua^{k+1}_j(\param_j^{k+1}) - \nabla \qua^{k+1}_j(\param_j^{k})}_2\\
&\leq
\reg(\Lambda + \eta)\ \norm{\param_j^{k+1} - \param_j^{k}}_2 + \reg\Gamma \norm{\param_j^{k+1} - \param_j^{k}}_2\\
& = \reg(\Lambda + \eta + \Gamma)\ \norm{\param_j^{k+1} - \param_j^{k}}_2,
\end{align*}
which yields,
\begin{align*}
\norm{\vec{\nu}_{\param}^{k+1} + \reg\ \partial_{\param} \qua(\sig^{k+1}, \param^{k+1})}_2
\leq \reg\ (\Lambda + \eta + \Gamma)\ \norm{\param^{k+1} - \param^{k}}_2,
\end{align*}
with $\vec{\nu}_{\param}^{k+1} = ((\vec{\nu}_{\param_1}^{k+1})^\adjoint, \ldots, (\vec{\nu}_{\param_\nbObs}^{k+1})^\adjoint)^\adjoint \in \partial \ind_{\ConsParam}(\param^{k+1})$. The sequence $(\sig^k, \param^k)_{k \geq k_0}$ thus satisfies conditions \refeq{eq:conditions_attouch} with $a = \min(\regMoveMin/(4 \mu), \reg\regMove_{\param}/2)$ and $b = \max(\regMoveMax/(2 \mu), \reg(\Lambda + \eta + \Gamma))$. This terminates the proof of the third point of Theorem \ref{th:main}.

\bibliographystyle{acm}
\bibliography{biblio}

\end{document}